\documentclass[10pt]{article}

\usepackage[accepted]{tmlr}  % CAMERA-READY: de-anonymized + "Published in TMLR" header + OpenReview line
\usepackage{hyperref}
\usepackage{url}

\usepackage{dblfloatfix}
\usepackage{caption}
\usepackage{placeins}
\usepackage{tikz}
\usetikzlibrary{shapes,shadows,positioning,calc}
\usepackage{subcaption}
\usepackage{xcolor}
\IfFileExists{fontawesome5.sty}{\usepackage{fontawesome5}}{%
  \newcommand{\faLightbulb}{\textbullet}%
  \newcommand{\faRobot}{\textbullet}%
  \newcommand{\faExclamationTriangle}{\textbf{!}}%
  \newcommand{\faCheckCircle}{\checkmark}%
  \newcommand{\faFlag}{\textbf{F}}%
}
\usepackage[utf8]{inputenc}
\usepackage{float}
\usepackage{microtype}
\usepackage{booktabs}
\usepackage{multirow}
\usepackage{graphicx}
\usepackage{amsmath}
\usepackage{amssymb}
\usepackage{enumitem}
\usepackage{colortbl}

\hypersetup{
  pdftitle={Small Updates, Big Doubts: Does Fine-tuning Enhance Uncertainty Awareness for Large Language Models?},
  pdfauthor={Xu Hu, Yifan Zhang, Songtao Wei, Chen Zhao, Qiannan Li, Bingzhe Li, Feng Chen},
  pdfsubject={Transactions on Machine Learning Research},
  pdfkeywords={hallucination detection, uncertainty quantification, parameter-efficient fine-tuning, LoRA, large language models},
  colorlinks=true,
  linkcolor=[rgb]{0.10,0.20,0.55},
  citecolor=[rgb]{0.10,0.35,0.45},
  urlcolor=[rgb]{0.45,0.12,0.12},
  breaklinks=true
}
\IfFileExists{algorithm.sty}{\usepackage{algorithm}\usepackage{algorithmic}}{}

\usepackage{tcolorbox}
\usepackage{listings}
\tcbuselibrary{skins,breakable,listings}

\definecolor{promptbg}{gray}{0.94}
\definecolor{promptframe}{gray}{0.42}
\lstdefinestyle{promptstyle}{
  basicstyle=\ttfamily\scriptsize,
  breaklines=true,
  breakatwhitespace=true,
  postbreak=\mbox{\textcolor{promptframe}{$\hookrightarrow$}\space},
  columns=fullflexible,
  keepspaces=true,
  showstringspaces=false,
  upquote=true,
  extendedchars=true,
  literate={’}{{'}}1 {‘}{{`}}1 {“}{{"}}1 {”}{{"}}1
           {…}{{...}}3 {→}{{$\rightarrow$}}1 {‑}{{-}}1
           {–}{{--}}1 {—}{{---}}1
}
\newtcblisting{promptbox}[1]{%
  breakable, enhanced, listing only,
  colback=promptbg, colframe=promptframe,
  coltitle=white, fonttitle=\bfseries,
  title={#1},
  arc=3pt, boxrule=0.8pt,
  left=4pt, right=4pt, top=4pt, bottom=4pt,
  listing options={style=promptstyle}
}
\newtcolorbox{casestudybox}[1]{
  colback=casebg,
  colframe=caseblue,
  fonttitle=\bfseries\large,
  title={#1},
  breakable,
  enhanced,
  boxrule=1pt,
  arc=3pt,
  left=8pt,right=8pt,top=8pt,bottom=8pt
}

\definecolor{graybg}{gray}{0.95}
\definecolor{dangerred}{RGB}{220,53,69}
\definecolor{safegreen}{RGB}{40,167,69}
\definecolor{warningorange}{RGB}{255,193,7}
\definecolor{caseblue}{RGB}{0,123,255}
\definecolor{casebg}{RGB}{248,249,250}
\definecolor{corrblue}{RGB}{220,240,255}
\definecolor{detectpink}{RGB}{255,230,235}
\definecolor{improveblue}{RGB}{230,240,250}
\definecolor{avegreen}{RGB}{255,230,235}

\newcommand{\takeawaymark}{{\color{orange}\faLightbulb}~}

\newcommand{\blfootnote}[1]{%
  \begingroup
  \renewcommand{\thefootnote}{}\footnote{#1}%
  \addtocounter{footnote}{-1}%
  \endgroup
}

\newcommand{\appendixtablestyle}{%
  \small
  \setlength{\tabcolsep}{4pt}%
  \renewcommand{\arraystretch}{1.05}%
}

\def\month{08}   %% 按 2026-08 填; TMLR 的 month/year 是【发表月份】, 若 AE 验收在别的月份请改
\def\year{2026} %% 同上
\def\openreview{\url{https://openreview.net/forum?id=BaW14onJX2}}

\title{Small Updates, Big Doubts:\\
Does Fine-tuning Enhance\\
Uncertainty Awareness for Large Language Models?}

\author{\name Xu Hu\thanks{Equal contribution.} \email Xu.Hu@utdallas.edu \\
      \addr The University of Texas at Dallas
      \AND
      \name Yifan Zhang\footnotemark[1] \email Yifan.Zhang3@utdallas.edu \\
      \addr The University of Texas at Dallas
      \AND
      \name Songtao Wei \email Songtao.Wei@utdallas.edu \\
      \addr The University of Texas at Dallas
      \AND
      \name Chen Zhao \email chen\_zhao@baylor.edu \\
      \addr Baylor University
      \AND
      \name Qiannan Li \email qiannan.li@ucdavis.edu \\
      \addr The University of California, Davis
      \AND
      \name Bingzhe Li\thanks{Corresponding author.} \email Bingzhe.Li@utdallas.edu \\
      \addr The University of Texas at Dallas
      \AND
      \name Feng Chen\footnotemark[2] \email feng.chen@utdallas.edu \\
      \addr The University of Texas at Dallas}

\begin{document}
\raggedbottom  % slack collapses to the page bottom instead of opening holes mid-column
\maketitle
% ---- TMLR 要求: LLM 使用声明必须是第一页脚注 (jmlr.org/tmlr/faq.html) ----
\blfootnote{\textbf{Use of Large Language Models.} We used AI assistants for manuscript polishing and figure design, and the authors verified all resulting content.}

\begin{abstract}
Fine-tuning is widely used to adapt large language models (LLMs), but does it also make their uncertainty more informative about when they are wrong? We study this question through answer-level hallucination detectability after in-domain adaptation in fact-seeking question answering. Operationally, we ask whether uncertainty scores better separate externally verified correct and incorrect answers after fine-tuning. We first characterize this phenomenon through a PEFT-centered study spanning three open-weight LLMs (LLaMA-3.2-3B, Qwen-2.5-3B, and Mistral-7B), three QA benchmarks (TriviaQA, NQ-Open, and SQuAD), three PEFT methods (LoRA, DoRA, and PiSSA), and seven black-box detectors. PEFT improves answer accuracy by only a few percentage points, yet raises AUROC for semantic-consistency and confidence-based detectors by up to approximately 14\% while token-level entropy detectors show weaker or mixed changes. To test whether this pattern is specific to parameter-efficient adaptation, we evaluate one matched full-parameter SFT configuration for each backbone–dataset pair. The same qualitative detector-family pattern appears across all nine evaluated configurations, showing that it is not unique to PEFT in these settings. Within the PEFT experiments, stronger multi-sample detector performance is associated with lower answer-length variability, which we use as a proxy for one dimension of output-format diversity. However, short-answer few-shot prompting changes response length and style without reproducing either the variability reduction or the detector gains. White-box linear probes of PEFT-adapted hidden states yield mixed results. Overall, both parameter-efficient and full-parameter fine-tuning can strengthen some externally observable uncertainty signals in the evaluated settings, but detector-family heterogeneity and the mixed PEFT white-box results do not support a detector-independent improvement in uncertainty awareness.
\end{abstract}

\begin{figure}[t]
\centering
\includegraphics[width=\textwidth]{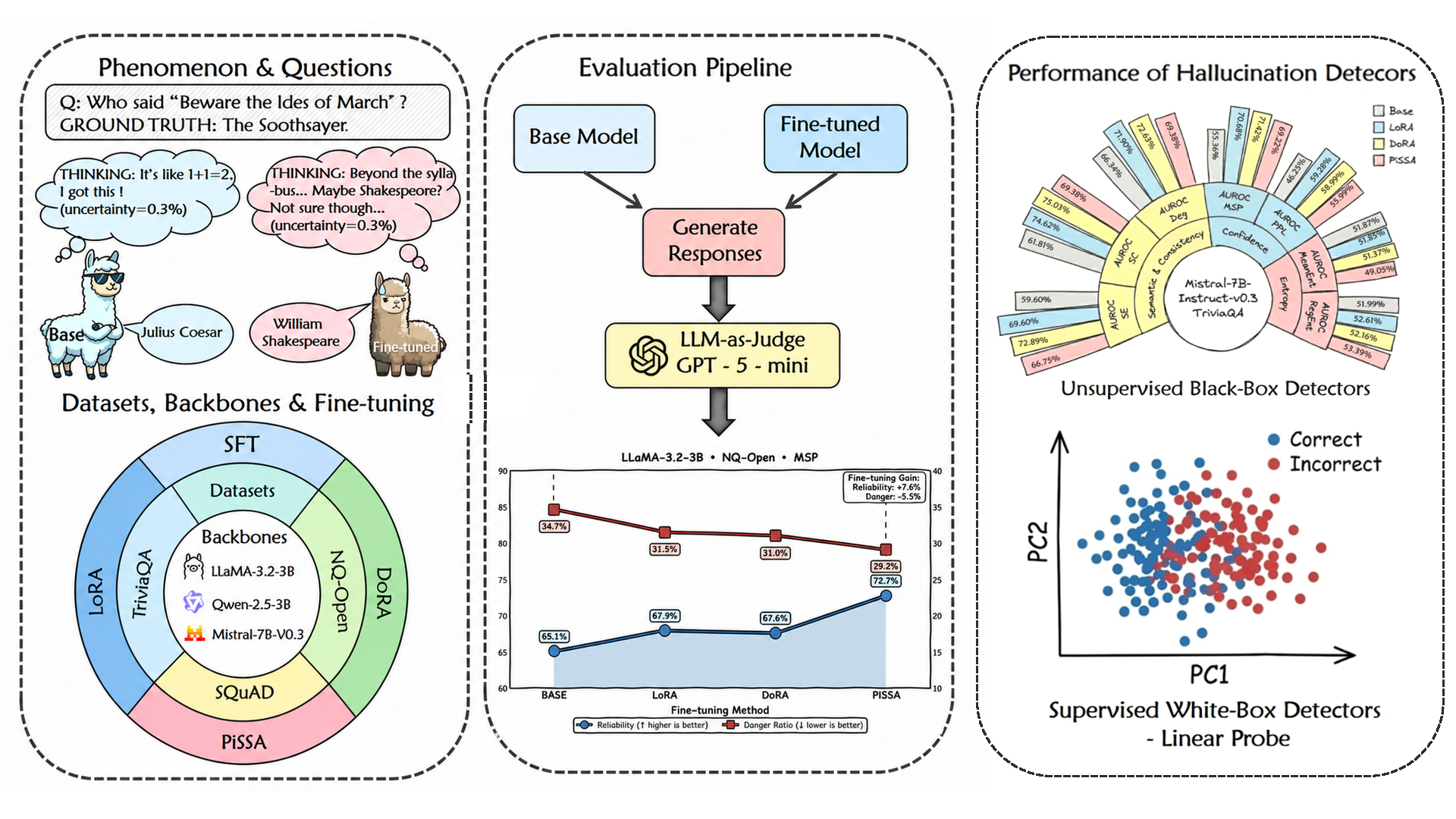}
\caption{Overview of our empirical study of how fine-tuning affects hallucination detection.}
\vspace{-1em}
\end{figure}

\section{Introduction}
Hallucination, the generation of fluent but fabricated or factually incorrect content, remains a central obstacle to deploying LLMs in knowledge-intensive applications \citep{du2024haloscope,manakul2023selfcheckgpt,qiu2024semantic}. Although often discussed as a factual failure, it also has an epistemic dimension: models can be confidently wrong \citep{tian2024factuality,kalai2025language}. In practice, safety improves not only by reducing incorrect answers, but also by ensuring that incorrect answers are accompanied by uncertainty signals that support abstention, fallback, escalation, or filtering. From this perspective, systems benefit not only when errors become rarer, but also when they become more detectable from model behavior \citep{wang2024factuality}.

At the same time, LLMs are routinely adapted through both parameter-efficient fine-tuning (PEFT), including LoRA~\citep{hu2022lora}, DoRA~\citep{liu2024dora}, and PiSSA~\citep{zhu2024pissa}, and full-parameter supervised fine-tuning (SFT). Fine tuning is typically evaluated through downstream accuracy or overall hallucination rate, leaving its effect on answer-level error detectability underexplored. This gap is important because widely used detectors rely on behavioral signals, such as entropy, self-consistency, and semantic agreement, that fine tuning may reshape even when answer accuracy changes little.

Fact-seeking open-ended QA benchmarks provide a natural setting for studying this question because they combine (i) generation freedom, (ii) factual grounding, and (iii) retrieval of knowledge from model parameters under open-ended generation. At the same time, this setting allows externally verifiable answer correctness, making it possible to distinguish ground-truth answer errors from the detector scores used to identify them. In this paper, we focus on answer-level hallucination detection in fact-seeking QA. 

Accordingly, we ask how fine-tuning changes the uncertainty signals that make externally verified incorrect answers more or less detectable in this controlled setting. We characterize the phenomenon primarily through three PEFT methods across three open-weight backbones (LLaMA-3.2-3B-Instruct, Qwen2.5-3B-Instruct, and Mistral-7B-Instruct-v0.3), three QA benchmarks (TriviaQA, NQ-Open, and SQuAD), and seven black-box detectors. We then use one matched full-parameter SFT configuration for each backbone--dataset pair as a specificity check: it tests whether the qualitative black-box pattern is unique to parameter-efficient adaptation, but it is not a hyperparameter sweep or a tuned comparison between PEFT and SFT. The output-format and white-box analyses remain scoped to PEFT. 

Our study yields three main conclusions:
\begin{itemize}[leftmargin=*]
    \item \textbf{Conclusion \#1: Accuracy changes are modest.} Fine-tuning yields only marginal gains in QA accuracy, suggesting limited direct reduction of answer errors through improved factual correctness alone.
    
    \item \textbf{Conclusion \#2: Incorrect answers become more detectable under semantic-consistency and confidence-based detectors.} Despite modest accuracy improvements, fine-tuning consistently improves the separability of externally verified incorrect answers from correct ones for semantic-consistency-based and confidence-based methods. A few-shot format control indicates that this gain is driven mainly by a reduction in output length diversity. 
    
    \item \textbf{Conclusion \#3: White-box probing results are mixed.} Building on the black-box results, we further examine parameter-efficient fine-tuning through white-box probing. PEFT induces structured shifts in hidden representations related to uncertainty, but it does not uniformly increase the linear separability of correctness labels. Supervised white-box linear-probe performance is less consistent than the improvements observed for black-box detector signals.
\end{itemize}

Overall, our findings suggest that, in fact-seeking QA, in-domain task adaptation primarily reduces the output-format diversity that semantic-consistency and confidence-based detectors rely on. And this mechanism makes the uncertainty easier to detect in QA tasks. To our knowledge, this is the first unified study to explore the fine-tuning from this perspective by jointly evaluating external answer-level detectors and internal representation probes in a controlled QA setting.

\section{Related Work}
\label{sec:related}

\vspace{-2mm}
\textbf{Uncertainty Estimation:} 
Fact-seeking QA is a standard evaluation setting for answer reliability and hallucination-related errors, with benchmarks such as SQuAD~\citep{rajpurkar2016squad}, TriviaQA~\citep{joshi2017triviaqa}, and Natural Questions~\citep{kwiatkowski2019natural} providing reference answers. A major line of prior work estimates epistemic uncertainty from sample-based consistency and semantic agreement. SelfCheckGPT compares a model's answer against self-generated alternatives \citep{manakul2023selfcheckgpt}, while Semantic Entropy aggregates semantic dispersion among sampled answers \citep{farquhar2024semanticentropy}. Follow-up work further proposes semantic entropy probes that map hidden representations to semantic clusters for cheaper detection \citep{kossen2024semantic}. Complementary logit-based signals adapt maximum softmax probability and predictive entropy from out-of-distribution detection \citep{hendrycks2017baseline,farquhar2024semanticentropy}, although calibration remains challenging in LLMs \citep{kadavath2022language,lin2022truthfulqa}. Despite rapid progress in uncertainty estimation and answer-level detectors, there is still limited systematic understanding of how fine-tuning, and parameter-efficient fine-tuning in particular, changes the uncertainty and consistency signals on which these methods rely.

\noindent \textbf{Hallucination Detection:}
Prior work on hallucination detection can be broadly grouped into two settings. The first studies answer-level hallucination detection in question answering, where the goal is to determine whether a model's answer to a question is correct or incorrect, typically using fact-seeking benchmarks with reference answers. This line includes methods based on self-consistency, semantic agreement, confidence, and entropy signals, which infer answer reliability from the model's own outputs or predictive uncertainty. The second studies claim-level hallucinated fact detection, which focuses on identifying fabricated or unsupported factual claims within generated text, often at the sentence or claim level in long-form generation, summarization, or dialogue. Compared with QA-based answer-level detection, this setting usually requires finer-grained claim verification, often with external evidence or retrieval. Our work belongs to the first category. Specifically, we study answer-level hallucination detection on fact-seeking QA tasks and investigate how fine-tuning affects the detector-relevant uncertainty signals used in this setting. Claim-level verification of individual facts in free-form text lies outside this scope. Our contribution analyzes how fine-tuning changes the detectability of externally verified answer errors under commonly used answer-level detectors.

\noindent \textbf{Parameter-Efficient Fine-Tuning and Knowledge Dynamics:}
Low-rank adaptation adapts LLMs by training a small set of additional parameters while often approaching full fine-tuning performance. We study three widely used low-rank methods, LoRA \citep{hu2022lora}, PiSSA \citep{zhu2024pissa}, and DoRA \citep{liu2024dora}, which share a common low-rank adaptation core but differ in their parameterization. LoRA adds trainable low-rank adapters to frozen weights \citep{hu2022lora}, PiSSA uses SVD-based initialization from the original weights \citep{zhu2024pissa}, and DoRA decomposes weights into magnitude and direction and adapts directions separately \citep{liu2024dora}. Beyond accuracy, fine-tuning reorganizes knowledge access beyond simply injecting facts: narrow supervision may increase related-query hallucinations \citep{gekhman2024does}, and knowledge editing can struggle with global consistency \citep{zhang2024truthx}. Fine-tuning on factuality preferences improves truthfulness \citep{tian2024factuality}. Preference-based adaptation also reshapes the model's own probabilities: RLHF worsens expected calibration error and concentrates probability mass on the most preferred answer, which leaves the model systematically overconfident. Motivated by these observations, we evaluate fine-tuning through the lens of answer-level error detectability and uncertainty signal reshaping, focusing on how fine-tuning changes detector-relevant behavior in fact-seeking QA. Claims about hallucination behavior across all generation settings lie outside this scope.
\setlength{\textfloatsep}{10pt plus 2pt minus 2pt}
\setlength{\floatsep}{8pt plus 2pt minus 2pt}
\setlength{\intextsep}{8pt plus 2pt minus 2pt}

\section{Experiment}
\subsection{Experimental Setup}
\textbf{Backbones and datasets.}
 For each backbone--dataset pair, the base and fine-tuned models are evaluated on the same benchmark split using a shared correctness-labeling and detector pipeline. This matched setup holds the task and evaluation procedure fixed, providing a controlled and well-established setting for assessing how detector behavior changes after fine-tuning. Experimental details appear in Appendix~\ref{sec:exper_details}.

 We consider three open-weight instruct models: LLaMA-3.2-3B-Instruct\cite{grattafiori2024llama}, Qwen2.5-3B-Instruct\cite{yang2025qwen3}, and Mistral-7B-Instruct-v0.3\cite{jiang2023mistral7b}.
For each backbone, we compare the base instruct model against its fine-tuned variants, covering both parameter-efficient fine-tuning (LoRA, PiSSA, DoRA) and full-parameter SFT, all fine-tuned in-domain on each dataset. We evaluate on TriviaQA (\texttt{rc.nocontext})~\citep{joshi2017triviaqa}\footnote{\url{https://huggingface.co/datasets/mandarjoshi/trivia_qa}}, Natural Questions (NQ-Open)~\citep{kwiatkowski2019natural}\footnote{\url{https://github.com/google-research-datasets/natural-questions}}, and SQuAD v1~\citep{rajpurkar2016squad}\footnote{\url{https://huggingface.co/datasets/rajpurkar/squad}}. These fact-seeking QA benchmarks provide short-form, externally verifiable reference answers and are established testbeds for uncertainty quantification and hallucination detection~\citep{farquhar2024semanticentropy,fadeeva2023lm,lin2023generating}. This setup holds the task and evaluation procedure fixed, providing a controlled and well-established setting for assessing how detector behavior changes after fine-tuning. Experimental details appear in Appendix~\ref{sec:exper_details}. 

\textbf{Parameter efficient fine-tuning methods and full-parameter SFT.} We fine-tune all models with the Hugging Face \texttt{peft} library using LoRA, PiSSA, and DoRA, with rank $r=32$, scaling factor $\alpha=64$, dropout $p=0.05$, AdamW at learning rate $2\times 10^{-5}$, global batch size 64, warmup ratio 0.03, and 1 epoch in bfloat16. All three methods target the attention projection layers (\texttt{q\_proj}, \texttt{k\_proj}, \texttt{v\_proj}, \texttt{o\_proj}). Training and evaluation loss curves for all methods and datasets appear in Appendix~\ref{sec:appendix_results} (Figures~\ref{fig:training_loss} and \ref{fig:eval_loss}). We run one full-parameter SFT configuration for each backbone--dataset pair on the same in-domain QA training sets. Black-box detection results appear in Table~\ref{tab:sft_blackbox}, with further discussion in Appendix~\ref{app:sft_baseline}. 

\paragraph{Black-box hallucination detectors.}
We evaluate task accuracy alongside seven black-box hallucination detectors, grouped into three families.
\textbf{Semantic-consistency-based methods} exploit multiple stochastic samples (we draw $K=10$ samples per question at decoding temperature $1.0$; the full sampling, clustering and judging protocol is given in Appendix~\ref{app:protocol_details}): Semantic Entropy (SE)~\citep{farquhar2024semanticentropy} measures entropy over semantically clustered responses, SelfCheckGPT (SC)~\citep{manakul2023selfcheckgpt} counts how many samples support the primary answer, and Degree of Uncertainty (Deg)~\citep{lin2023generating} measures the connectivity between sampled responses.
\textbf{Confidence-based methods} score an answer by the likelihood the model assigns to it. Maximum Sequence Probability (MSP)~\citep{fadeeva2023lm} uses the probability of the most likely generation,
\[
\mathrm{MSP}(\mathbf{y}\mid \mathbf{x},\theta)=1-P(\mathbf{y}\mid \mathbf{x},\theta),
\]
and Perplexity is the length-normalized inverse likelihood,
\[
P(\mathbf{y},\mathbf{x};\theta)=\exp\left\{-\frac{1}{L}\log P(\mathbf{y}\mid \mathbf{x},\theta)\right\}.
\]
\textbf{Entropy-based methods} measure predictive dispersion: Predictive Entropy~\citep{farquhar2024semanticentropy} (also called Monte Carlo sequence entropy) is the negative average log-probability of the response, and Mean Token Entropy~\citep{fadeeva2023lm} averages the per-token entropy across the sequence.

\paragraph{White-box hallucination detector: linear probe.} 
Following \cite{du2024haloscope,kadavath2022language,kossen2024semantic}, we adopt the standard approach of training an external classifier for hallucination prediction using the backbone LLM’s hidden states as input. Concretely, we train logistic-regression models on a labeled training set, where labels indicate response correctness. We split the original validation set into two equal halves, which we use as a new validation set and a held-out test set, respectively. We select the hidden-state layer used as classifier input by maximizing performance on the new validation set.

\paragraph{Evaluation metrics.}
Following \citet{yao2025reasoning}, we adopt an LLM-as-judge protocol, using GPT-5-mini to label each generated response as correct or incorrect. We report AUROC (area under the receiver operating characteristic curve) as the primary detection metric. For SQuAD we additionally emphasize AUPR (area under the precision-recall curve) because of severe class imbalance: as an extractive reading-comprehension task whose answers are explicitly present in the context, SQuAD yields high model accuracy ($>90\%$) and correspondingly low hallucination rates ($<10\%$). Under such imbalance, AUPR assesses detection performance on the minority class (hallucinations) more informatively than AUROC~\citep{davis2006relationship}.
To validate the reliability of this protocol, Appendix~\ref{app:judge_consistency} reports both a human-annotation study and a cross-judge consistency analysis. A panel of five annotators labeling $100$ TriviaQA samples agrees with the GPT-5-mini judge on $99\%$ of cases (Fleiss' $\kappa>0.85$), and Tables~\ref{tab:judge_verdict_distribution} and~\ref{tab:judge_pairwise_agreement} further show highly consistent verdicts across three independent frontier judges, with pairwise agreement above $97\%$.

\paragraph{Variance across random seeds.}
To verify the robustness of the experiment results, we repeat the experiment with three independently sampled random seeds. For the three sample-based semantic-consistency detectors (SE, SC, Deg), Tables~\ref{tab:llama_auroc},~\ref{tab:qwen_auroc} and~\ref{tab:mistral_auroc} report the resulting mean $\pm$ standard deviation on all three backbones, and Table~\ref{tab:squad_aupr} does the same for AUPR on SQuAD.

\subsection{Experimental Results}

\subsubsection{Black-Box Detectors}
\label{sec:blackbox}

\paragraph{\takeawaymark Takeaway~\#1: Fine-tuning produces modest accuracy gains but substantially larger black-box detectability gains in the evaluated settings.}

\begin{figure}[!htbp]
    \centering
    \includegraphics[width=\textwidth]{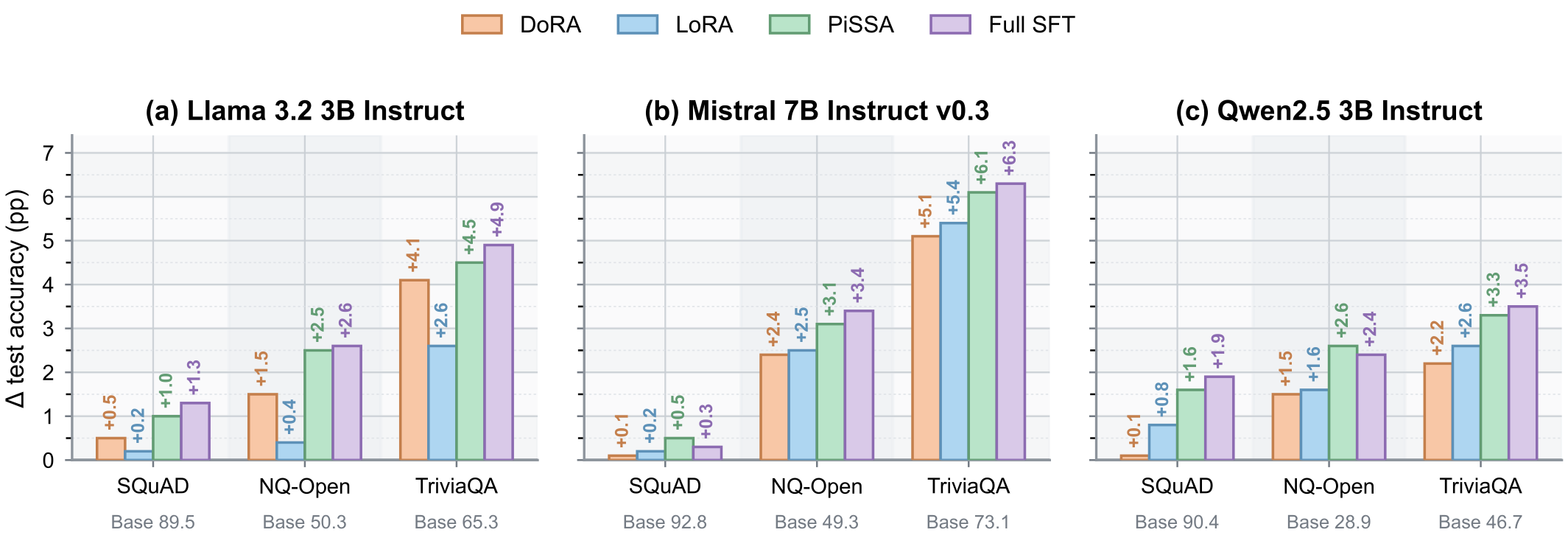}
    \caption{Test accuracy of the Base model and its parameter-efficiently and fully fine-tuned variants across three backbones and three datasets. Bars report absolute accuracy, placing the modest changes after fine-tuning in the context of each Base model's overall performance. In this paper, we define a change as marginal when it is within 1 percentage point.}
    \label{fig:test_accuracy}
\end{figure}

Figure~\ref{fig:test_accuracy} compares absolute test-set accuracy before and after fine-tuning across models and datasets.
In NQ-Open, where base accuracy is overall low, LoRA, PiSSA, and DoRA lift performance by only 0.4\% to 3.1\%, around 1.5\% hallucination mitigation on average. In SQuAD, base models already approach the performance ceiling ($\approx$90\%), and parameter-efficient fine-tuning yields negligible gains of 0.1\% to 1.6\%.  In TriviaQA, the parameter-efficient improvements range from 2.6\% to 6.1\%, about 3\% hallucination mitigation on average. Among the parameter-efficient methods, PiSSA tends to achieve the largest gains across all configurations. We attribute PiSSA’s stronger performance to its initialization scheme. PiSSA applies SVD to the pretrained weight matrix $W_0$ and initializes the adapter from its principal singular directions while keeping the residual components frozen, so adaptation begins in an informative subspace aligned with the dominant directions of $W_0$, where LoRA begins from random low-rank directions~\citep{zhu2024pissa}. This alignment lets the early gradients carry more task-relevant signal, which is consistent with the faster convergence and lower final loss reported for PiSSA. By contrast, DoRA decouples magnitude and direction updates but still initializes its directional component randomly, and therefore does not address this initialization issue. The full-parameter SFT reference in Figure~\ref{fig:test_accuracy} does not change this picture. It stays within $0.4$ percentage points of the best parameter-efficient method in every backbone--dataset pair, and it is not uniformly the strongest: PiSSA remains ahead on Mistral-7B/SQuAD and on Qwen2.5-3B/NQ-Open. Every one of these differences is marginal by the criterion stated in the caption. Accuracy thus stays modest under full-parameter adaptation as well, which sharpens Takeaway~\#1: the detector gains reported in Tables~\ref{tab:llama_auroc},~\ref{tab:qwen_auroc} and~\ref{tab:mistral_auroc} are an order of magnitude larger than the accuracy movement that accompanies them, under both parameter-efficient and full-parameter adaptation.

\paragraph{\takeawaymark Takeaway~\#2: Semantic-consistency and confidence-based hallucination detectors improve significantly, but entropy-based detectors improve only marginally after fine-tuning.}
\label{sec:takeaway2} Tables~\ref{tab:llama_auroc},~\ref{tab:qwen_auroc}, and~\ref{tab:mistral_auroc} characterize this pattern for PEFT. Semantic-consistency and confidence-based detectors improve most consistently, whereas token-level entropy detectors show weaker or mixed changes. This family-level heterogeneity, rather than a gain shared by every detector, is the black-box pattern examined throughout the paper.

\begin{table}[!htbp]
\centering
\small
\caption{AUROC scores for hallucination detection on LLaMA-3.2-3B-Instruct. For the three sample-based semantic-consistency detectors (SE, SC, Deg) we report the mean $\pm$ standard deviation over three random seeds; the confidence and entropy detectors (MSP, Perplexity, Mean Ent., Pre Ent.) use a single representative seed. The highest value in each column, including Base, is bolded. \textbf{Avg Impr.} (green) is the mean improvement of DoRA, LoRA and PiSSA over Base. Full-parameter SFT is reported separately in Table~\ref{tab:sft_blackbox}.}
\renewcommand{\arraystretch}{1.2}
\resizebox{\textwidth}{!}{%
\begin{tabular}{llccccccc}
\toprule
& & \multicolumn{3}{c}{\textbf{Semantic Consistency}} & \multicolumn{2}{c}{\textbf{Confidence}} & \multicolumn{2}{c}{\textbf{Entropy}} \\
\cmidrule(lr){3-5} \cmidrule(lr){6-7} \cmidrule(lr){8-9}
\textbf{Dataset} & \textbf{Method} & \textbf{SE} & \textbf{SC} & \textbf{Deg} & \textbf{MSP} & \textbf{Perplexity} & \textbf{Mean Ent.} & \textbf{Pre Ent.} \\ 
\midrule
\multirow{5}{*}{NQ-Open}
& \cellcolor{graybg}Base & \cellcolor{graybg}$0.7006\pm0.0019$ & \cellcolor{graybg}$0.7423\pm0.0013$ & \cellcolor{graybg}$0.7373\pm0.0027$ & \cellcolor{graybg}0.6977 & \cellcolor{graybg}0.7263 & \cellcolor{graybg}0.7295 & \cellcolor{graybg}0.6649 \\
& DoRA & $0.7403\pm0.0071$ & $0.7611\pm0.0016$ & $0.7387\pm0.0011$ & 0.7585 & 0.7445 & 0.7163 & 0.7030 \\
& LoRA & $0.7408\pm0.0060$ & $0.7517\pm0.0030$ & $0.7379\pm0.0052$ & 0.7620 & 0.7507 & 0.7233 & 0.7058 \\
& PiSSA & $\mathbf{0.7589\pm0.0583}$ & $\mathbf{0.7740\pm0.0235}$ & $\mathbf{0.7602\pm0.0641}$ & \textbf{0.7773} & \textbf{0.7648} & \textbf{0.7381} & \textbf{0.7309} \\
& \cellcolor{avegreen}$\blacktriangle$Avg Impr. & \cellcolor{avegreen}+4.61\% & \cellcolor{avegreen}+2.00\% & \cellcolor{avegreen}+0.83\% & \cellcolor{avegreen}+6.82\% & \cellcolor{avegreen}+2.70\% & \cellcolor{avegreen}-0.36\% & \cellcolor{avegreen}+4.83\% \\
\midrule

\multirow{5}{*}{TriviaQA}
& \cellcolor{graybg}Base & \cellcolor{graybg}$0.8149\pm0.0031$ & \cellcolor{graybg}$0.8585\pm0.0101$ & \cellcolor{graybg}$0.8625\pm0.0052$ & \cellcolor{graybg}0.8154 & \cellcolor{graybg}0.8328 & \cellcolor{graybg}\textbf{0.8457} & \cellcolor{graybg}0.7819 \\
& DoRA & $0.8685\pm0.0196$ & $0.8724\pm0.0161$ & $0.8620\pm0.0167$ & 0.8866 & \textbf{0.8605} & 0.8202 & 0.8138 \\
& LoRA & $0.8534\pm0.0287$ & $0.8734\pm0.0219$ & $0.8575\pm0.0288$ & 0.8656 & 0.8597 & 0.8303 & \textbf{0.8216} \\
& PiSSA & $\mathbf{0.8693\pm0.0147}$ & $\mathbf{0.8859\pm0.0052}$ & $\mathbf{0.8687\pm0.0120}$ & \textbf{0.8878} & 0.8595 & 0.8275 & 0.8087 \\
& \cellcolor{avegreen}$\blacktriangle$Avg Impr. & \cellcolor{avegreen}+4.88\% & \cellcolor{avegreen}+1.87\% & \cellcolor{avegreen}+0.02\% & \cellcolor{avegreen}+6.46\% & \cellcolor{avegreen}+2.71\% & \cellcolor{avegreen}-1.97\% & \cellcolor{avegreen}+3.28\% \\
\midrule

\multirow{5}{*}{SQuAD}
& \cellcolor{gray!20}Base & \cellcolor{gray!20}$0.7147\pm0.0014$ & \cellcolor{gray!20}$0.6766\pm0.0205$ & \cellcolor{gray!20}$0.7131\pm0.0325$ & \cellcolor{gray!20}0.6972 & \cellcolor{gray!20}0.6429 & \cellcolor{gray!20}0.6381 & \cellcolor{gray!20}\textbf{0.6808} \\
& DoRA & $0.7884\pm0.0153$ & $\mathbf{0.7667\pm0.0068}$ & $0.7381\pm0.0821$ & 0.6974 & 0.6901 & 0.6923 & 0.6336 \\
& LoRA & $\mathbf{0.8052\pm0.0035}$ & $0.7381\pm0.0355$ & $\mathbf{0.7758\pm0.0630}$ & \textbf{0.7346} & \textbf{0.7074} & \textbf{0.7006} & 0.6476 \\
& PiSSA & $0.7278\pm0.0703$ & $0.6911\pm0.0277$ & $0.6692\pm0.0230$ & 0.6988 & 0.6834 & 0.6863 & 0.6107 \\
& \cellcolor{avegreen}$\blacktriangle$Avg Impr. & \cellcolor{avegreen}+5.91\% & \cellcolor{avegreen}+5.54\% & \cellcolor{avegreen}+1.46\% & \cellcolor{avegreen}+1.31\% & \cellcolor{avegreen}+5.07\% & \cellcolor{avegreen}+5.50\% & \cellcolor{avegreen}-5.02\% \\
\bottomrule
\end{tabular}}
\label{tab:llama_auroc}
\end{table}

\begin{table}[!htbp]
\centering
\small
\caption{AUROC scores for hallucination detection on Qwen2.5-3B-Instruct. Reporting protocol, bolding and \textbf{Avg Impr.} follow Table~\ref{tab:llama_auroc}.}
\renewcommand{\arraystretch}{1.2}
\resizebox{\textwidth}{!}{%
\begin{tabular}{llccccccc}
\toprule
& & \multicolumn{3}{c}{\textbf{Semantic Consistency}} & \multicolumn{2}{c}{\textbf{Confidence}} & \multicolumn{2}{c}{\textbf{Entropy}} \\
\cmidrule(lr){3-5} \cmidrule(lr){6-7} \cmidrule(lr){8-9}
\textbf{Dataset} & \textbf{Method} & \textbf{SE} & \textbf{SC} & \textbf{Deg} & \textbf{MSP} & \textbf{Perplexity} & \textbf{Mean Ent.} & \textbf{Pre Ent.} \\ 
\midrule

\multirow{5}{*}{NQ-Open}
& \cellcolor{graybg}Base & \cellcolor{graybg}$0.7666\pm0.0021$ & \cellcolor{graybg}$0.7761\pm0.0034$ & \cellcolor{graybg}$0.7912\pm0.0029$ & \cellcolor{graybg}0.7193 & \cellcolor{graybg}0.7173 & \cellcolor{graybg}\textbf{0.7225} & \cellcolor{graybg}0.6920 \\
& DoRA & $\mathbf{0.7780\pm0.0088}$ & $\mathbf{0.8233\pm0.0072}$ & $0.7990\pm0.0106$ & 0.7883 & 0.7264 & 0.6843 & 0.7018 \\
& LoRA & $0.7723\pm0.0095$ & $0.8189\pm0.0058$ & $0.7936\pm0.0121$ & 0.7817 & 0.7219 & 0.6800 & 0.7041 \\
& PiSSA & $0.7765\pm0.0247$ & $0.8178\pm0.0163$ & $\mathbf{0.7995\pm0.0288}$ & \textbf{0.7946} & \textbf{0.7297} & 0.6874 & \textbf{0.7116} \\
& \cellcolor{avegreen}$\blacktriangle$Avg Impr. & \cellcolor{avegreen}+0.90\% & \cellcolor{avegreen}+4.39\% & \cellcolor{avegreen}+0.62\% & \cellcolor{avegreen}+6.89\% & \cellcolor{avegreen}+0.87\% & \cellcolor{avegreen}-3.86\% & \cellcolor{avegreen}+1.38\% \\
% \addlinespace
\midrule

\multirow{5}{*}{TriviaQA}
& \cellcolor{graybg}Base & \cellcolor{graybg}$0.8318\pm0.0038$ & \cellcolor{graybg}$0.8551\pm0.0062$ & \cellcolor{graybg}$0.8548\pm0.0044$& \cellcolor{graybg}0.8132 & \cellcolor{graybg}0.8155 & \cellcolor{graybg}\textbf{0.8320} & \cellcolor{graybg}0.7808 \\
& DoRA & $\mathbf{0.8707\pm0.0113}$ & $0.8756\pm0.0084$ & $0.8642\pm0.0128$ & 0.8938 & 0.8508 & 0.8076 & 0.7899 \\
& LoRA & $0.8694\pm0.0146$ & $0.8796\pm0.0117$ & $0.8647\pm0.0159$ & 0.8971 & \textbf{0.8549} & 0.8145 & \textbf{0.7918} \\
& PiSSA & $0.8692\pm0.0192$ & $\mathbf{0.8828\pm0.0071}$ & $\mathbf{0.8710\pm0.0135}$ & \textbf{0.8973} & 0.8497 & 0.8114 & 0.7898 \\
& \cellcolor{avegreen}$\blacktriangle$Avg Impr. & \cellcolor{avegreen}+3.80\% & \cellcolor{avegreen}+2.42\% & \cellcolor{avegreen}+1.18\% & \cellcolor{avegreen}+8.29\% & \cellcolor{avegreen}+3.63\% & \cellcolor{avegreen}-2.08\% & \cellcolor{avegreen}+0.97\% \\
% \addlinespace
\midrule

\multirow{5}{*}{SQuAD}
& \cellcolor{graybg}Base & \cellcolor{graybg}$0.6669\pm0.0026$ & \cellcolor{graybg}$0.6109\pm0.0187$ & \cellcolor{graybg}$0.6599\pm0.0243$ & \cellcolor{graybg}0.6683 & \cellcolor{graybg}0.6519 & \cellcolor{graybg}0.6558 & \cellcolor{graybg}0.6314 \\
& DoRA & $0.7629\pm0.0208$ & $0.7412\pm0.0141$ & $0.7782\pm0.0396$ & 0.7552 & 0.7303 & 0.7313 & 0.6266 \\
& LoRA & $0.7533\pm0.0164$ & $0.7566\pm0.0273$ & $\mathbf{0.7810\pm0.0451}$ & 0.7436 & 0.7295 & 0.7410 & \textbf{0.6488} \\
& PiSSA & $\mathbf{0.7959\pm0.0512}$ & $\mathbf{0.7608\pm0.0329}$ & $0.7728\pm0.0287$ & \textbf{0.7596} & \textbf{0.7460} & \textbf{0.7570} & 0.6246 \\
& \cellcolor{avegreen}$\blacktriangle$Avg Impr. & \cellcolor{avegreen}+10.38\% & \cellcolor{avegreen}+14.20\% & \cellcolor{avegreen}+11.74\% & \cellcolor{avegreen}+8.45\% & \cellcolor{avegreen}+8.34\% & \cellcolor{avegreen}+8.73\% & \cellcolor{avegreen}+0.19\% \\
\bottomrule
\end{tabular}%
}
\label{tab:qwen_auroc}
\end{table}

\begin{table}[!htbp]
\centering
\small
\caption{AUROC scores for hallucination detection on Mistral-7B-Instruct-v0.3. Reporting protocol, bolding and \textbf{Avg Impr.} follow Table~\ref{tab:llama_auroc}.}
\renewcommand{\arraystretch}{1.2}
\resizebox{\textwidth}{!}{%
\begin{tabular}{llccccccc}
\toprule
& & \multicolumn{3}{c}{\textbf{Semantic Consistency}} & \multicolumn{2}{c}{\textbf{Confidence}} & \multicolumn{2}{c}{\textbf{Entropy}} \\
\cmidrule(lr){3-5} \cmidrule(lr){6-7} \cmidrule(lr){8-9}
\textbf{Dataset} & \textbf{Method} & \textbf{SE} & \textbf{SC} & \textbf{Deg} & \textbf{MSP} & \textbf{Perplexity} & \textbf{Mean Ent.} & \textbf{Pre Ent.} \\
\midrule
\multirow{5}{*}{NQ-Open}
& \cellcolor{gray!20}Base & \cellcolor{gray!20}$0.7047\pm0.0024$ & \cellcolor{gray!20}$0.7357\pm0.0041$ & \cellcolor{gray!20}$0.7578\pm0.0033$ & \cellcolor{gray!20}0.6791 & \cellcolor{gray!20}0.6499 & \cellcolor{gray!20}0.6680 & \cellcolor{gray!20}0.6658 \\
& DoRA & $0.7690\pm0.0102$ & $0.7875\pm0.0079$ & $0.7730\pm0.0143$ & 0.7826 & 0.7271 & 0.6953 & 0.7123 \\
& LoRA & $0.7684\pm0.0087$ & $0.7922\pm0.0064$ & $\mathbf{0.7769\pm0.0118}$ & \textbf{0.7894} & \textbf{0.7368} & \textbf{0.7058} & \textbf{0.7267} \\
& PiSSA & $\mathbf{0.7712\pm0.0316}$ & $\mathbf{0.7937\pm0.0198}$ & $0.7746\pm0.0374$ & 0.7850 & 0.7318 & 0.7038 & 0.7193 \\
& \cellcolor{avegreen}$\blacktriangle$Avg Impr. & \cellcolor{avegreen}+6.48\% & \cellcolor{avegreen}+5.54\% & \cellcolor{avegreen}+1.70\% & \cellcolor{avegreen}+10.66\% & \cellcolor{avegreen}+8.20\% & \cellcolor{avegreen}+3.36\% & \cellcolor{avegreen}+5.36\% \\
\addlinespace
\multirow{5}{*}{TriviaQA}
& \cellcolor{gray!20}Base & \cellcolor{gray!20}$0.7869\pm0.0045$ & \cellcolor{gray!20}$0.7966\pm0.0088$ & \cellcolor{gray!20}$0.8243\pm0.0057$ & \cellcolor{gray!20}0.7571 & \cellcolor{gray!20}0.7185 & \cellcolor{gray!20}0.7396 & \cellcolor{gray!20}0.7448 \\
& DoRA & $\mathbf{0.8891\pm0.0127}$ & $\mathbf{0.9017\pm0.0093}$ & $\mathbf{0.8992\pm0.0146}$ & \textbf{0.8730} & \textbf{0.8229} & 0.7815 & 0.7972 \\
& LoRA & $0.8790\pm0.0168$ & $0.8999\pm0.0134$ & $0.8966\pm0.0201$ & 0.8694 & 0.8219 & \textbf{0.7824} & 0.7953 \\
& PiSSA & $0.8790\pm0.0221$ & $0.9008\pm0.0086$ & $0.8957\pm0.0157$ & 0.8644 & 0.8101 & 0.7703 & \textbf{0.8009} \\
& \cellcolor{avegreen}$\blacktriangle$Avg Impr. & \cellcolor{avegreen}+9.55\% & \cellcolor{avegreen}+10.42\% & \cellcolor{avegreen}+7.29\% & \cellcolor{avegreen}+11.18\% & \cellcolor{avegreen}+9.98\% & \cellcolor{avegreen}+3.85\% & \cellcolor{avegreen}+5.30\% \\
\addlinespace
\multirow{5}{*}{SQuAD}
& \cellcolor{gray!20}Base & \cellcolor{gray!20}$0.6833\pm0.0031$ & \cellcolor{gray!20}$0.6121\pm0.0216$ & \cellcolor{gray!20}$0.6472\pm0.0298$ & \cellcolor{gray!20}0.7032 & \cellcolor{gray!20}0.6547 & \cellcolor{gray!20}0.6715 & \cellcolor{gray!20}\textbf{0.6704} \\
& DoRA & $\mathbf{0.7614\pm0.0187}$ & $\mathbf{0.7407\pm0.0124}$ & $0.7560\pm0.0433$ & \textbf{0.7350} & 0.7234 & 0.7322 & 0.6445 \\
& LoRA & $0.7440\pm0.0152$ & $0.7328\pm0.0301$ & $\mathbf{0.7638\pm0.0578}$ & 0.7247 & \textbf{0.7315} & \textbf{0.7503} & 0.6611 \\
& PiSSA & $0.7596\pm0.0641$ & $0.6858\pm0.0342$ & $0.7282\pm0.0264$ & 0.7348 & 0.7275 & 0.7403 & 0.6330 \\
& \cellcolor{avegreen}$\blacktriangle$Avg Impr. & \cellcolor{avegreen}+7.17\% & \cellcolor{avegreen}+10.77\% & \cellcolor{avegreen}+10.21\% & \cellcolor{avegreen}+2.83\% & \cellcolor{avegreen}+7.28\% & \cellcolor{avegreen}+6.94\% & \cellcolor{avegreen}-2.42\% \\
\bottomrule
\end{tabular}%
}
\label{tab:mistral_auroc}
\end{table}

Comparing the based and PEFT models across all datasets, PEFT consistently improves semantic consistency based and confidence based methods across all backbones and datasets. For the average improvement of AUROC, the semantic-consistency-based and confidence-based methods systematically raise by ranging from 0.02\% (TriviaQA with degree of uncertainty) to 6.82\% (NQ-Open with MSP) on LLaMA-3.2-3B. And for Qwen-2.5-3B, the performance has been uplifted by from 0.62\% (NQ-Open with degree of uncertainty) to 14.20\% (SQuAD with SelfCheckGPT). As for Mistral-7B, the scores are increased by 1.70\% (NQ-Open with degree of uncertainty) to 11.18\% (TriviaQA with MSP).

In contrast to semantic consistency-based and confidence-based hallucination detectors, entropy-based detectors exhibit inconsistent performance. Based on our analysis, this is related to reduced output diversity, detailed in \ref{sec:why_detection}

\begin{table}[!htbp]
\centering
\caption{AUPR scores for hallucination detection on SQuAD across three backbones. Reporting protocol, bolding and \textbf{Avg Impr.} follow Table~\ref{tab:llama_auroc}.}
\resizebox{\textwidth}{!}{%
\renewcommand{\arraystretch}{1.25}
\begin{tabular}{l l c c c c c c c}
\toprule
 &  & \multicolumn{3}{c}{\textbf{Semantic Consistency}} & \multicolumn{2}{c}{\textbf{Confidence}} & \multicolumn{2}{c}{\textbf{Entropy}} \\
\cmidrule(lr){3-5} \cmidrule(lr){6-7} \cmidrule(lr){8-9}
\textbf{Model} & \textbf{Method} & \textbf{SE} & \textbf{SC} & \textbf{Deg} & \textbf{MSP} & \textbf{Perplexity} & \textbf{Mean Ent.} & \textbf{Pre Ent.} \\
\midrule
\multirow{5}{*}{\textbf{LLaMA-3.2-3B-Instruct}}
  & \cellcolor{gray!20}Base & \cellcolor{gray!20}$0.2456\pm0.0044$ & \cellcolor{gray!20}$0.2110\pm0.0065$ & \cellcolor{gray!20}$0.1601\pm0.0058$ & \cellcolor{gray!20}0.1952 & \cellcolor{gray!20}0.1849 & \cellcolor{gray!20}0.1845 & \cellcolor{gray!20}\textbf{0.2125} \\
  & DoRA & $0.2951\pm0.0183$ &  $\mathbf{0.3433\pm0.0148}$ & $0.1961\pm0.0186$ & 0.2219 & 0.2044 & 0.2092 & 0.1873 \\
  & LoRA & $\mathbf{0.3400\pm0.0174}$ & $0.3312\pm0.0252$ & $\mathbf{0.2773\pm0.0241}$ &  \textbf{0.3101} & \textbf{0.2652} & \textbf{0.2754} & 0.1751 \\
  & PiSSA & $0.2640\pm0.0393$ & $0.2340\pm0.0232$ & $0.1853\pm0.0139$ & 0.2186 & 0.1951 & 0.2055 & 0.1242 \\
& \cellcolor{avegreen}$\blacktriangle$Avg Impr. & \cellcolor{avegreen}+5.41\% & \cellcolor{avegreen}+9.18\% & \cellcolor{avegreen}+5.95\% & \cellcolor{avegreen}+5.50\% & \cellcolor{avegreen}+3.67\% & \cellcolor{avegreen}+4.55\% & \cellcolor{avegreen}-5.03\% \\
\midrule
\multirow{5}{*}{\textbf{Mistral-7B-Instruct-v0.3}}
  & \cellcolor{gray!20}Base & \cellcolor{gray!20}$0.1875\pm0.0045$ & \cellcolor{gray!20}$0.1288\pm0.0053$ & \cellcolor{gray!20}$0.1608\pm0.0072$ & \cellcolor{gray!20}0.1736 & \cellcolor{gray!20}0.1386 & \cellcolor{gray!20}0.1503 & \cellcolor{gray!20} \textbf{0.1478} \\
  & DoRA &  $\mathbf{0.2519\pm0.0171}$ &  $\mathbf{0.2758\pm0.0143}$ &  $\mathbf{0.2773\pm0.0286}$ &  \textbf{0.2618} &  \textbf{0.1983} &  \textbf{0.1992} & 0.1195 \\
  & LoRA & $0.2176\pm0.0161$ & $0.2294\pm0.0204$ & $0.2628\pm0.0331$ & 0.1969 & 0.1747 & 0.1898 & 0.1253 \\
  & PiSSA & $0.2516\pm0.0438$ & $0.2190\pm0.0245$ & $0.2403\pm0.0195$ & 0.2121 & 0.1784 & 0.1933 & 0.1071 \\
& \cellcolor{avegreen}$\blacktriangle$Avg Impr. & \cellcolor{avegreen}+5.29\% & \cellcolor{avegreen}+11.26\% & \cellcolor{avegreen}+9.93\% & \cellcolor{avegreen}+5.00\% & \cellcolor{avegreen}+4.52\% & \cellcolor{avegreen}+4.38\% & \cellcolor{avegreen}-3.05\% \\
\midrule
\multirow{5}{*}{\textbf{Qwen2.5-3B-Instruct}}
  & \cellcolor{gray!20}Base & \cellcolor{gray!20}$0.1787\pm0.0046$ & \cellcolor{gray!20}$0.1767\pm0.0060$ & \cellcolor{gray!20}$0.2167\pm0.0085$ & \cellcolor{gray!20}0.1658 & \cellcolor{gray!20}0.1534 & \cellcolor{gray!20}0.1578 & \cellcolor{gray!20}0.1348 \\
  & DoRA & $0.3021\pm0.0175$ & $0.2659\pm0.0168$ & $0.3172\pm0.0289$ &  \textbf{0.2572} & 0.2528 & 0.2587 & 0.1408 \\
  & LoRA & $0.2635\pm0.0190$ & $0.2665\pm0.0224$ &  $\mathbf{0.3330\pm0.0393}$ & 0.2198 & 0.2603 & 0.2762 &  \textbf{0.1549} \\
  & PiSSA &  $\mathbf{0.3113\pm0.0423}$ &  $\mathbf{0.2952\pm0.0298}$ & $0.3214\pm0.0248$ & 0.2282 &  \textbf{0.2622} &  \textbf{0.2816} & 0.1391 \\
& \cellcolor{avegreen}$\blacktriangle$Avg Impr. & \cellcolor{avegreen}+11.36\% & \cellcolor{avegreen}+9.92\% & \cellcolor{avegreen}+10.72\% & \cellcolor{avegreen}+6.93\% & \cellcolor{avegreen}+10.50\% & \cellcolor{avegreen}+11.44\% & \cellcolor{avegreen}+1.01\% \\
\bottomrule
\end{tabular}%
}
\label{tab:squad_aupr}
\end{table}

Additionally, we report AUPR scores as a more informative metric for SQuAD in Table~\ref{tab:squad_aupr} due to severe class imbalance in SQuAD (high accuracy, low hallucination rate). PEFT also improves hallucination detection performance systematically on SQuAD across all models observed from AUPR consistent with AUROC. These results confirm that PEFT enhances uncertainty detectability even under imbalanced class conditions where AUROC may be misleadingly optimistic. To complement this SQuAD-specific analysis, we also report AUPR results for TriviaQA and NQ-Open in Appendix Tables~\ref{tab:triviaqa_aupr} and~\ref{tab:nqopen_aupr}, which help contextualize why AUROC gains in open-domain QA do not always translate into equally consistent AUPR gains.

\paragraph{From parameter-efficient to full-parameter fine-tuning.}
Takeaway~\#2 establishes the black-box detectability pattern for the three low-rank methods. Low-rank adaptation confines the weight update to a small subspace, which raises a specificity question: does the pattern depend on that constraint, or does it follow from in-domain supervision itself?  Table~\ref{tab:sft_blackbox} compares one matched full-parameter SFT configuration per backbone--dataset pair with the corresponding Base model.

The same qualitative family-level pattern appears across the nine evaluated configurations. Semantic-consistency detectors improve in all nine backbone--dataset pairs: SE gains range from $5.38\%$ to $14.24\%$, SC from $2.49\%$ to $16.36\%$, and Deg from $0.79\%$ to $14.63\%$. MSP also improves in all nine pairs, by $4.53\%$ to $12.81\%$. Token-level entropy is less uniform: Mean Token Entropy ranges from $-3.45\%$ to $+10.62\%$ and is negative in three pairs, while Predictive Entropy is negative in one. Thus, the broad detector-family pattern is not unique to PEFT in these tested settings.

This specificity check does not establish that every full-SFT recipe behaves this way. Because we evaluate only one configuration per backbone--dataset pair, the results do not support full-SFT hyperparameter invariance, a ranking between PEFT and full SFT, or a shared PEFT/SFT mechanism. The output-format and white-box results below remain claims about the PEFT experiments.

\begin{table}[!htbp]
\centering
\small
\caption{Black-box hallucination detection under full-parameter SFT. Each Full SFT cell reports the score followed by its change over Base in percentage points; SE, SC and Deg are averaged over three seeds.}
\label{tab:sft_blackbox}
\renewcommand{\arraystretch}{1.2}
\resizebox{\textwidth}{!}{%
\begin{tabular}{lllllllll}
\toprule
& & \multicolumn{3}{l}{\textbf{Semantic Consistency}} & \multicolumn{2}{l}{\textbf{Confidence}} & \multicolumn{2}{l}{\textbf{Entropy}} \\
\cmidrule(lr){3-5} \cmidrule(lr){6-7} \cmidrule(lr){8-9}
\textbf{Backbone} & \textbf{Method} & \textbf{SE} & \textbf{SC} & \textbf{Deg} & \textbf{MSP} & \textbf{Perplexity} & \textbf{Mean Ent.} & \textbf{Pre Ent.} \\
\midrule
\multicolumn{9}{l}{\cellcolor{gray!15}\textbf{NQ-Open}} \\
\multirow{2}{*}{LLaMA-3.2-3B} & Base & $0.7006\pm0.0019$ & $0.7423\pm0.0013$ & $0.7373\pm0.0027$ & 0.6977 & 0.7263 & 0.7295 & 0.6649 \\
& Full SFT & $0.7746\pm0.0124$~\textcolor{safegreen}{(+7.40)} & $0.7972\pm0.0068$~\textcolor{safegreen}{(+5.49)} & $0.7784\pm0.0093$~\textcolor{safegreen}{(+4.11)} & 0.7972~\textcolor{safegreen}{(+9.95)} & 0.7517~\textcolor{safegreen}{(+2.54)} & 0.7293~\textcolor{dangerred}{(-0.02)} & 0.7515~\textcolor{safegreen}{(+8.66)} \\
\addlinespace[2pt]
\multirow{2}{*}{Qwen2.5-3B} & Base & $0.7666\pm0.0021$ & $0.7761\pm0.0034$ & $0.7912\pm0.0029$ & 0.7193 & 0.7173 & 0.7225 & 0.6920 \\
& Full SFT & $0.8204\pm0.0134$~\textcolor{safegreen}{(+5.38)} & $0.8372\pm0.0097$~\textcolor{safegreen}{(+6.11)} & $0.8225\pm0.0152$~\textcolor{safegreen}{(+3.13)} & 0.8318~\textcolor{safegreen}{(+11.25)} & 0.7841~\textcolor{safegreen}{(+6.68)} & 0.7425~\textcolor{safegreen}{(+2.00)} & 0.7667~\textcolor{safegreen}{(+7.47)} \\
\addlinespace[2pt]
\multirow{2}{*}{Mistral-7B} & Base & $0.7047\pm0.0024$ & $0.7357\pm0.0041$ & $0.7578\pm0.0033$ & 0.6791 & 0.6499 & 0.6680 & 0.6658 \\
& Full SFT & $0.7874\pm0.0141$~\textcolor{safegreen}{(+8.27)} & $0.7942\pm0.0106$~\textcolor{safegreen}{(+5.85)} & $0.7657\pm0.0169$~\textcolor{safegreen}{(+0.79)} & 0.8072~\textcolor{safegreen}{(+12.81)} & 0.7608~\textcolor{safegreen}{(+11.09)} & 0.7263~\textcolor{safegreen}{(+5.83)} & 0.7531~\textcolor{safegreen}{(+8.73)} \\
\midrule
\multicolumn{9}{l}{\cellcolor{gray!15}\textbf{TriviaQA}} \\
\multirow{2}{*}{LLaMA-3.2-3B} & Base & $0.8149\pm0.0031$ & $0.8585\pm0.0101$ & $0.8625\pm0.0052$ & 0.8154 & 0.8328 & 0.8457 & 0.7819 \\
& Full SFT & $0.8872\pm0.0231$~\textcolor{safegreen}{(+7.23)} & $0.8834\pm0.0163$~\textcolor{safegreen}{(+2.49)} & $0.8736\pm0.0208$~\textcolor{safegreen}{(+1.11)} & 0.8824~\textcolor{safegreen}{(+6.70)} & 0.8371~\textcolor{safegreen}{(+0.43)} & 0.8112~\textcolor{dangerred}{(-3.45)} & 0.8024~\textcolor{safegreen}{(+2.05)} \\
\addlinespace[2pt]
\multirow{2}{*}{Qwen2.5-3B} & Base & $0.8318\pm0.0038$ & $0.8551\pm0.0062$ & $0.8548\pm0.0044$ & 0.8132 & 0.8155 & 0.8320 & 0.7808 \\
& Full SFT & $0.8904\pm0.0176$~\textcolor{safegreen}{(+5.86)} & $0.8925\pm0.0129$~\textcolor{safegreen}{(+3.74)} & $0.8778\pm0.0184$~\textcolor{safegreen}{(+2.30)} & 0.9024~\textcolor{safegreen}{(+8.92)} & 0.8566~\textcolor{safegreen}{(+4.11)} & 0.8235~\textcolor{dangerred}{(-0.85)} & 0.7912~\textcolor{safegreen}{(+1.04)} \\
\addlinespace[2pt]
\multirow{2}{*}{Mistral-7B} & Base & $0.7869\pm0.0045$ & $0.7966\pm0.0088$ & $0.8243\pm0.0057$ & 0.7571 & 0.7185 & 0.7396 & 0.7448 \\
& Full SFT & $0.8416\pm0.0193$~\textcolor{safegreen}{(+5.47)} & $0.8463\pm0.0148$~\textcolor{safegreen}{(+4.97)} & $0.8465\pm0.0212$~\textcolor{safegreen}{(+2.22)} & 0.8328~\textcolor{safegreen}{(+7.57)} & 0.7741~\textcolor{safegreen}{(+5.56)} & 0.7433~\textcolor{safegreen}{(+0.37)} & 0.7817~\textcolor{safegreen}{(+3.69)} \\
\midrule
\multicolumn{9}{l}{\cellcolor{gray!15}\textbf{SQuAD}} \\
\multirow{2}{*}{LLaMA-3.2-3B} & Base & $0.7147\pm0.0014$ & $0.6766\pm0.0205$ & $0.7131\pm0.0325$ & 0.6972 & 0.6429 & 0.6381 & 0.6808 \\
& Full SFT & $0.7804\pm0.0287$~\textcolor{safegreen}{(+6.57)} & $0.7472\pm0.0246$~\textcolor{safegreen}{(+7.06)} & $0.7675\pm0.0412$~\textcolor{safegreen}{(+5.44)} & 0.7425~\textcolor{safegreen}{(+4.53)} & 0.7314~\textcolor{safegreen}{(+8.85)} & 0.7443~\textcolor{safegreen}{(+10.62)} & 0.6701~\textcolor{dangerred}{(-1.07)} \\
\addlinespace[2pt]
\multirow{2}{*}{Qwen2.5-3B} & Base & $0.6669\pm0.0026$ & $0.6109\pm0.0187$ & $0.6599\pm0.0243$ & 0.6683 & 0.6519 & 0.6558 & 0.6314 \\
& Full SFT & $0.8093\pm0.0294$~\textcolor{safegreen}{(+14.24)} & $0.7745\pm0.0238$~\textcolor{safegreen}{(+16.36)} & $0.8062\pm0.0367$~\textcolor{safegreen}{(+14.63)} & 0.7417~\textcolor{safegreen}{(+7.34)} & 0.7383~\textcolor{safegreen}{(+8.64)} & 0.7427~\textcolor{safegreen}{(+8.69)} & 0.6635~\textcolor{safegreen}{(+3.21)} \\
\addlinespace[2pt]
\multirow{2}{*}{Mistral-7B} & Base & $0.6833\pm0.0031$ & $0.6121\pm0.0216$ & $0.6472\pm0.0298$ & 0.7032 & 0.6547 & 0.6715 & 0.6704 \\
& Full SFT & $0.7645\pm0.0276$~\textcolor{safegreen}{(+8.12)} & $0.7273\pm0.0229$~\textcolor{safegreen}{(+11.52)} & $0.7588\pm0.0389$~\textcolor{safegreen}{(+11.16)} & 0.7712~\textcolor{safegreen}{(+6.80)} & 0.7549~\textcolor{safegreen}{(+10.02)} & 0.7634~\textcolor{safegreen}{(+9.19)} & 0.6772~\textcolor{safegreen}{(+0.68)} \\
\bottomrule
\end{tabular}}
\end{table}

\subsubsection{White-Box Probes}

While the unsupervised black-box detectors (semantic consistency based and confidence based) benefit from parameter-efficient fine-tuning, it remains unclear whether this improvement extends to other detection paradigms. We examine linear probing, a supervised white-box approach that directly classifies hidden representations to test whether parameter-efficient fine-tuning universally enhances hallucination detectability.

\begin{table}[t]
\centering
\footnotesize
\captionsetup{skip=5pt,width=0.78\textwidth}
\caption{Best-layer linear probe results for hallucination detection. Bold values indicate the best AUROC or AUPR within each model--dataset block.}
\label{tab:best_probe_auroc_aupr}
\setlength{\tabcolsep}{3.8pt}
\renewcommand{\arraystretch}{1.06}
\begin{tabular}{@{}>{\raggedright\arraybackslash}p{0.16\textwidth}>{\raggedright\arraybackslash}p{0.105\textwidth}>{\raggedright\arraybackslash}p{0.10\textwidth}>{\centering\arraybackslash}p{0.085\textwidth}>{\centering\arraybackslash}p{0.10\textwidth}>{\centering\arraybackslash}p{0.10\textwidth}@{}}
\toprule
\textbf{Model} & \textbf{Dataset} & \textbf{Method} & \shortstack[c]{\textbf{Best}\\\textbf{Layer}} & \textbf{AUROC} & \textbf{AUPR} \\
\midrule
\multirow{12}{=}{\textbf{LLaMA-3.2-3B-Instruct}}
& \cellcolor{gray!12}\textbf{NQ-Open} & \cellcolor{gray!12}\textbf{Base} & \cellcolor{gray!12}13 & \cellcolor{gray!12}\textbf{0.7525} & \cellcolor{gray!12}\textbf{0.7562} \\
& & DoRA & 13 & 0.7062 & 0.6914 \\
& & LoRA & 14 & 0.7219 & 0.7124 \\
& & PiSSA & 14 & 0.7440 & 0.7478 \\
\addlinespace[3pt]
& \cellcolor{gray!12}\textbf{TriviaQA} & \cellcolor{gray!12}\textbf{Base} & \cellcolor{gray!12}18 & \cellcolor{gray!12}0.8089 & \cellcolor{gray!12}0.6823 \\
& & DoRA & 12 & 0.8498 & 0.7398 \\
& & LoRA & 16 & \textbf{0.8644} & \textbf{0.7608} \\
& & PiSSA & 12 & 0.8512 & 0.7523 \\
\addlinespace[3pt]
& \cellcolor{gray!12}\textbf{SQuAD} & \cellcolor{gray!12}\textbf{Base} & \cellcolor{gray!12}23 & \cellcolor{gray!12}0.7097 & \cellcolor{gray!12}0.2842 \\
& & DoRA & 14 & \textbf{0.7788} & \textbf{0.3309} \\
& & LoRA & 22 & 0.7691 & 0.3056 \\
& & PiSSA & 22 & 0.7172 & 0.3029 \\
\midrule
\multirow{12}{=}{\textbf{Mistral-7B-Instruct-v0.3}}
& \cellcolor{gray!12}\textbf{NQ-Open} & \cellcolor{gray!12}\textbf{Base} & \cellcolor{gray!12}15 & \cellcolor{gray!12}\textbf{0.7748} & \cellcolor{gray!12}\textbf{0.7908} \\
& & DoRA & 14 & 0.7262 & 0.7058 \\
& & LoRA & 15 & 0.7279 & 0.7066 \\
& & PiSSA & 15 & 0.7227 & 0.6901 \\
\addlinespace[3pt]
& \cellcolor{gray!12}\textbf{TriviaQA} & \cellcolor{gray!12}\textbf{Base} & \cellcolor{gray!12}14 & \cellcolor{gray!12}0.8667 & \cellcolor{gray!12}0.7031 \\
& & DoRA & 16 & 0.8676 & 0.7176 \\
& & LoRA & 16 & \textbf{0.8678} & \textbf{0.7193} \\
& & PiSSA & 16 & 0.8670 & 0.7035 \\
\addlinespace[3pt]
& \cellcolor{gray!12}\textbf{SQuAD} & \cellcolor{gray!12}\textbf{Base} & \cellcolor{gray!12}13 & \cellcolor{gray!12}\textbf{0.7805} & \cellcolor{gray!12}\textbf{0.3336} \\
& & DoRA & 22 & 0.7604 & 0.2522 \\
& & LoRA & 19 & 0.7453 & 0.2182 \\
& & PiSSA & 18 & 0.7592 & 0.2522 \\
\midrule
\multirow{12}{=}{\textbf{Qwen2.5-3B-Instruct}}
& \cellcolor{gray!12}\textbf{NQ-Open} & \cellcolor{gray!12}\textbf{Base} & \cellcolor{gray!12}24 & \cellcolor{gray!12}\textbf{0.7603} & \cellcolor{gray!12}\textbf{0.8807} \\
& & DoRA & 24 & 0.7515 & 0.8646 \\
& & LoRA & 24 & 0.7268 & 0.8458 \\
& & PiSSA & 24 & 0.7496 & 0.8557 \\
\addlinespace[3pt]
& \cellcolor{gray!12}\textbf{TriviaQA} & \cellcolor{gray!12}\textbf{Base} & \cellcolor{gray!12}27 & \cellcolor{gray!12}0.8564 & \cellcolor{gray!12}0.8683 \\
& & DoRA & 24 & 0.8661 & 0.8698 \\
& & LoRA & 26 & 0.8652 & \textbf{0.8760} \\
& & PiSSA & 27 & \textbf{0.8678} & 0.8707 \\
\addlinespace[3pt]
& \cellcolor{gray!12}\textbf{SQuAD} & \cellcolor{gray!12}\textbf{Base} & \cellcolor{gray!12}26 & \cellcolor{gray!12}0.7356 & \cellcolor{gray!12}\textbf{0.2617} \\
& & DoRA & 24 & 0.7365 & 0.2465 \\
& & LoRA & 24 & 0.7409 & 0.2430 \\
& & PiSSA & 34 & \textbf{0.7456} & 0.2434 \\
\bottomrule
\end{tabular}
\vspace{-0.2em}
\end{table}

\begin{figure}[t!]
\centering
\includegraphics[width=\textwidth]{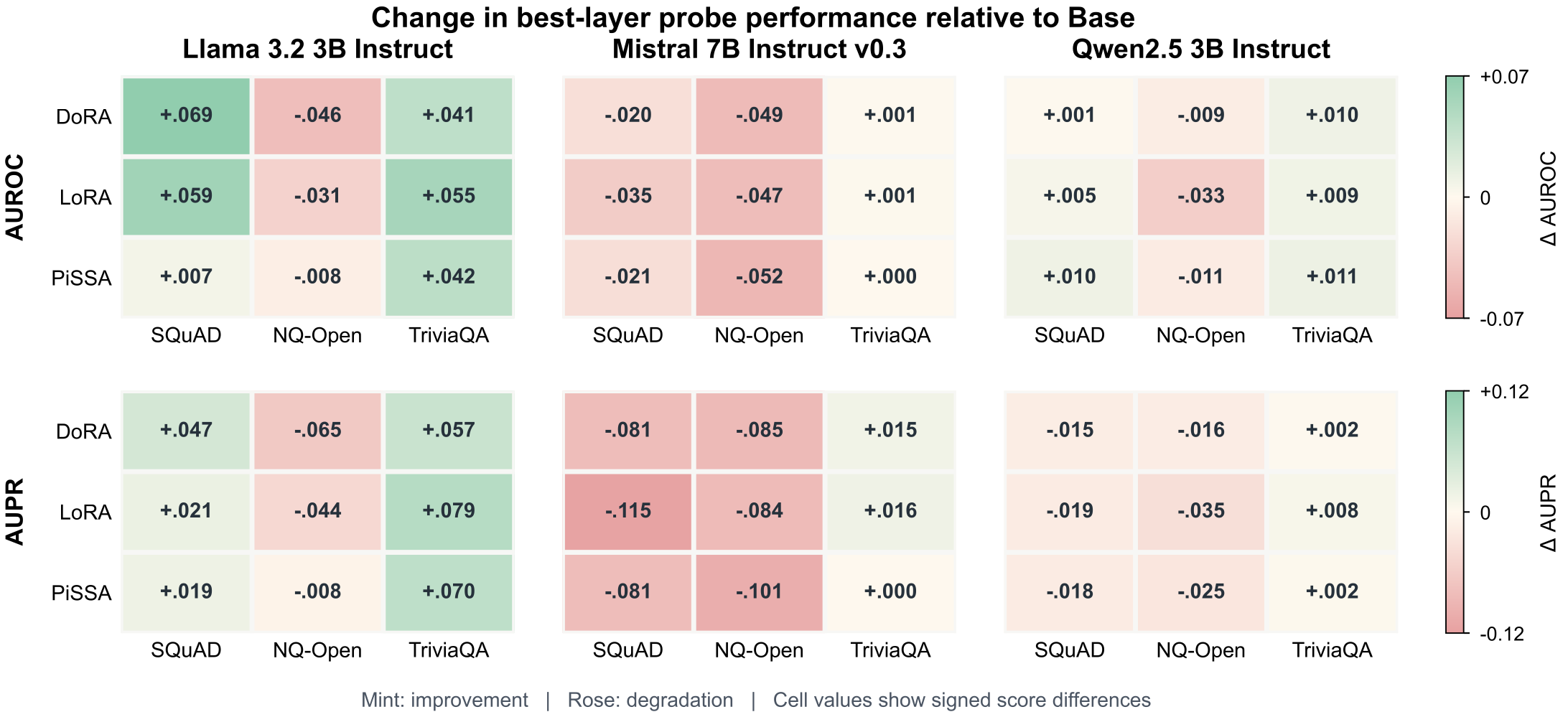}
\caption{%
\textbf{Change in probe hallucination detection scores relative to Base.}
We train logistic regression probes on every layer's hidden states to detect hallucination. Each cell reports the signed difference in the best-performing layer's validation AUROC or AUPR after parameter-efficient fine-tuning; mint indicates improvement and rose indicates degradation. Absolute scores are reported in Table~\ref{tab:best_probe_auroc_aupr}.}
\label{fig:probe_results}
\end{figure}

\paragraph{\takeawaymark Takeaway~\#3: Parameter-efficient fine-tuning raises black-box hallucination detectors but leaves linear-probe detection inconsistent.}
Figure~\ref{fig:probe_results} and Table~\ref{tab:best_probe_auroc_aupr} reveal striking inconsistency: on TriviaQA, parameter-efficient fine-tuning improves probing performance; on NQ-Open, it uniformly degrades; on SQuAD, effects vary by model. This inconsistency contrasts sharply with black-box unsupervised uncertainty detectors (Table~\ref{tab:llama_auroc},~\ref{tab:qwen_auroc} and Table~\ref{tab:squad_aupr}), which improve consistently almost across all settings after parameter-efficient fine-tuning. We stress that this separability claim is quantitative and high-dimensional: the linear probe is a logistic-regression classifier trained on the full hidden-state vector (e.g., $3072$ dimensions for LLaMA-3.2-3B), so Table~\ref{tab:best_probe_auroc_aupr} already measures linear separability in the full representation space, not in a reduced projection. The PCA visualizations in Appendix~Figure~\ref{fig:pca_plots} are consistent with this observation: after parameter-efficient fine-tuning, hidden representations do not exhibit clearer separation between correct and incorrect predictions in the leading components, and in some cases clusters become less distinguishable. Because the top two principal components capture only a small fraction of the total variance, we do not treat these plots as evidence of inseparability on their own; the quantitative claim rests on the full-dimensional probe in Table~\ref{tab:best_probe_auroc_aupr}. Taken together, these results support Takeaway~\#3 by showing that, unlike black-box uncertainty detectors, probe-based methods do not consistently benefit from parameter-efficient fine-tuning and may even degrade, highlighting a representational shift induced by parameter-efficient fine-tuning.

\paragraph{Scope of the full-parameter SFT arm.}
The full-parameter SFT runs in Table~\ref{tab:sft_blackbox} answer one question: whether the black-box detectability gain of Section~\ref{sec:blackbox} depends on the low-rank constraint. A specificity check of this kind applies to a positive effect, and the black-box gain is the effect that raises the question. The probe result above is mixed, so running the same check under full-parameter SFT would ask whether an absent effect is also absent under a different parameterization, which says little about the mechanism. Section~\ref{sec:not_separability} and Section~\ref{sec:behavioral} decompose how the gains arise within parameter-efficient adaptation, and we run and report them for the parameter-efficient models alone. Every claim outside Table~\ref{tab:sft_blackbox} is therefore stated for parameter-efficient fine-tuning. A full-parameter probing and behavioral arm is a direct extension that we leave to future work.

\section{Why Detection Improves}
\label{sec:why_detection}

\subsection{Hidden-State Separability Does Not Explain the Gains}
\label{sec:not_separability}

We read this divergence as informative. The black-box detectors and the linear probe ask different questions: the former measure whether the model's output behavior separates correct from incorrect answers, while the latter measures whether correctness is linearly encoded in the hidden states. That black-box detectability improves while linear separability does not tells us the gains do not come from a stronger linear encoding of correctness in the representations; they instead live at the output-behavior level, consistent with the reduction in output-format diversity identified in Section~\ref{sec:format_control}. This divergence also clarifies which detector family is affected, and it falls on the less deployable one: black-box detectors are unsupervised and require only the model's outputs, whereas linear probes are supervised, require labeled hidden states that are often unavailable in practice, and can degrade under train/test distribution shift. We therefore do not claim that parameter-efficient fine-tuning improves the internal linear encoding of correctness. We instead make the qualified claim that in-domain task adaptation reshapes the output-behavioral signals that black-box detectors rely on, while leaving hidden-state linear separability mixed. Characterizing this representational shift and developing probes robust to parameter-efficient fine-tuning remain promising directions for future work.

\subsection{Output-Format Diversity Does: A Few-Shot Control}
\label{sec:format_control}

\begin{table}[H]
\centering
\small
\caption{Few-shot format control: hallucination-detection AUROC for the LLaMA-3.2-3B base model prompted with short-answer demonstrations (\texttt{base\_fewshot}), against the base model and its weight-adapted variants. Reporting protocol follows Table~\ref{tab:llama_auroc}.}
\label{tab:fewshot-control-llama}
\setlength{\tabcolsep}{4pt}
\renewcommand{\arraystretch}{1.2}
\resizebox{0.98\textwidth}{!}{%
\begin{tabular}{@{}llccccccc@{}}
\toprule
\multicolumn{2}{c}{} & \multicolumn{3}{c}{\textbf{Semantic consistency}} & \textbf{Confidence} & \multicolumn{3}{c}{\textbf{Token-level entropy}} \\
\cmidrule(lr){3-5}\cmidrule(lr){6-6}\cmidrule(lr){7-9}
\textbf{Dataset} & \textbf{Method} & \textbf{SE} & \textbf{SC} & \textbf{Deg} & \textbf{MSP} & \textbf{Perplexity} & \textbf{Mean Ent.} & \textbf{Pre Ent.} \\
\midrule
\multirow{6}{*}{\textbf{NQ-Open}}
& Base          & $0.7006\pm0.0019$ & $0.7423\pm0.0013$ & $0.7373\pm0.0027$ & 0.6977 & 0.7263 & 0.7295 & 0.6649 \\
& \cellcolor{gray!22}\textbf{Base\_fewshot} & \cellcolor{gray!22}$0.6751\pm0.0138$ & \cellcolor{gray!22}$0.7274\pm0.0074$ & \cellcolor{gray!22}$0.7248\pm0.0116$ & \cellcolor{gray!22}0.6256 & \cellcolor{gray!22}0.7020 & \cellcolor{gray!22}0.7008 & \cellcolor{gray!22}0.6249 \\
\cmidrule(lr){2-9}
& DoRA          & $0.7403\pm0.0071$ & $0.7611\pm0.0016$ & $0.7387\pm0.0011$ & 0.7585 & 0.7445 & 0.7163 & 0.7030 \\
& LoRA          & $0.7408\pm0.0060$ & $0.7517\pm0.0030$ & $0.7379\pm0.0052$ & 0.7620 & 0.7507 & 0.7233 & 0.7058 \\
& PiSSA         & $0.7589\pm0.0583$ & $0.7740\pm0.0235$ & $0.7602\pm0.0641$ & 0.7773 & 0.7648 & 0.7381 & 0.7309 \\
& Full SFT      & $0.7746\pm0.0124$ & $0.7972\pm0.0068$ & $0.7784\pm0.0093$ & 0.7972 & 0.7517 & 0.7293 & 0.7515 \\
\midrule
\multirow{6}{*}{\textbf{TriviaQA}}
& Base          & $0.8149\pm0.0031$ & $0.8585\pm0.0101$ & $0.8625\pm0.0052$ & 0.8154 & 0.8328 & 0.8457 & 0.7819 \\
& \cellcolor{gray!22}\textbf{Base\_fewshot} & \cellcolor{gray!22}$0.8067\pm0.0243$ & \cellcolor{gray!22}$0.8520\pm0.0187$ & \cellcolor{gray!22}$0.8412\pm0.0221$ & \cellcolor{gray!22}0.7999 & \cellcolor{gray!22}0.7775 & \cellcolor{gray!22}0.7802 & \cellcolor{gray!22}0.6886 \\
\cmidrule(lr){2-9}
& DoRA          & $0.8685\pm0.0196$ & $0.8724\pm0.0161$ & $0.8620\pm0.0167$ & 0.8866 & 0.8605 & 0.8202 & 0.8138 \\
& LoRA          & $0.8534\pm0.0287$ & $0.8734\pm0.0219$ & $0.8575\pm0.0288$ & 0.8656 & 0.8597 & 0.8303 & 0.8216 \\
& PiSSA         & $0.8693\pm0.0147$ & $0.8859\pm0.0052$ & $0.8687\pm0.0120$ & 0.8878 & 0.8595 & 0.8275 & 0.8087 \\
& Full SFT      & $0.8872\pm0.0231$ & $0.8834\pm0.0163$ & $0.8736\pm0.0208$ & 0.8824 & 0.8371 & 0.8112 & 0.8024 \\
\midrule
\multirow{6}{*}{\textbf{SQuAD}}
& Base          & $0.7147\pm0.0014$ & $0.6766\pm0.0205$ & $0.7131\pm0.0325$ & 0.6972 & 0.6429 & 0.6381 & 0.6808 \\
& \cellcolor{gray!22}\textbf{Base\_fewshot} & \cellcolor{gray!22}$0.7670\pm0.0196$ & \cellcolor{gray!22}$0.7104\pm0.0312$ & \cellcolor{gray!22}$0.7646\pm0.0574$ & \cellcolor{gray!22}0.7436 & \cellcolor{gray!22}0.7455 & \cellcolor{gray!22}0.7696 & \cellcolor{gray!22}0.7401 \\
\cmidrule(lr){2-9}
& DoRA          & $0.7884\pm0.0153$ & $0.7667\pm0.0068$ & $0.7381\pm0.0821$ & 0.6974 & 0.6901 & 0.6923 & 0.6336 \\
& LoRA          & $0.8052\pm0.0035$ & $0.7381\pm0.0355$ & $0.7758\pm0.0630$ & 0.7346 & 0.7074 & 0.7006 & 0.6476 \\
& PiSSA         & $0.7278\pm0.0703$ & $0.6911\pm0.0277$ & $0.6692\pm0.0230$ & 0.6988 & 0.6834 & 0.6863 & 0.6107 \\
& Full SFT      & $0.7804\pm0.0287$ & $0.7472\pm0.0246$ & $0.7675\pm0.0412$ & 0.7425 & 0.7314 & 0.7443 & 0.6701 \\
\bottomrule
\end{tabular}%
}
\end{table}

A natural concern is whether the detector gains are a format artifact or a consequence of the weight update. In-domain adaptation makes the model answer in a concise and short style, which can also be induced through prompting. However, our results show that the improved performance of consistency-based uncertainty scores is associated with the reduction in response diversity after fine-tuning, with response length playing no comparable role. On generative tasks, prompting changes response length and style but does not produce the same diversity reduction or corresponding uncertainty-estimation gains; these effects emerge only after weight-level adaptation.

\paragraph{Length alone is not the only driver.}
Table~\ref{tab:generation_length_statistics} (Appendix~\ref{app:response_length}) confirms that adapted models generate shorter answers. e.g., on NQ-Open the mean length drops from 11.92 to roughly 5.3 tokens. However, the gains are concentrated in the Semantic-consistency-based detectors SE, SC, Deg and the confidence detector MSP, whereas token-level entropy detectors, which would be most directly affected by a length change, improve only marginally. Showing that token-level entropy does not improve only rules out a change in mean length; it does not rule out a change in the signal the multi-sample detectors actually use, namely the diversity of the sampled outputs. We therefore measure diversity directly, by the standard deviation (std) of answer length: a large std means the model answers with varying length, whereas a small std means uniform outputs.

\paragraph{A few-shot control separates format from weights.}
To separate the short-answer format from the weight update, we prompt the LLaMA-3.2-3B instruct model with short-answer demonstrations and evaluate on all three datasets. Both few-shot prompting and weight-level adaptation shorten answers, but only adaptation changes the weights, so the control asks whether the diversity drop and the detector gain come from the format (which a prompt can produce) or from the weights (which a prompt cannot). For TriviaQA and NQ-Open we use a $10$-shot short-answer template; for SQuAD we use a $5$-shot template instructing the model to answer using the exact words from the context (prompts in Appendix~\ref{app:fewshot_prompts}). Table~\ref{tab:fewshot-control-llama} reports the control AUROC, and Appendix Table~\ref{tab:generation_length_statistics} reports the matching few-shot length statistics.

The detector gains follow the std, not the mean. On the two generative sets the few-shot prompt does not lower the std (NQ-Open $5.02 \to 9.01$; TriviaQA $4.30 \to 4.42$), and the sampling-based detectors do not improve (SE/SC/Deg average $0.709$ vs.\ base $0.727$ on NQ-Open; $0.833$ vs.\ $0.845$ on TriviaQA). SQuAD is the only set where the prompt lowers the std ($4.72 \to 2.42$), and there the detectors do improve (SE $0.767$, Deg $0.765$ vs.\ base $0.715$ and $0.713$). SQuAD behaves differently because it is extractive: the answer is a span already present in the prompt, so the few-shot examples alone make the output uniform. On the generative sets the answer is not in the prompt, so a prompt cannot reproduce the effect; only weight-level adaptation lowers the std there.

This rules out the simple explanations. The few-shot prompt both shortens answers and adds in-context examples, yet on the generative sets the std stays high and the detectors do not improve. The reduction in output-format diversity, not response length or extra in-context examples, is what tracks the detector gains, and on generative tasks only weight-level adaptation produces it. Accordingly, we describe the effect as in-domain task adaptation reducing output-format diversity, and we refrain from framing it as a calibration or ``uncertainty awareness'' gain, which our data do not establish. The same qualitative pattern is robust to sampling temperature and adaptation configuration (Appendices~\ref{app:temperature_sensitivity} and~\ref{app:rank_sensitivity_mlp}).

\section{Behavioral Analysis}
\label{sec:behavioral}

\subsection{Case Study}

We show two typical cases generated from LLaMA-3.2-3B in the boxes below to illustrate the two hallucination behaviors. Their uncertainty scores are calculated by SelfCheckGPT. Additionally, the first case is from NQ-Open and the second is from SQuAD. Case 1 demonstrates how parameter-efficient fine-tuning handles a factual recall question while case 2 shows an extractive QA scenario.

\begin{figure}[!htbp]
\noindent
\begin{minipage}[t]{0.48\textwidth}
% Case Study 1
\begin{tcolorbox}[
  colback=casebg,
  colframe=caseblue,
  fonttitle=\bfseries,
  title={\faRobot~Case 1: \\``5000 Dollar Bill '' Hallucination},
  boxrule=0.8pt,
  arc=2pt,
  left=3pt,right=3pt,top=4pt,bottom=4pt,
  equal height group=casepair
]

\small
\vspace{0.3em}
\noindent\textbf{Q 1441:} Who is pictured on the 5000 dollar bill ? \\
\noindent\textbf{Ground Truth:} James Madison

\vspace{0.3em}
\centering
\resizebox{\linewidth}{!}{%
\footnotesize
\begin{tabular}{@{}llcl@{}}
\toprule
\textbf{Model} & \textbf{Response} & \textbf{Uncert.} & \textbf{Analysis} \\
\midrule
\rowcolor{gray!20}\textbf{Base} & ``George Washington.'' & \textcolor{dangerred}{\textbf{0.1024}} & \textcolor{dangerred}{\faExclamationTriangle~Dangerous} \\
LoRA & ``James Madison.'' & \textcolor{blue}{\textbf{0.6994}} & \textcolor{blue}{\faCheckCircle~Correct} \\
PiSSA & ``Ulysses S. Grant.'' & \textcolor{safegreen}{\textbf{0.9633}} & \textcolor{safegreen}{\faFlag~Detectable} \\
DoRA & ``James Madison.'' & \textcolor{blue}{\textbf{0.7893}} & \textcolor{blue}{\faCheckCircle~Correct} \\
\bottomrule
\end{tabular}%
}

\vspace{0.4em}
\raggedright
\fbox{\parbox{0.96\linewidth}{
\scriptsize\faLightbulb~\textbf{Insight:} \textbf{Base} model outputs an incorrect answer with low uncertainty (0.10), making the error undetectable. \textbf{LoRA} and \textbf{DoRA} correct the prediction, while \textbf{PiSSA} remains incorrect but exhibits the highest uncertainty (0.96), making the error \textbf{most detectable}.
}}

\end{tcolorbox}
\end{minipage}%
\hfill
\begin{minipage}[t]{0.48\textwidth}
% Case Study 2
\begin{tcolorbox}[
  colback=casebg,
  colframe=caseblue,
  fonttitle=\bfseries,
  title={\faRobot~Case 2: \\``Income Inequality Increase'' Hallucination},
  boxrule=0.8pt,
  arc=2pt,
  left=3pt,right=3pt,top=4pt,bottom=4pt,
  equal height group=casepair
]
\small
\vspace{0.3em}
\noindent\textbf{Q 220:} When did income inequality begin to increase in the US?
\noindent\textbf{Ground Truth:} 1970s.
\vspace{0.3em}
\centering
\resizebox{\linewidth}{!}{%
\footnotesize
\begin{tabular}{@{}llcl@{}}
\toprule
\textbf{Model} & \textbf{Response} & \textbf{Uncert.} & \textbf{Analysis} \\
\midrule
\rowcolor{gray!20}\textbf{Base} & ``After the 1970s.'' & \textcolor{dangerred}{\textbf{0.0185}} & \textcolor{dangerred}{\faExclamationTriangle~Dangerous} \\
LoRA & ``the 1970s.'' & \textcolor{blue}{\textbf{0.0296}} & \textcolor{blue}{\faCheckCircle~Correct} \\
PiSSA & ``after the 1970s.'' & \textcolor{safegreen}{\textbf{0.6651}} & \textcolor{safegreen}{\faFlag~Detectable} \\
DoRA & ``after the 1970s.'' & \textcolor{safegreen}{\textbf{0.4810}} & \textcolor{safegreen}{\faFlag~Detectable} \\
\bottomrule
\end{tabular}%
}
\vspace{0.4em}
\raggedright
\fbox{\parbox{0.96\linewidth}{
\scriptsize\faLightbulb~\textbf{Insight:} \textbf{Base} model outputs an incorrect and confident answer with high confidence. After parameter-efficient fine-tuning, \textbf{LoRA} directly corrects the prediction, while \textbf{PiSSA} remains incorrect but exhibits the highest uncertainty, making the error \textbf{most detectable} among the parameter-efficient variants; \textbf{DoRA} also stays incorrect with elevated uncertainty, but less than PiSSA.
}}
\end{tcolorbox}
\end{minipage}
\end{figure}

% 先排空 4.2 的浮动队列, 使 Figure 4 能靠近 4.3 正文
\FloatBarrier

\subsection{Where the Detector Scores Move}

\paragraph{\takeawaymark Takeaway~\#4: Parameter-efficient fine-tuning improves the performance of hallucination detectors by shifting scores away from the overconfident regime.}

\begin{figure}[!tbp]
\centering
\begin{subfigure}[b]{0.325\textwidth}
    \includegraphics[width=\textwidth]{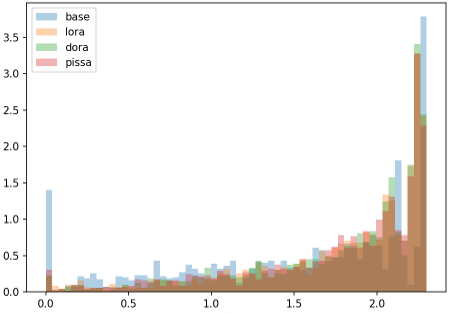}
    \caption{Semantic Entropy}
\end{subfigure}
\hfill
\begin{subfigure}[b]{0.325\textwidth}
    \includegraphics[width=\textwidth]{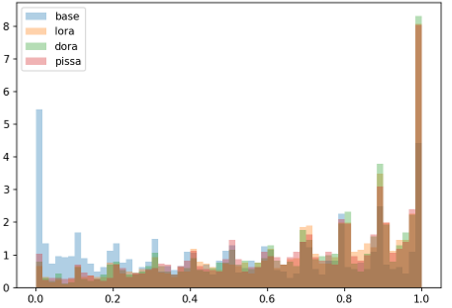}
    \caption{SelfCheckGPT}
\end{subfigure}
\hfill
\begin{subfigure}[b]{0.325\textwidth}
    \includegraphics[width=\textwidth]{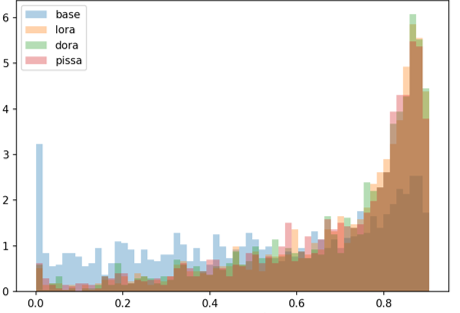}
    \caption{Degree of Uncertainty}
\end{subfigure}
\vspace{0.15em}

\begin{subfigure}[b]{0.243\textwidth}
    \includegraphics[width=\textwidth]{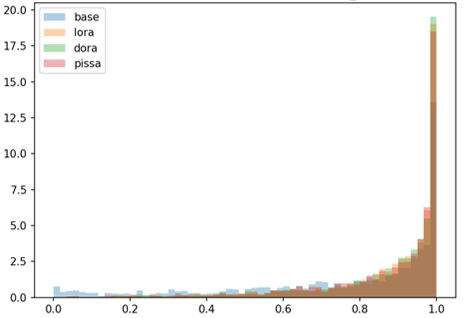}
    \caption{MSP}
\end{subfigure}
\hfill
\begin{subfigure}[b]{0.243\textwidth}
    \includegraphics[width=\textwidth]{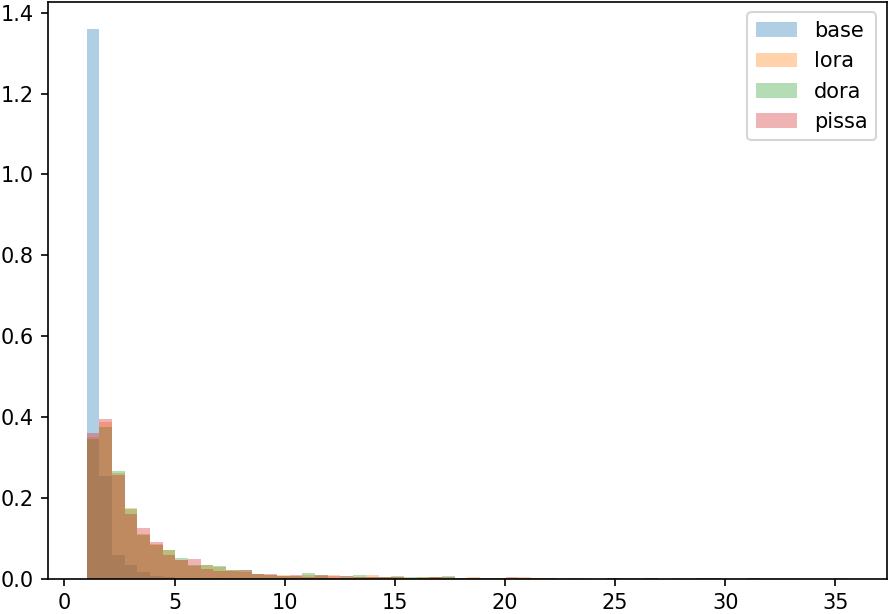}
    \caption{Perplexity}
\end{subfigure}
\hfill
\begin{subfigure}[b]{0.243\textwidth}
    \includegraphics[width=\textwidth]{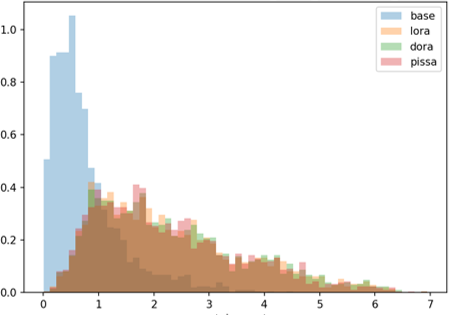}
    \caption{Mean Token Entropy}
\end{subfigure}
\hfill
\begin{subfigure}[b]{0.243\textwidth}
    \includegraphics[width=\textwidth]{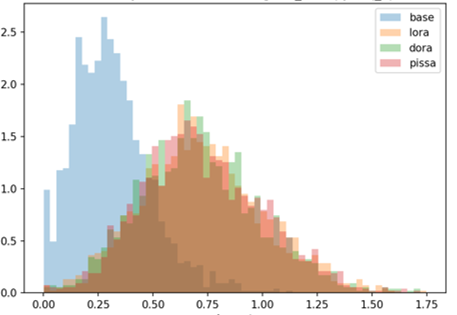}
    \caption{Predictive Entropy}
\end{subfigure}
\caption{Uncertainty score density distributions across parameter-efficient methods on Qwen-NQ-Open. X-axis represents the uncertainty score.}
\label{fig:qwen_nq_density}
\end{figure}

\begin{figure}[H]
\centering
\includegraphics[width=\linewidth]{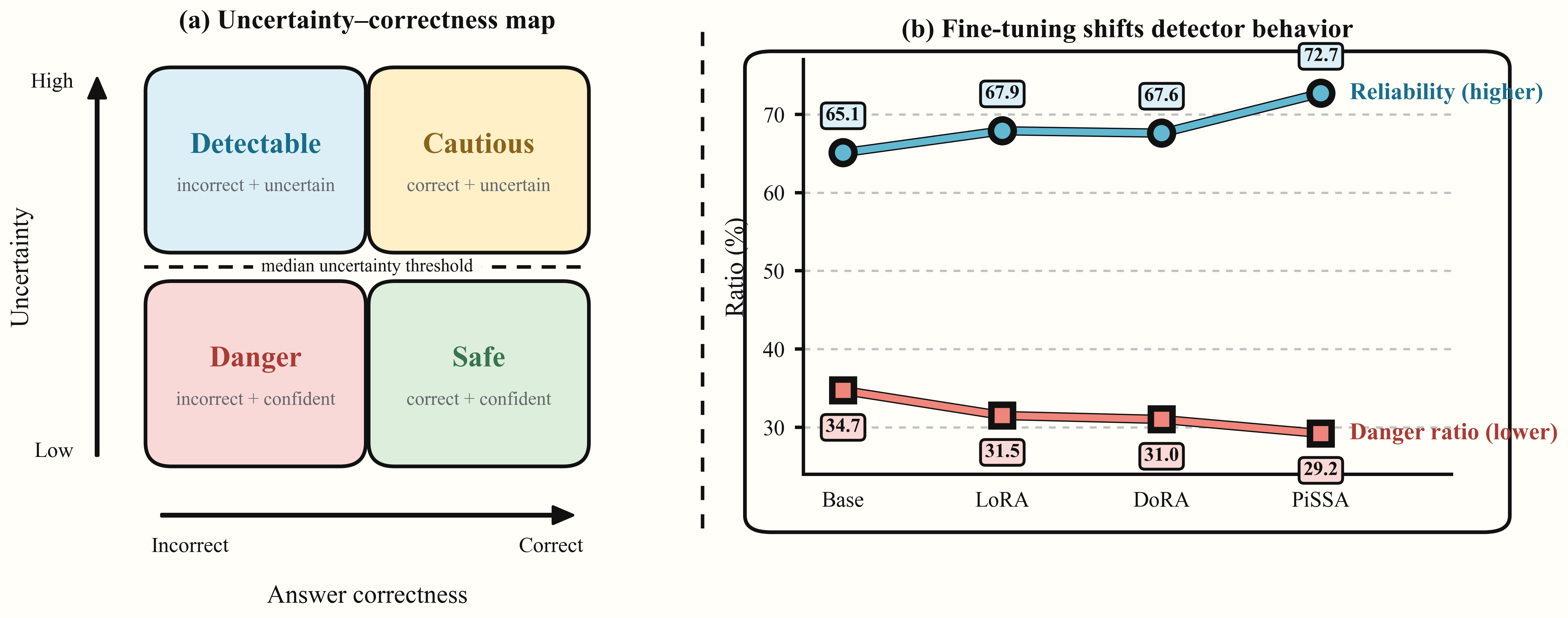}
\caption{Uncertainty--correctness analysis. (a) Operational quadrants induced by answer correctness and the model-specific median uncertainty threshold. (b) Reliability and danger ratios on NQ-Open using MSP with Llama-3.2-3B.}
\label{fig:uc_quadratic}
\end{figure}

The distribution plots (Figure~\ref{fig:qwen_nq_density}) of uncertainty scores place heavy probability mass concentrated near the lower bound of the score range (full results in Appendix~\ref{sec:density_appendix}; see Figures~\ref{fig:llama_squad_density} and~\ref{fig:mistral_triviaqa_density}). We further verify that such concentration reflects systematic overconfidence of hallucination detectors when applied to the base model. Specifically, we partition the model outputs into four quadrants based on uncertainty level and correctness (Figure~\ref{fig:uc_quadratic}). The right is the correct zone while the left is the incorrect zone. The top row is the unconfident zone while the bottom row is the confident zone. The uncertainty threshold for each model is set to its own median uncertainty score. We define the \textbf{reliability ratio} as safe / (safe + cautious) and the \textbf{danger ratio} as danger / (danger + detectable).

As shown in Figure~\ref{fig:uc_quadratic} (Right), the danger ratio, which reflects overconfidence, is at a high level of 34.7\%. After parameter-efficient fine-tuning, it decreases consistently: reliability increases from 65.11\% (Base) to 72.74\% (PiSSA), indicating that responses become obviously more trustworthy after parameter-efficient fine-tuning, while the danger ratio also decreases by 5\% (PiSSA) at most. Appendix~\ref{sec:appendix_cc_analysis} extends this analysis across datasets, and Figure~\ref{fig:cc_nqopen} shows the same qualitative shift on NQ-Open. Conversely, the reliability ratio increases after parameter-efficient fine-tuning, indicating that correct outputs are more often assigned low uncertainty. Together, these shifts help explain the performance gains of hallucination detection methods under parameter-efficient fine-tuning.

\section{Conclusions, Limitations, and Future Directions}

Across three open-weight LLMs, three QA benchmarks, and seven answer-level detectors, fine-tuning improves answer accuracy by only 1--3\% on average, yet boosts detection AUROC by up to roughly 14 \%, and full-parameter SFT reproduces this pattern on all nine backbone--dataset combinations. Within our parameter-efficient experiments, the detector gains track a reduction in output-format diversity, with answer length playing no comparable role, while white-box linear probes on the adapted hidden states show mixed results.

As a limitation, our analysis is specific to answer-level detection in fact-seeking QA, and its generalization to other tasks, model scales, and adaptation settings remains an open question that we leave to future work. On model coverage, we deliberately chose LLaMA-3.2-3B, Qwen-2.5-3B, and Mistral-7B to span distinct pretraining corpora, tokenizers, and architectural choices; we did not include families such as Gemma or DeepSeek, but because the effect we identify is mediated by generic output-behavior signals (semantic consistency and sequence confidence) that are independent of family-specific design, we expect the same qualitative trend to hold for them, and verifying this is a straightforward extension. More broadly, future work should extend this framework to settings such as claim-level verification, long-form generation, reasoning-intensive tasks, code generation, multimodal problems, and interactive dialogue.

\subsubsection*{Acknowledgments}
This work was supported by the National Institute of Standards and Technology (NIST) under Grant No.~14147259 and by the National Science Foundation (NSF) CyberTraining program under Grant No.~OAC-2417747 and 2417748. We thank Haoming Li for helpful discussions on the rebuttal and revision stage.

% \newpage
\bibliographystyle{tmlr}
\bibliography{references}

@article{yao2025reasoning,
  title={Are Reasoning Models More Prone to Hallucination?},
  author={Yao, Zijun and Liu, Yantao and Chen, Yanxu and Chen, Jianhui and Fang, Junfeng and Hou, Lei and Li, Juanzi and Chua, Tat-Seng},
  journal={arXiv preprint arXiv:2505.23646},
  year={2025}
}

@inproceedings{wang2024factuality,
 title={Factuality of large language models: A survey},
 author={Wang, Yuxia and Wang, Minghan and Manzoor, Muhammad Arslan and Liu, Fei and Georgiev, Georgi Nenkov and Das, Rocktim Jyoti and Nakov, Preslav},
 booktitle={Proceedings of the 2024 Conference on Empirical Methods in Natural Language Processing},
 pages={19519--19529},
 year={2024}
}

@article{rajpurkar2016squad,
 title={Squad: 100,000+ questions for machine comprehension of text},
 author={Rajpurkar, Pranav and Zhang, Jian and Lopyrev, Konstantin and Liang, Percy},
 journal={arXiv preprint arXiv:1606.05250},
 year={2016}

}

@inproceedings{joshi2017triviaqa,
  title     = {{TriviaQA}: A Large-Scale Distantly Supervised Challenge Dataset for Reading Comprehension},
  author    = {Joshi, Mandar and Choi, Eunsol and Weld, Daniel S. and Zettlemoyer, Luke},
  booktitle = {Proceedings of the 55th Annual Meeting of the Association for Computational Linguistics (Volume 1: Long Papers)},
  pages     = {1601--1611},
  year      = {2017}
}

@article{kwiatkowski2019natural,
  title     = {Natural Questions: A Benchmark for Question Answering Research},
  author    = {Kwiatkowski, Tom and Palomaki, Jennimaria and Redfield, Olivia and Collins, Michael and Parikh, Ankur and Alberti, Chris and Epstein, Danielle and Polosukhin, Illia and Devlin, Jacob and Lee, Kenton and others},
  journal   = {Transactions of the Association for Computational Linguistics},
  volume    = {7},
  pages     = {453--466},
  year      = {2019},
  publisher = {MIT Press}
}

@article{farquhar2024semanticentropy,
  title   = {Detecting Hallucinations in Large Language Models Using Semantic Entropy},
  author  = {Farquhar, Sebastian and Kossen, Jannik and K{\"u}hn, Lorenz and Gal, Yarin},
  journal = {Nature},
  volume  = {630},
  number  = {8017},
  pages   = {625--630},
  year    = {2024},
  publisher = {Nature Publishing Group}
}

@inproceedings{hu2022lora,
  title     = {{LoRA}: Low-Rank Adaptation of Large Language Models},
  author    = {Hu, Edward J and Shen, Yelong and Wallis, Phillip and Allen-Zhu, Zeyuan and Li, Yuanzhi and Wang, Shean and Wang, Lu and Chen, Weizhu},
  booktitle = {International Conference on Learning Representations},
  year      = {2022}
}

@article{liu2024dora,
  title   = {{DoRA}: Weight-Decomposed Low-Rank Adaptation},
  author  = {Liu, Shih-Yang and Wang, Chien-Yi and Yin, Hongxu and Molchanov, Pavlo and Wang, Yu-Chiang Frank and Cheng, Kwang-Ting and Chen, Min-Hung},
  journal = {arXiv preprint arXiv:2402.09353},
  year    = {2024}
}

@article{zhu2024pissa,
  title   = {{PiSSA}: Principal Singular Values and Singular Vectors Adaptation of Large Language Models},
  author  = {Zhu, Jiali and He, Ze-Feng and others},
  journal = {arXiv preprint arXiv:2404.02948},
  year    = {2024}
}

@inproceedings{gekhman2024does,
  title     = {Does Fine-Tuning LLMs on New Knowledge Encourage Hallucinations?},
  author    = {Gekhman, Zorik and Yona, Gal and Aharoni, Roee and Eyal, Matan and Feder, Amir and Reichart, Roi and Herzig, Jonathan},
  booktitle = {Proceedings of the 2024 Conference on Empirical Methods in Natural Language Processing},
  year      = {2024}
}

@inproceedings{zhang2024truthx,
  title     = {{TruthX}: Alleviating Hallucinations by Editing Large Language Models in Truthful Space},
  author    = {Zhang, Shaolei and Yu, Tian and Feng, Yang},
  booktitle = {Proceedings of the 62nd Annual Meeting of the Association for Computational Linguistics (Volume 1: Long Papers)},
  pages     = {8908--8949},
  year      = {2024}
}

@inproceedings{tian2024factuality,
  title     = {Fine-tuning Language Models for Factuality},
  author    = {Tian, Katherine and Mitchell, Eric and Yao, Huaxiu and Manning, Christopher D and Finn, Chelsea},
  booktitle = {International Conference on Learning Representations},
  year      = {2024}
}

@inproceedings{lin2022truthfulqa,
  title     = {{TruthfulQA}: Measuring How Models Mimic Human Falsehoods},
  author    = {Lin, Stephanie and Hilton, Jacob and Evans, Owain},
  booktitle = {Proceedings of the 60th Annual Meeting of the Association for Computational Linguistics (Volume 1: Long Papers)},
  pages     = {3214--3252},
  year      = {2022}
}

@article{kalai2025language,
  title={Why language models hallucinate},
  author={Kalai, Adam Tauman and Nachum, Ofir and Vempala, Santosh S and Zhang, Edwin},
  journal={arXiv preprint arXiv:2509.04664},
  year={2025}
}

@inproceedings{hendrycks2017baseline,
  title     = {A Baseline for Detecting Misclassified and Out-of-Distribution Examples in Neural Networks},
  author    = {Hendrycks, Dan and Gimpel, Kevin},
  booktitle = {Proceedings of ICLR},
  year      = {2017}
}

@article{du2024haloscope,
  title={Haloscope: Harnessing unlabeled llm generations for hallucination detection},
  author={Du, Xuefeng and Xiao, Chaowei and Li, Sharon},
  journal={Advances in Neural Information Processing Systems},
  volume={37},
  pages={102948--102972},
  year={2024}
}

@article{kossen2024semantic,
  title={Semantic entropy probes: Robust and cheap hallucination detection in llms},
  author={Kossen, Jannik and Han, Jiatong and Razzak, Muhammed and Schut, Lisa and Malik, Shreshth and Gal, Yarin},
  journal={arXiv preprint arXiv:2406.15927},
  year={2024}
}

@article{kadavath2022language,
  title={Language models (mostly) know what they know},
  author={Kadavath, Saurav and Conerly, Tom and Askell, Amanda and Henighan, Tom and Drain, Dawn and Perez, Ethan and Schiefer, Nicholas and Hatfield-Dodds, Zac and DasSarma, Nova and Tran-Johnson, Eli and others},
  journal={arXiv preprint arXiv:2207.05221},
  year={2022}
}

@inproceedings{manakul2023selfcheckgpt,
  title={Selfcheckgpt: Zero-resource black-box hallucination detection for generative large language models},
  author={Manakul, Potsawee and Liusie, Adian and Gales, Mark},
  booktitle={Proceedings of the 2023 conference on empirical methods in natural language processing},
  pages={9004--9017},
  year={2023}
}

@article{qiu2024semantic,
  title={Semantic density: Uncertainty quantification for large language models through confidence measurement in semantic space},
  author={Qiu, Xin and Miikkulainen, Risto},
  journal={Advances in neural information processing systems},
  volume={37},
  pages={134507--134533},
  year={2024}
}

@inproceedings{davis2006relationship,
  title={The Relationship Between Precision-Recall and ROC Curves},
  author={Davis, Jesse and Goadrich, Mark},
  booktitle={Proceedings of the 23rd International Conference on Machine Learning},
  pages={233--240},
  year={2006},
  publisher={ACM}
}

@article{lin2023generating,
  title={Generating with confidence: Uncertainty quantification for black-box large language models},
  author={Lin, Zhen and Trivedi, Shubhendu and Sun, Jimeng},
  journal={arXiv preprint arXiv:2305.19187},
  year={2023}
}

@article{grattafiori2024llama,
  title={The llama 3 herd of models},
  author={Grattafiori, Aaron and Dubey, Abhimanyu and Jauhri, Abhinav and Pandey, Abhinav and Kadian, Abhishek and Al-Dahle, Ahmad and Letman, Aiesha and Mathur, Akhil and Schelten, Alan and Vaughan, Alex and others},
  journal={arXiv preprint arXiv:2407.21783},
  year={2024}
}

@article{yang2025qwen3,
  title={Qwen3 technical report},
  author={Yang, An and Li, Anfeng and Yang, Baosong and Zhang, Beichen and Hui, Binyuan and Zheng, Bo and Yu, Bowen and Gao, Chang and Huang, Chengen and Lv, Chenxu and others},
  journal={arXiv preprint arXiv:2505.09388},
  year={2025}
}

@misc{jiang2023mistral7b,
      title={Mistral 7B}, 
      author={Albert Q. Jiang and Alexandre Sablayrolles and Arthur Mensch and Chris Bamford and Devendra Singh Chaplot and Diego de las Casas and Florian Bressand and Gianna Lengyel and Guillaume Lample and Lucile Saulnier and Lélio Renard Lavaud and Marie-Anne Lachaux and Pierre Stock and Teven Le Scao and Thibaut Lavril and Thomas Wang and Timothée Lacroix and William El Sayed},
      year={2023},
      eprint={2310.06825},
      archivePrefix={arXiv},
      primaryClass={cs.CL},
      url={https://arxiv.org/abs/2310.06825}, 
}

@article{fadeeva2023lm,
 title={LM-polygraph: Uncertainty estimation for language models},
 author={Fadeeva, Ekaterina and Vashurin, Roman and Tsvigun, Akim and Vazhentsev, Artem and Petrakov, Sergey and Fedyanin, Kirill and Vasilev, Daniil and Goncharova, Elizaveta and Panchenko, Alexander and Panov, Maxim and others},
 journal={arXiv preprint arXiv:2311.07383},
 year={2023}
}
\clearpage
\appendix

\setlength{\textfloatsep}{8pt plus 2pt minus 2pt}
\setlength{\floatsep}{8pt plus 2pt minus 2pt}
\setlength{\intextsep}{8pt plus 2pt minus 2pt}

\section{Experimental Details}
\label{sec:exper_details}
\begin{itemize}
\item All models are open-weight and are fine-tuned and evaluated locally on 8~$\times$~A6000 GPUs. Instances the model refuses to answer are handled as described in Appendix~\ref{app:protocol_details}.

\item For instruction tuning, we use a consistent prompt: \texttt{"Answer the question in a short phrase.\textbackslash n\textbackslash nQuestion: \{question\}"} for closed-book QA, and prepend the passage for SQuAD.

\item For TriviaQA and NQ-Open, we adopt a closed-book setting and input only the question, testing retrieval from parametric memory. For SQuAD, we provide both context and question, testing grounding in provided text. Since the official test sets for TriviaQA and NQ-Open often lack public groundtruth, we adopt a split-validation strategy on the validation sets.
We use 2{,}500 samples for training, 500 for validation, and the remaining $\sim$2{,}000 (1{,}805 for NQ-Open) for testing.

\item While the original evaluation protocol for TriviaQA is exact match, we find it unsuitable for generative language models, because the same answer can be expressed in many different surface forms. For example, the question ``What is the capital of the USA?'' has the ground truth ``Washington, D.C.'', yet a model answering ``Washington'' or ``Washington DC'' may not be scored as correct under exact match. In addition, some questions are refused rather than answered, and EM/F1 would require further pattern matching to distinguish these refusals from incorrect answers. We therefore use LLM-as-a-Judge for the QA datasets. We deploy GPT-5-mini as our judging model and report accuracy as a percentage. 

\end{itemize}

\section{Detection and Judging Protocol Details}
\label{app:protocol_details}
This section specifies the detection and judging protocols used throughout the paper.

\paragraph{Stochastic sampling for sample-based detectors.}
For the sample-based detectors (Semantic Entropy, SelfCheckGPT, and Degree of Uncertainty), we draw $K=10$ stochastic samples per question. Unless otherwise noted (e.g., the temperature study in Appendix~\ref{app:temperature_sensitivity}), the decoding temperature is $1.0$.

\paragraph{Semantic clustering and semantic entropy.}
We cluster the $K=10$ sampled responses by semantic equivalence using \texttt{microsoft/deberta-v2-xlarge-mnli} as the NLI judge: two responses are assigned to the same cluster only when they mutually entail each other. After clustering, we aggregate the normalized response likelihoods within each semantic cluster,
\[
p(C_m \mid x) =
\frac{\sum_{r_i \in C_m} p(r_i \mid x)}
{\sum_j p(r_j \mid x)} ,
\]
implemented with log-sum-exp for numerical stability, and compute semantic entropy over the cluster distribution,
\[
H_{\mathrm{sem}}(x) = - \sum_m p(C_m \mid x)\log p(C_m \mid x).
\]
This follows \citet{farquhar2024semanticentropy}, treating paraphrases as the same answer while separating semantically distinct generations.

\paragraph{Probe token position.}
For the white-box linear probe, the input representation is the hidden state of the second-to-last token of the response.

\paragraph{Judge prompt and reference visibility.}
The judge sees the reference answers. The prompt differs by dataset; all three judge prompts are given below. For NQ-Open, a simplified two-label variant is used when refusal detection is disabled (\texttt{detect\_unk=False}); the refusal counts in Table~\ref{tab:refusal_counts} come from the three-label prompt.

\begin{promptbox}{TriviaQA judge prompt}
System: You are an expert answer-grader.
Inputs: Question; Candidate Answer; Reference Answers (a list, any one may be correct).
Labels:
  YES:     Candidate states the same core fact(s) as any reference
           (synonyms, reordering, extra words allowed).
  NO:      Candidate answers but does not match any reference,
           or contradicts a reference, or adds factual errors.
  UNKNOWN: Candidate does not actually answer the question
           (says it does not know, refuses, or asks for more info).
Procedure (in order):
  1) If an UNKNOWN condition holds, label UNKNOWN.
  2) Otherwise compare to the references: a match is YES, otherwise NO.
  3) Give a one-sentence rationale.
Output:
  Label: YES | NO | UNKNOWN
  Rationale: <one short sentence>
\end{promptbox}

\begin{promptbox}{NQ-Open judge prompt}
[System]
You are an expert answer-grader.

### Inputs
- Question
- Candidate Answer
- Reference Answers (a list; any one may be correct)

### Labels
- YES: Candidate clearly states the same core fact(s) as any reference
  (synonyms/reordering/extra fluff allowed).
- NO: Candidate gives an answer but it does not match any reference, or it
  contradicts a reference / adds factual errors.
- UNKNOWN: Candidate does **not** actually answer the question.

### What counts as "does not answer" (label UNKNOWN)
- Says they don't know/can't answer/refuses/asks for more info only
  (e.g., "I can't determine from the context," "not sure," "as an AI...").

### Procedure (apply in order)
1) **Non-answer check**: If any UNKNOWN condition holds, label **UNKNOWN**.
2) **Match check**: If the candidate asserts a concrete answer, compare to
   references.
   - If it matches any reference's core fact(s) -> **YES**.
   - Otherwise, or if it contradicts a reference/has factual errors -> **NO**.
3) Provide a one-sentence rationale.

### Output (exactly this format)
Label: YES | NO | UNKNOWN
Rationale: <one short sentence>

[User]
Question:
{question}

Candidate Answer:
{predicted_answer}

Reference Answers:
{', '.join(aliases)}
\end{promptbox}

\begin{promptbox}{NQ-Open judge prompt, simplified variant (\texttt{detect\_unk=False})}
You are an expert answer-grader.

### Task
1. Compare **Candidate Answer** to the list of **Reference Answers**.
2. If the candidate expresses the *same core fact(s)* as **any** reference
   (allowing for synonyms, re-ordering, or extra irrelevant detail), mark it **YES**.
3. Otherwise mark **NO**.
4. If the candidate contradicts a reference or introduces factual errors, mark **NO**
   even if it partly overlaps.
5. Also include a one-sentence rationale.
\end{promptbox}

\begin{promptbox}{SQuAD judge prompt}
[System]
You are an expert answer-grader.

[User]
We are assessing the quality of answers to the following question:
{question}

The proposed answer is:
{predicted_answer}

The expected answer is:
{', '.join(aliases)}

Within the context of the question,
does the proposed answer mean the same as the expected answer?
Respond only with yes or no.
\end{promptbox}

\paragraph{Few-shot format-control prompts.}
\label{app:fewshot_prompts}
The format control in Section~\ref{sec:format_control} prompts the LLaMA-3.2-3B base model with short-answer demonstrations and no weight update. For TriviaQA and NQ-Open we use the following template with $10$-shot examples:
\begin{promptbox}{Few-shot format-control prompt: TriviaQA and NQ-Open ($10$-shot)}
system: You are a helpful assistant.
user:   Answer the question in a short phrase.
Question: {sample}
\end{promptbox}
For SQuAD we use the following template with $5$-shot examples (the SQuAD sample with context is too long for $10$ shots):
\begin{promptbox}{Few-shot format-control prompt: SQuAD ($5$-shot)}
system: You are a helpful assistant. Answer only the current question in a
short phrase, using the exact words from the context.
user:   {sample}
assistant:
{answer to that sample}
\end{promptbox}

\paragraph{Refusal counts.}
Refusals correspond to the judge's UNKNOWN label and are rare (Table~\ref{tab:refusal_counts}). The highest refusal rate is on LLaMA + NQ-Open + Base, at $5.87\%$. Across datasets and models, the base model refuses far more often than its adapted variants. Because the overall refusal rate is low, we discard the refused instances from the corresponding evaluation.

\begin{table}[!htbp]
\centering
\appendixtablestyle
\caption{Refusal counts and rates, computed as the UNKNOWN-label count divided by the number of judged examples.}
\label{tab:refusal_counts}
\begin{tabular}{llccccc}
\toprule
Model & Dataset & $N$ & Base & LoRA & DoRA & PiSSA \\
\midrule
Llama & trivia\_qa & 2000 & 77 (3.85\%) & 10 (0.50\%) & 9 (0.45\%) & 8 (0.40\%) \\
Llama & squad & 2000 & 1 (0.05\%) & 1 (0.05\%) & 1 (0.05\%) & 1 (0.05\%) \\
Llama & nq\_open & 1805 & 106 (5.87\%) & 3 (0.17\%) & 2 (0.11\%) & 0 (0.00\%) \\
Qwen & trivia\_qa & 2000 & 7 (0.35\%) & 0 (0.00\%) & 1 (0.05\%) & 0 (0.00\%) \\
Qwen & squad & 2000 & 0 (0.00\%) & 0 (0.00\%) & 0 (0.00\%) & 0 (0.00\%) \\
Qwen & nq\_open & 1805 & 98 (5.43\%) & 3 (0.17\%) & 7 (0.39\%) & 2 (0.11\%) \\
Mistral & trivia\_qa & 2000 & 1 (0.05\%) & 0 (0.00\%) & 0 (0.00\%) & 0 (0.00\%) \\
Mistral & squad & 2000 & 0 (0.00\%) & 0 (0.00\%) & 0 (0.00\%) & 0 (0.00\%) \\
Mistral & nq\_open & 1805 & 68 (3.77\%) & 5 (0.28\%) & 5 (0.28\%) & 2 (0.11\%) \\
\bottomrule
\end{tabular}
\end{table}

\section{LLM-as-Judge Validation and Consistency Analysis}
\label{app:judge_consistency}
We validate the LLM-as-judge protocol from two complementary angles: (i) agreement with human annotators, which grounds the judge in human feedback, and (ii) consistency across multiple independent frontier judges.

\paragraph{Human validation.}
To verify that the GPT-5-mini judge is grounded in human feedback, we recruited five computer-science PhD students to independently label $100$ questions sampled from the TriviaQA validation set (answers generated by LLaMA-3.2-3B-Instruct), and took the majority vote as the final human label. The human labels agree with the GPT-5-mini judge on $99\%$ of the samples, with high inter-annotator agreement (Fleiss' $\kappa>0.85$). The single disagreement came from a response that was off-topic and incorrect but contained the phrase ``I don't know'': the LLM judge interpreted this as a refusal and labeled it UNKNOWN, whereas the human annotators labeled it NO (incorrect). This high agreement indicates that the automatic judge closely tracks human correctness judgments on fact-seeking QA.

\paragraph{Cross-judge consistency.}
In the main paper, we use GPT-5-mini as the judge. We further conduct a cross-judge consistency study using three frontier models as independent judges: GPT-5-mini, Claude Opus 4.6, and Gemini 3 Pro Preview. We evaluate the first 200 test samples from TriviaQA, assessing answers generated by LLaMA-3.2-3B-Instruct. We restrict this study to a single model and dataset for cost reasons: running Claude and Gemini as judges on 200 items costs roughly \$150. The three APIs also apply different safety filters and decline to label questions they deem sensitive; GPT-5-mini and Claude skip only a handful, whereas Gemini 3 Pro Preview skips nearly $30\%$. When computing agreement we keep only the questions that no judge skipped, which leaves fewer than 200 items, so the reported rates are not exact multiples of $0.5\%$.

\paragraph{Judges.}
Azure GPT-5-mini / Claude Opus 4.6 / Gemini 3 Pro Preview

\begin{table}[!htbp]
\centering
\appendixtablestyle
\caption{Judge verdict distribution on TriviaQA (test split). Percentages are computed over the items each judge actually labeled; agreement in Table~\ref{tab:judge_pairwise_agreement} is computed over the $\le 200$ items labeled by all three judges.}
\label{tab:judge_verdict_distribution}
\begin{tabular}{lcc}
\toprule
\textbf{Judge} & \textbf{Correct} & \textbf{Incorrect} \\
\midrule
Azure GPT-5-mini   & 67.50\% & 32.50\% \\
Claude Opus 4.6    & 69.67\% & 30.33\% \\
Gemini 3 Pro Preview & 66.87\% & 33.13\% \\
\bottomrule
\end{tabular}
\end{table}

\begin{table}[!htbp]
\centering
\appendixtablestyle
\caption{Pairwise agreement rates among judges.}
\label{tab:judge_pairwise_agreement}
\begin{tabular}{lc}
\toprule
\textbf{Judge Pair} & \textbf{Agreement Rate} \\
\midrule
Azure GPT-5-mini vs. Claude Opus 4.6   & 98.36\% \\
Azure GPT-5-mini vs. Gemini 3 Pro Preview & 98.80\% \\
Claude Opus 4.6 vs. Gemini 3 Pro Preview  & 97.86\% \\
All three judges agree & 97.73\% \\
\bottomrule
\end{tabular}
\end{table}

Pairwise agreement exceeds 97\% across all comparisons, suggesting that for factual QA tasks such as TriviaQA, frontier models apply highly consistent evaluation criteria.

\section{Fine-tuning Loss Curves}
\label{sec:appendix_results}
\begin{figure}[p]
\centering
\includegraphics[width=0.99\textwidth]{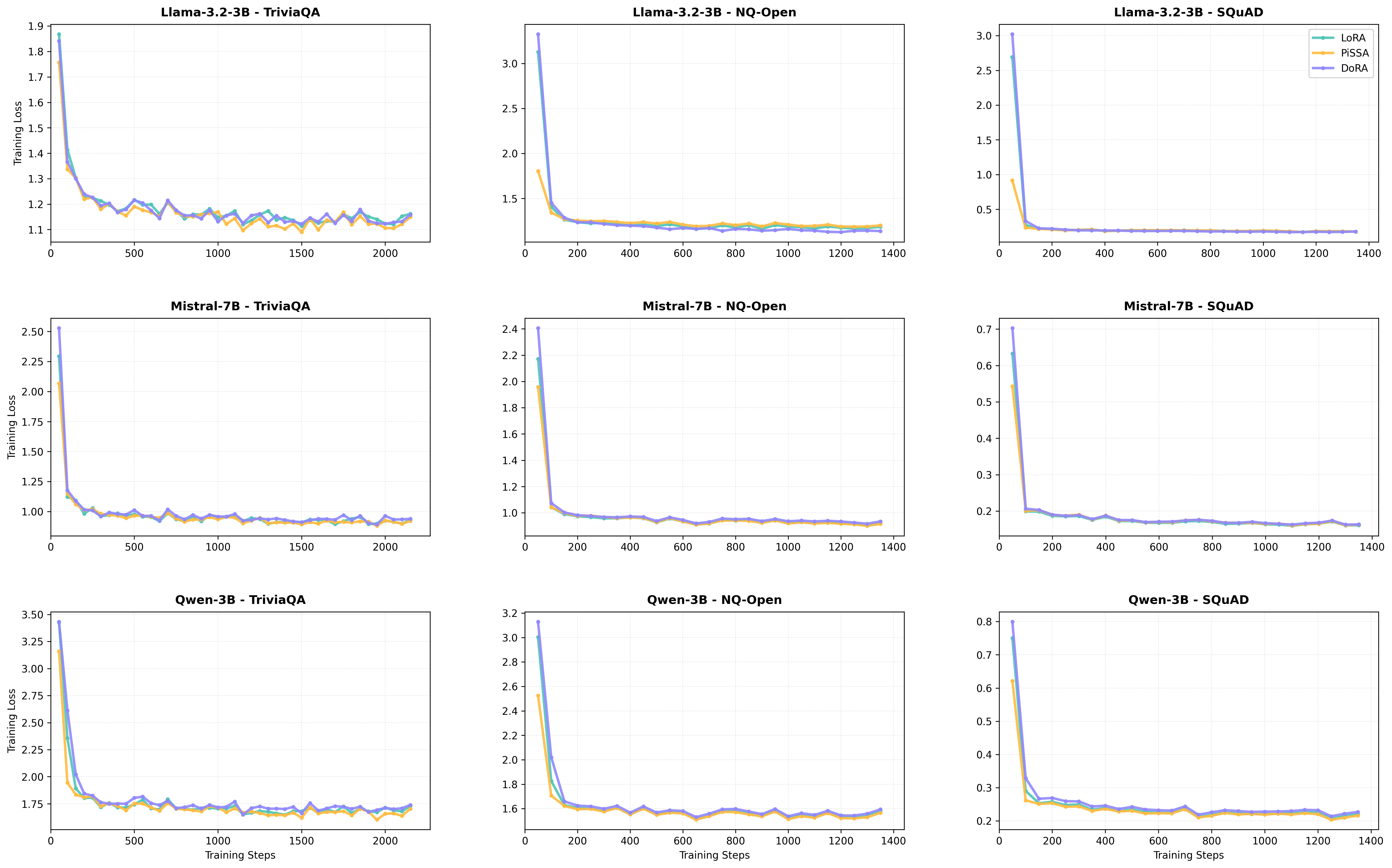}
\caption{\textbf{Training loss curves.} }
\label{fig:training_loss}
\end{figure}

\begin{figure}[p]
\centering
\includegraphics[width=0.99\textwidth]{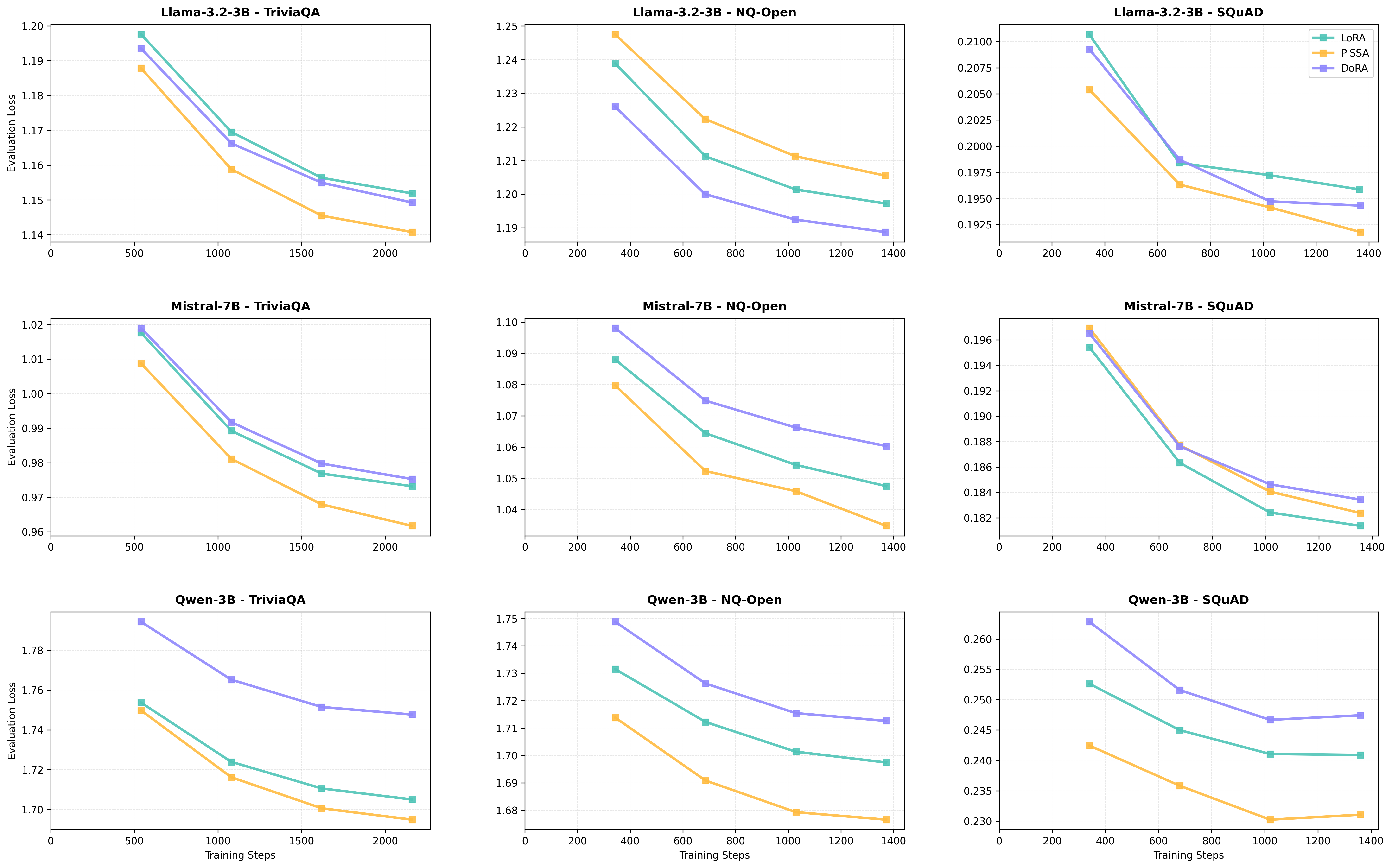}
\caption{\textbf{Evaluation loss curves.} }
\label{fig:eval_loss}
\end{figure}

(Figure~\ref{fig:training_loss} and~\ref{fig:eval_loss}). This section reports training and evaluation loss curves for all parameter-efficient configurations across models and datasets. The primary purpose of these figures is to verify the correctness and stability of the fine-tuning process. As shown in Figures~\ref{fig:training_loss} and~\ref{fig:eval_loss}, LoRA, DoRA, and PiSSA consistently exhibit rapid convergence, smooth loss decay, and no signs of divergence or overfitting on LLaMA, Mistral, and Qwen backbones. Training and evaluation losses closely track each other across datasets, indicating that the observed differences in hallucination detection performance in the main paper are not artifacts of unstable training, optimization failure, or misconfiguration, but reflect genuine epistemic effects induced by parameter-efficient fine-tuning.

\section{TriviaQA and NQ-Open hallucination detection AUPR score}
(Table.~\ref{tab:triviaqa_aupr}~\ref{tab:nqopen_aupr}).
This section reports AUPR scores for TriviaQA and NQ-Open, complementing the AUROC analysis in the main paper. Because open-domain QA exhibits less severe class imbalance than SQuAD, these results illustrate why fine-tuning-induced improvements in uncertainty detectors do not always translate into consistent AUPR gains, supporting the discussion in Takeaway~\#2.

\begin{table}[!htbp]
\centering
\appendixtablestyle
\caption{AUPR scores for hallucination detection baselines on TriviaQA. The Full SFT row reports full-parameter supervised fine-tuning as a reference point: a single run at three decimals, not hyperparameter-tuned, and excluded from the column-wise bolding (Appendix~\ref{app:sft_baseline}).}
\resizebox{0.98\textwidth}{!}{%
\begin{tabular}{llccccccc}
\toprule
& & \multicolumn{3}{c}{\textbf{Semantic Consistency}} & \multicolumn{2}{c}{\textbf{Confidence}} & \multicolumn{2}{c}{\textbf{Entropy}} \\
\cmidrule(lr){3-5} \cmidrule(lr){6-7} \cmidrule(lr){8-9}
\textbf{Model} & \textbf{Method} & \textbf{SE} & \textbf{SC} & \textbf{Deg} & \textbf{MSP} & \textbf{Perplexity} & \textbf{MeanEnt} & \textbf{PreEnt} \\ 
\midrule
\multirow{5}{*}{\shortstack[l]{LLaMA-3.2\\-3B-Instruct}}
& \cellcolor{gray!20}Base & \cellcolor{gray!20}0.6750 & \cellcolor{gray!20}0.7942 & \cellcolor{gray!20}0.7669 & \cellcolor{gray!20}0.6761 & \cellcolor{gray!20}0.7245 & \cellcolor{gray!20}0.7450 & \cellcolor{gray!20}0.6198 \\
& DoRA & 0.7618 & 0.7768 & 0.7535 & 0.8011 & 0.7431 & 0.6907 & 0.6585 \\
& LoRA & 0.7155 & 0.7461 & 0.7030 & 0.7652 & 0.7071 & 0.6643 & 0.6198 \\
& PiSSA & \textbf{0.7782} & \textbf{0.8081} & \textbf{0.7686} & \textbf{0.8260} & \textbf{0.7826} & \textbf{0.7487} & \textbf{0.6865} \\
& Full SFT & 0.791 & 0.797 & 0.767 & 0.813 & 0.715 & 0.688 & 0.676 \\
\addlinespace
\multirow{5}{*}{\shortstack[l]{Mistral-7B\\-Instruct-v0.3}}
& \cellcolor{gray!20}Base & \cellcolor{gray!20}0.5960 & \cellcolor{gray!20}0.6181 & \cellcolor{gray!20}0.6634 & \cellcolor{gray!20}0.5536 & \cellcolor{gray!20}0.4625 & \cellcolor{gray!20}\textbf{0.5187} & \cellcolor{gray!20}0.5199 \\
& DoRA & \textbf{0.7289} & \textbf{0.7503} & \textbf{0.7263} & \textbf{0.7142} & 0.5899 & 0.5137 & 0.5216 \\
& LoRA & 0.6960 & 0.7462 & 0.7190 & 0.7068 & \textbf{0.5928} & 0.5185 & 0.5261 \\
& PiSSA & 0.6675 & 0.7241 & 0.6938 & 0.6922 & 0.5599 & 0.4905 & \textbf{0.5339} \\
& Full SFT & 0.843 & 0.848 & 0.844 & 0.840 & 0.766 & 0.734 & 0.784 \\
\addlinespace
\multirow{5}{*}{\shortstack[l]{Qwen2.5\\-3B-Instruct}}
& \cellcolor{gray!20}Base & \cellcolor{gray!20}0.8437 & \cellcolor{gray!20}\textbf{0.8668} & \cellcolor{gray!20}\textbf{0.8758} & \cellcolor{gray!20}0.8156 & \cellcolor{gray!20}0.8265 & \cellcolor{gray!20}\textbf{0.8488} & \cellcolor{gray!20}\textbf{0.8049} \\
& DoRA & 0.8544 & 0.8660 & 0.8514 & 0.8986 & \textbf{0.8530} & 0.8088 & 0.7904 \\
& LoRA & \textbf{0.8591} & 0.8657 & 0.8472 & \textbf{0.9016} & 0.8525 & 0.8102 & 0.7774 \\
& PiSSA & 0.8433 & 0.8633 & 0.8519 & 0.9004 & 0.8377 & 0.7992 & 0.7771 \\
& Full SFT & 0.870 & 0.868 & 0.842 & 0.896 & 0.838 & 0.797 & 0.766 \\
\bottomrule
\end{tabular}%
}
\label{tab:triviaqa_aupr}
\end{table}

\begin{table}[!htbp]
\centering
\appendixtablestyle
\caption{AUPR scores for hallucination detection baselines on NQ-Open. The Full SFT row reports full-parameter supervised fine-tuning as a reference point: a single run at three decimals, not hyperparameter-tuned, and excluded from the column-wise bolding (Appendix~\ref{app:sft_baseline}).}
\resizebox{0.98\textwidth}{!}{%
\begin{tabular}{llccccccc}
\toprule
& & \multicolumn{3}{c}{\textbf{Semantic Consistency}} & \multicolumn{2}{c}{\textbf{Confidence}} & \multicolumn{2}{c}{\textbf{Entropy}} \\
\cmidrule(lr){3-5} \cmidrule(lr){6-7} \cmidrule(lr){8-9}
\textbf{Model} & \textbf{Method} & \textbf{SE} & \textbf{SC} & \textbf{Deg} & \textbf{MSP} & \textbf{Perplexity} & \textbf{MeanEnt} & \textbf{PreEnt} \\ 
\midrule
\multirow{5}{*}{\shortstack[l]{LLaMA-3.2\\-3B-Instruct}}
& \cellcolor{gray!20}Base & \cellcolor{gray!20}0.6698 & \cellcolor{gray!20}0.7459 & \cellcolor{gray!20}0.7391 & \cellcolor{gray!20}0.6758 & \cellcolor{gray!20}0.7188 & \cellcolor{gray!20}0.7203 & \cellcolor{gray!20}0.6319 \\
& DoRA & 0.7062 & 0.7418 & 0.7070 & 0.7404 & 0.7097 & 0.6810 & 0.6558 \\
& LoRA & 0.7115 & 0.7412 & 0.7060 & \textbf{0.7470} & 0.7280 & 0.7003 & 0.6738 \\
& PiSSA & \textbf{0.7581} & \textbf{0.7908} & \textbf{0.7619} & 0.7803 & \textbf{0.7628} & \textbf{0.7372} & \textbf{0.7146} \\
& Full SFT & 0.776 & 0.805 & 0.779 & 0.798 & 0.762 & 0.739 & 0.768 \\
\addlinespace
\multirow{5}{*}{\shortstack[l]{Mistral-7B\\-Instruct-v0.3}}
& \cellcolor{gray!20}Base & \cellcolor{gray!20}0.6920 & \cellcolor{gray!20}0.7600 & \cellcolor{gray!20}\textbf{0.7759} & \cellcolor{gray!20}0.6742 & \cellcolor{gray!20}0.6400 & \cellcolor{gray!20}0.6617 & \cellcolor{gray!20}0.6393 \\
& DoRA & 0.7193 & 0.7682 & 0.7515 & 0.7435 & 0.6689 & 0.6406 & 0.6810 \\
& LoRA & \textbf{0.7284} & \textbf{0.7716} & 0.7532 & \textbf{0.7495} & \textbf{0.6809} & \textbf{0.6522} & \textbf{0.6930} \\
& PiSSA & 0.7281 & 0.7708 & 0.7483 & 0.7394 & 0.6660 & 0.6444 & 0.6743 \\
& Full SFT & 0.861 & 0.878 & 0.853 & 0.887 & 0.844 & 0.818 & 0.844 \\
\addlinespace
\multirow{5}{*}{\shortstack[l]{Qwen2.5\\-3B-Instruct}}
& \cellcolor{gray!20}Base & \cellcolor{gray!20}\textbf{0.8769} & \cellcolor{gray!20}0.8904 & \cellcolor{gray!20}\textbf{0.8964} & \cellcolor{gray!20}0.8428 & \cellcolor{gray!20}\textbf{0.8532} & \cellcolor{gray!20}\textbf{0.8603} & \cellcolor{gray!20}0.8265 \\
& DoRA & 0.8620 & \textbf{0.9049} & 0.8811 & 0.8770 & 0.8407 & 0.8154 & \textbf{0.8385} \\
& LoRA & 0.8557 & 0.9003 & 0.8771 & \textbf{0.8779} & 0.8418 & 0.8161 & 0.8391 \\
& PiSSA & 0.8538 & 0.8902 & 0.8789 & 0.8859 & 0.8385 & 0.8095 & 0.8309 \\
& Full SFT & 0.874 & 0.885 & 0.876 & 0.880 & 0.839 & 0.811 & 0.828 \\
\bottomrule
\end{tabular}%
}
\label{tab:nqopen_aupr}
\end{table}

\section{The Uncertainty Score Density}
\label{sec:density_appendix}
\begin{figure}[htbp]
\centering
% 第一行：3张图
\begin{subfigure}[b]{0.32\textwidth}
    \includegraphics[width=\textwidth]{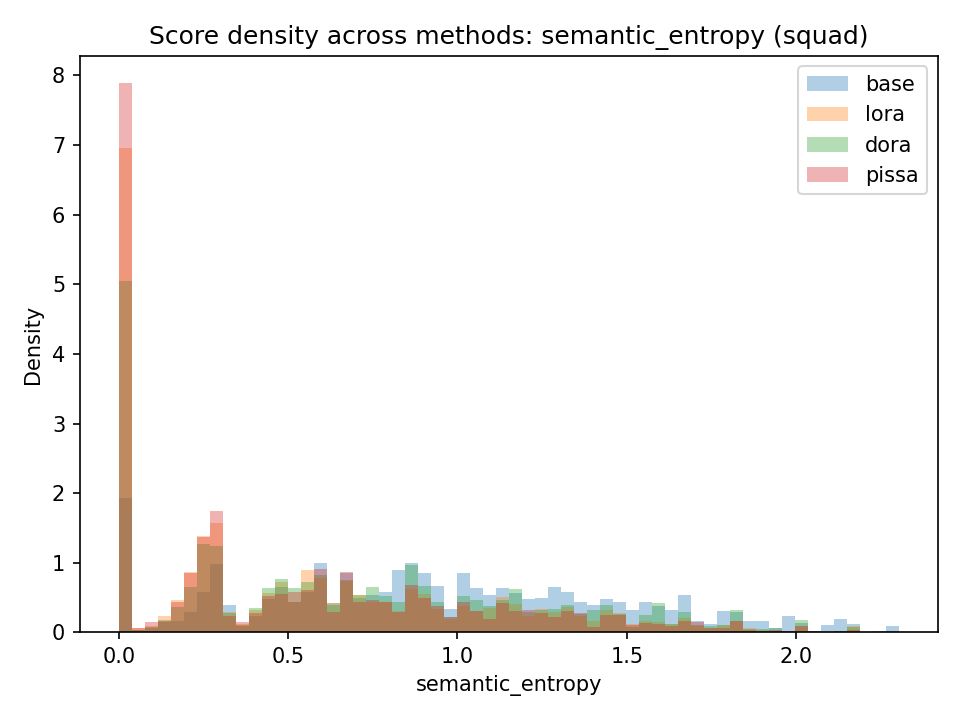}
    \caption{Semantic Entropy}
\end{subfigure}
\hfill
\begin{subfigure}[b]{0.32\textwidth}
    \includegraphics[width=\textwidth]{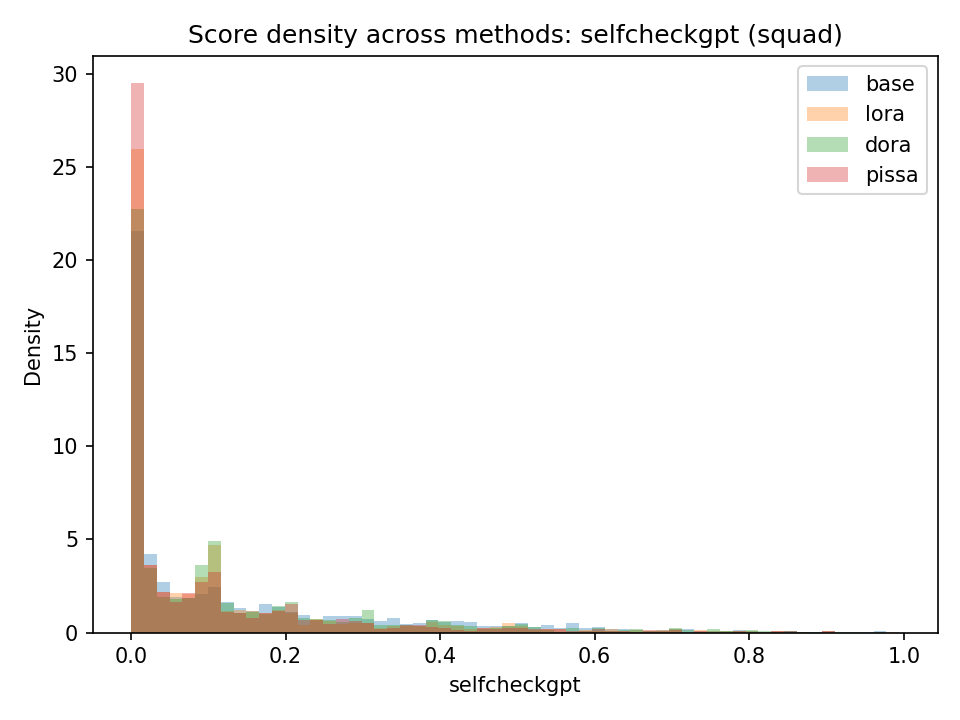}
    \caption{SelfCheckGPT}
\end{subfigure}
\hfill
\begin{subfigure}[b]{0.32\textwidth}
    \includegraphics[width=\textwidth]{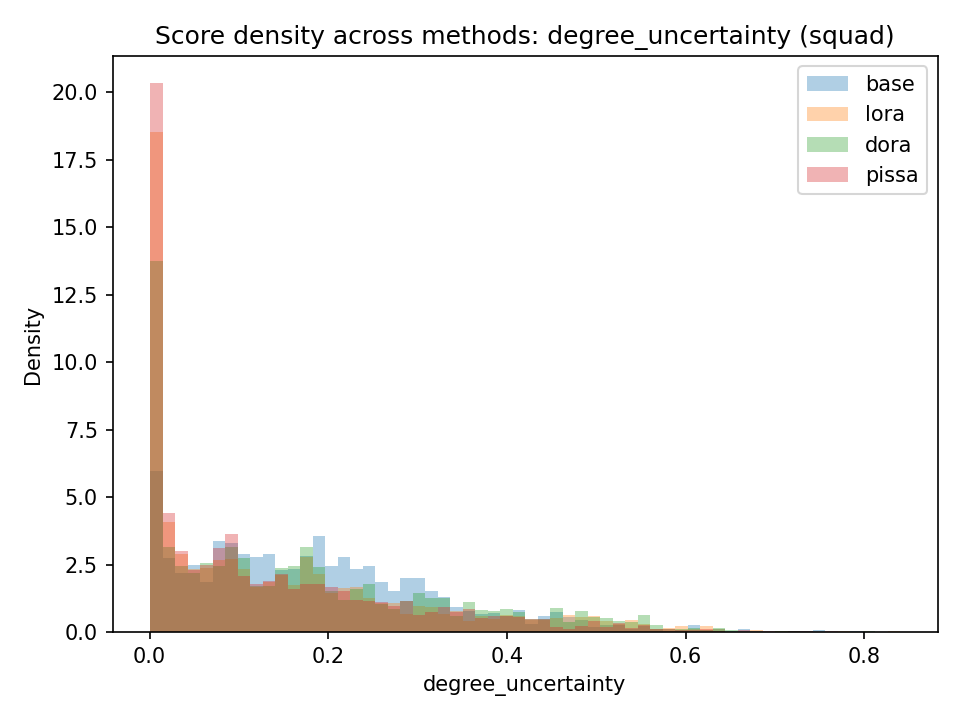}
    \caption{Degree of Uncertainty}
\end{subfigure}

\vspace{0.5em}

% 第二行：4张图
\begin{subfigure}[b]{0.24\textwidth}
    \includegraphics[width=\textwidth]{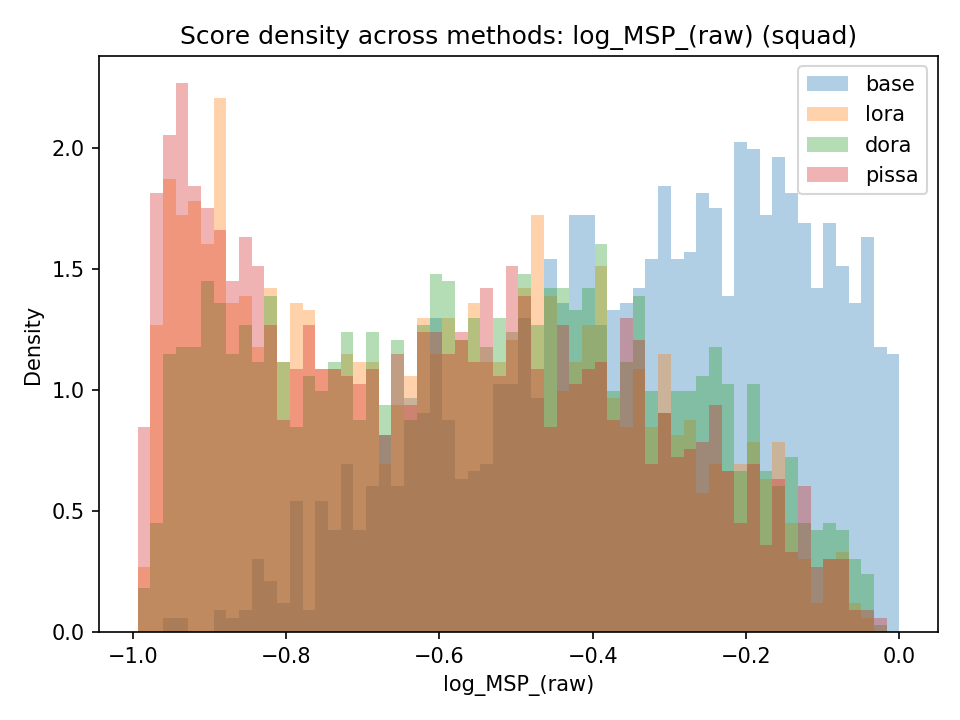}
    \caption{MSP}
\end{subfigure}
\hfill
\begin{subfigure}[b]{0.24\textwidth}
    \includegraphics[width=\textwidth]{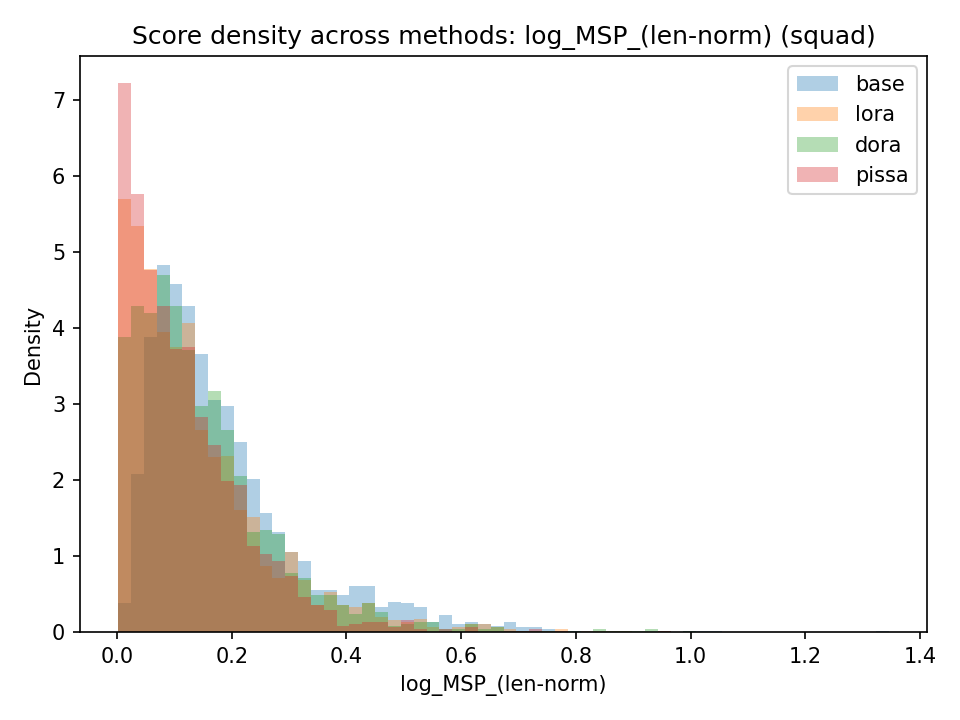}
    \caption{Perplexity}
\end{subfigure}
\hfill
\begin{subfigure}[b]{0.24\textwidth}
    \includegraphics[width=\textwidth]{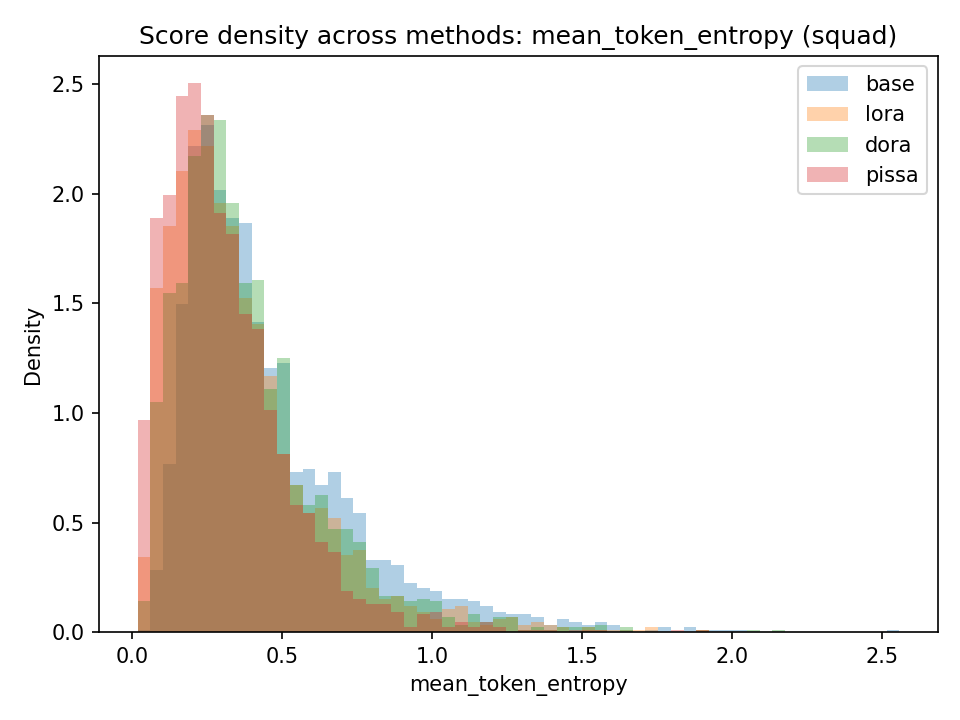}
    \caption{Mean Token Entropy}
\end{subfigure}
\hfill
\begin{subfigure}[b]{0.24\textwidth}
    \includegraphics[width=\textwidth]{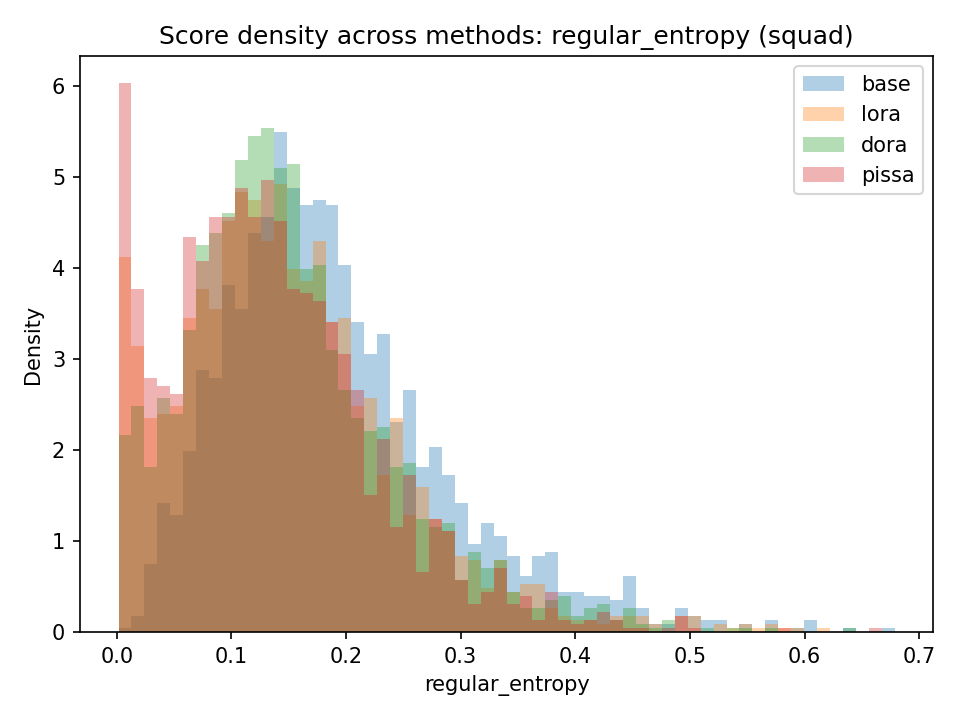}
    \caption{Predictive Entropy}
\end{subfigure}

\caption{Uncertainty score density distributions across parameter-efficient methods on LLaMA-3.2-3B-Instruct (SQuAD). Top row: semantic-level detectors. Bottom row: token-level  detectors. X-axis represents the uncertainty score.}
\label{fig:llama_squad_density}
\end{figure}

\begin{figure}[htbp]
\centering

% 第一行：3张图
\begin{subfigure}[b]{0.32\textwidth}
    \includegraphics[width=\textwidth]{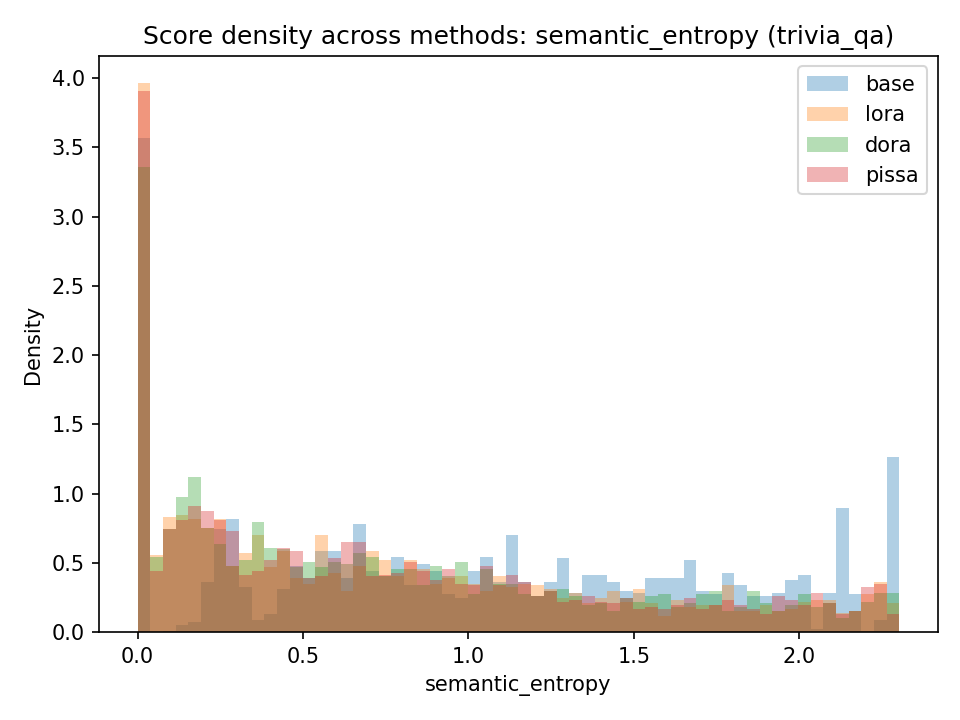}
    \caption{Semantic Entropy}
\end{subfigure}
\hfill
\begin{subfigure}[b]{0.32\textwidth}
    \includegraphics[width=\textwidth]{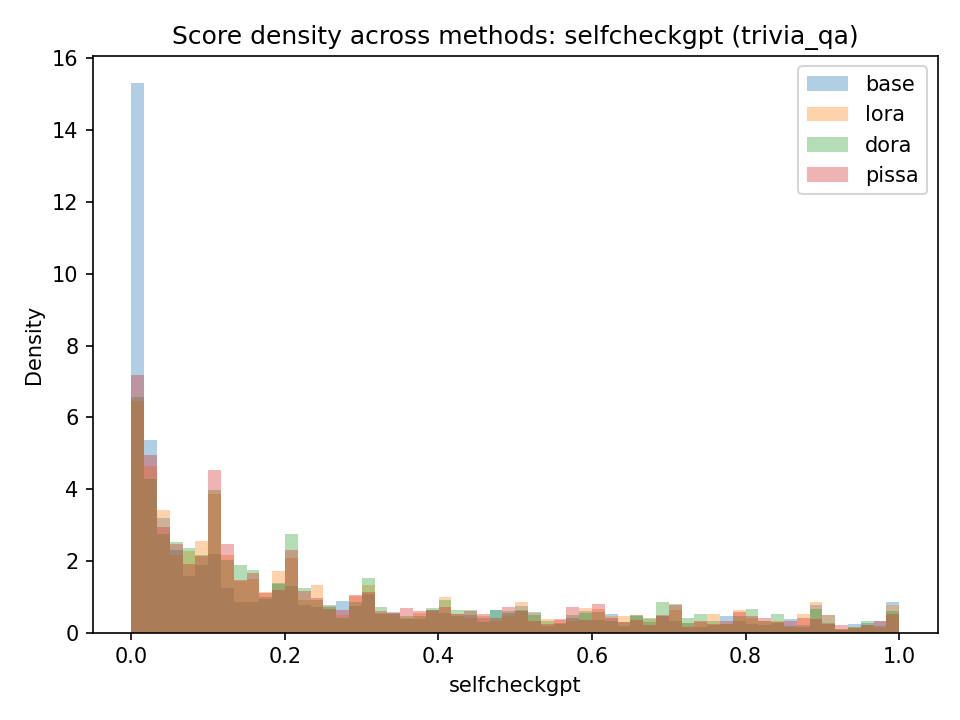}
    \caption{SelfCheckGPT}
\end{subfigure}
\hfill
\begin{subfigure}[b]{0.32\textwidth}
    \includegraphics[width=\textwidth]{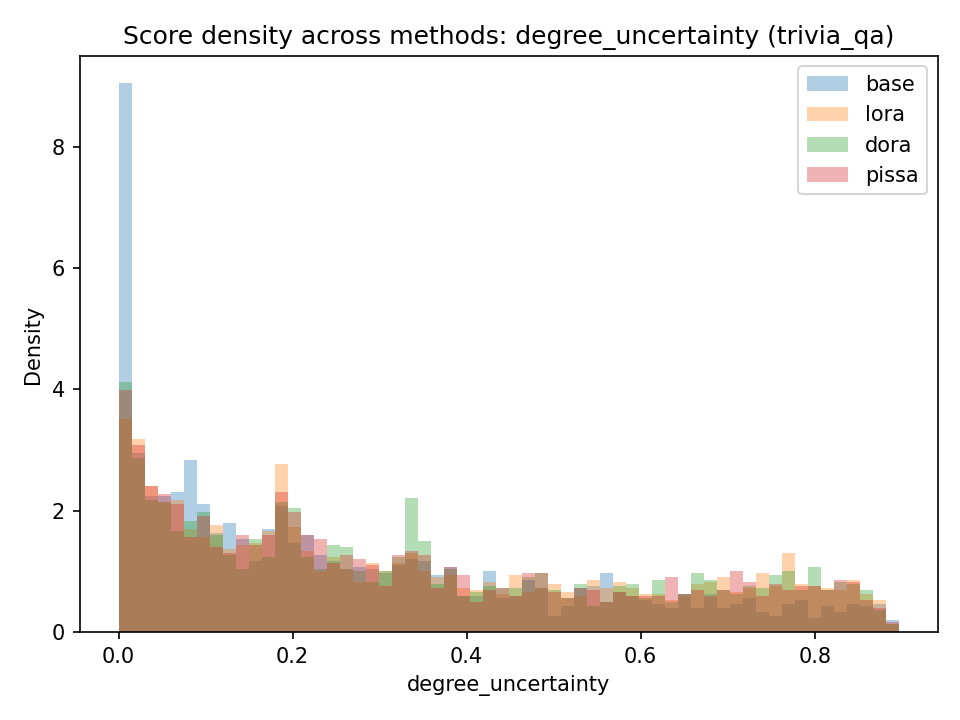}
    \caption{Degree of Uncertainty}
\end{subfigure}
\vspace{0.5em}

\begin{subfigure}[b]{0.24\textwidth}
    \includegraphics[width=\textwidth]{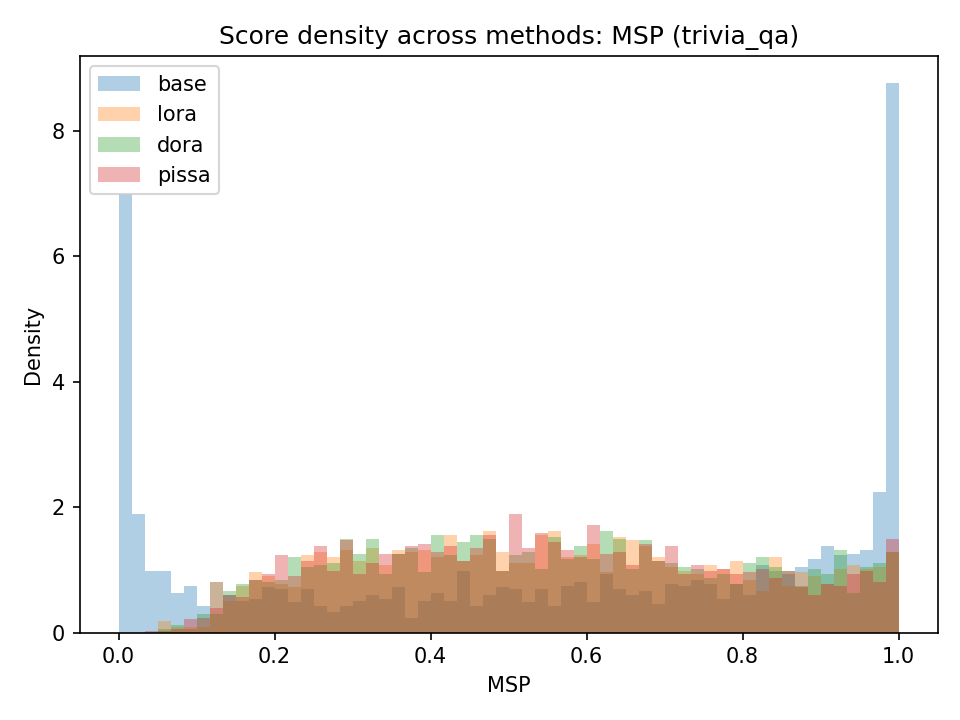}
    \caption{MSP}
\end{subfigure}
\hfill
\begin{subfigure}[b]{0.24\textwidth}
    \includegraphics[width=\textwidth]{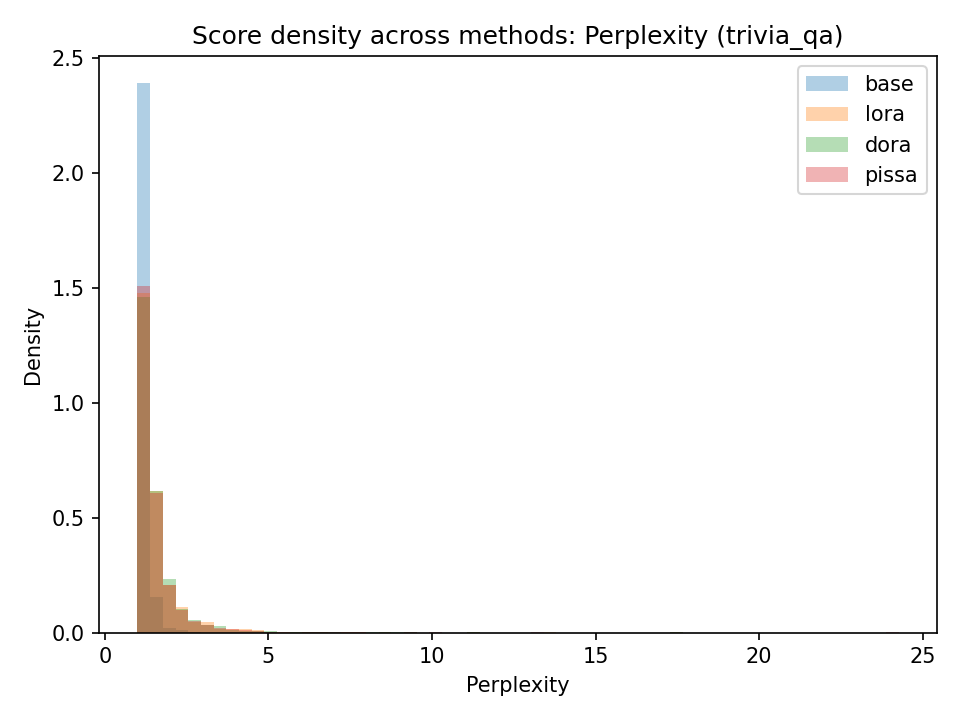}
    \caption{Perplexity}
\end{subfigure}
\hfill
\begin{subfigure}[b]{0.24\textwidth}
    \includegraphics[width=\textwidth]{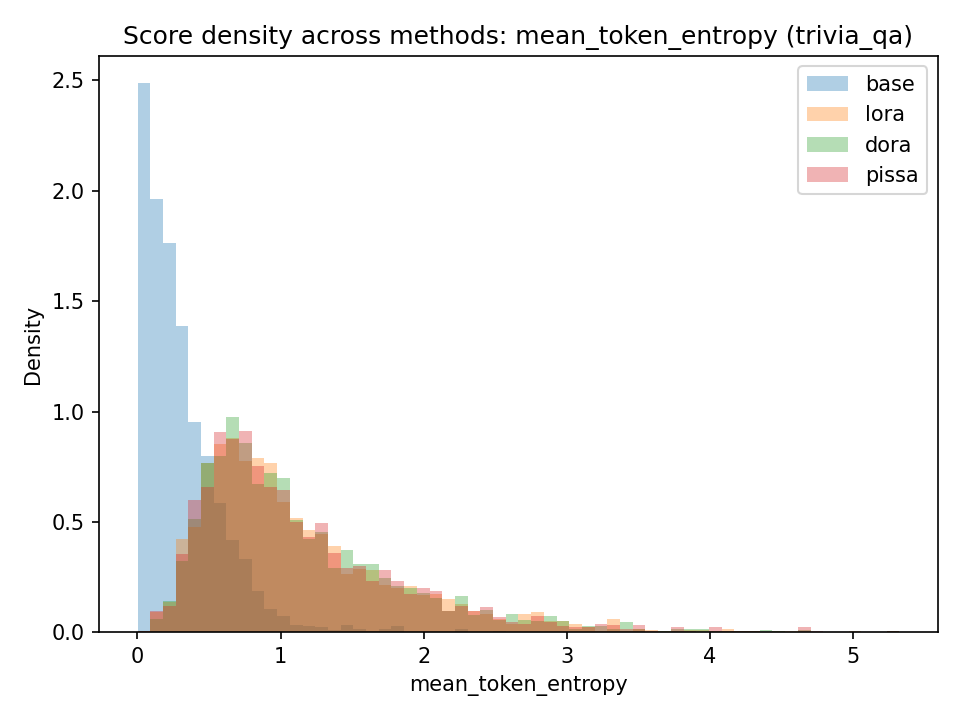}
    \caption{Mean Token Entropy}
\end{subfigure}
\hfill
\begin{subfigure}[b]{0.24\textwidth}
    \includegraphics[width=\textwidth]{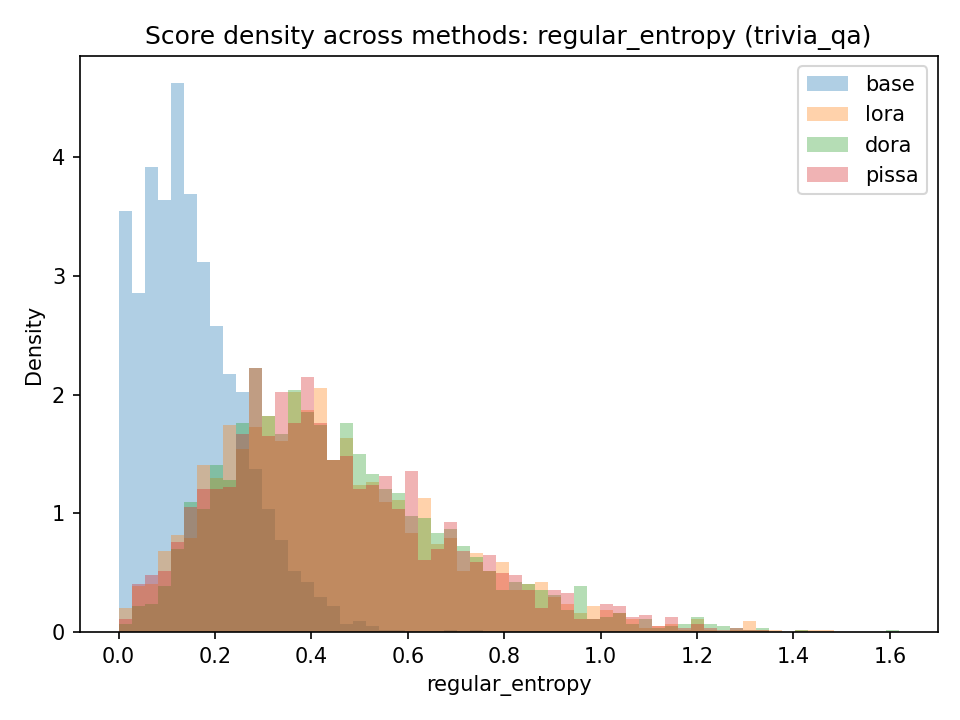}
    \caption{Predictive Entropy}
\end{subfigure}

\caption{Uncertainty score density distributions across parameter-efficient methods on Mistral-7B-Instruct-v0.3 (TriviaQA). Top row: semantic-level detectors. Bottom row: token-level  detectors. X-axis represents the uncertainty score.}
\label{fig:mistral_triviaqa_density}
\end{figure}

(Figure~\ref{fig:llama_squad_density} and Figure~\ref{fig:mistral_triviaqa_density}).
This section visualizes the distribution of uncertainty scores produced by different hallucination detectors before and after parameter-efficient fine-tuning. These density plots provide qualitative evidence for Takeaway~\#4, showing how parameter-efficient fine-tuning disperses degenerate near-zero uncertainty mass and restores meaningful separation between correct and incorrect responses.

\section{The reliability and danger ratio on Llama-3.2-3B across all datasets}
\label{sec:appendix_cc_analysis}
\begin{figure}[!htb]
\centering
\includegraphics[width=\textwidth]{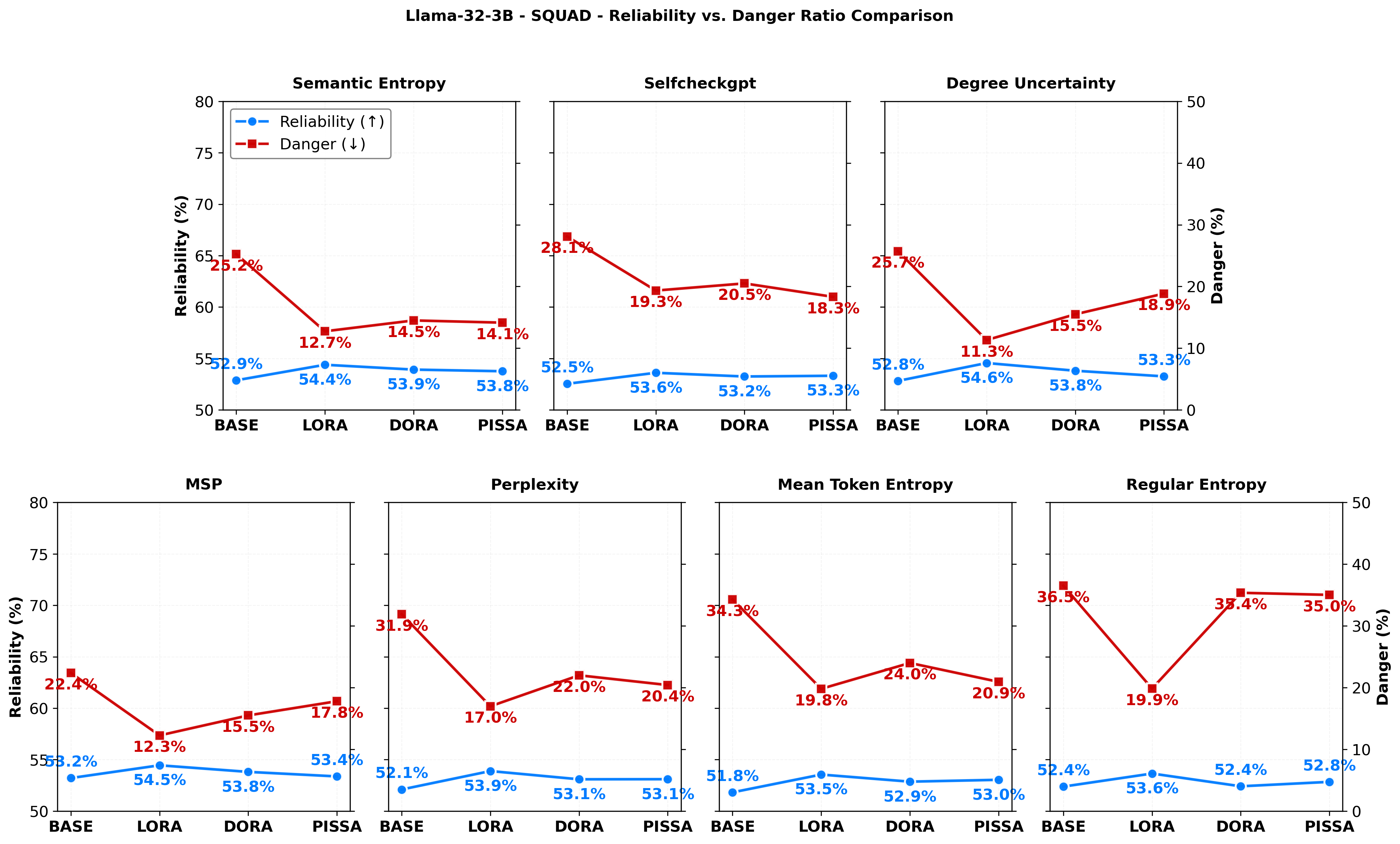}
\caption{Confidence-correctness safety analysis on SQuAD (Llama-3.2-3B). 
Reliability ($\uparrow$): precision of high-confidence predictions. 
Danger ($\downarrow$): fraction of errors delivered with high confidence. 
fine-tuning methods reduce dangerous hallucinations while maintaining prediction reliability.}
\label{fig:cc_squad}
\end{figure}

\begin{figure}[!htb]
\centering
\includegraphics[width=\textwidth]{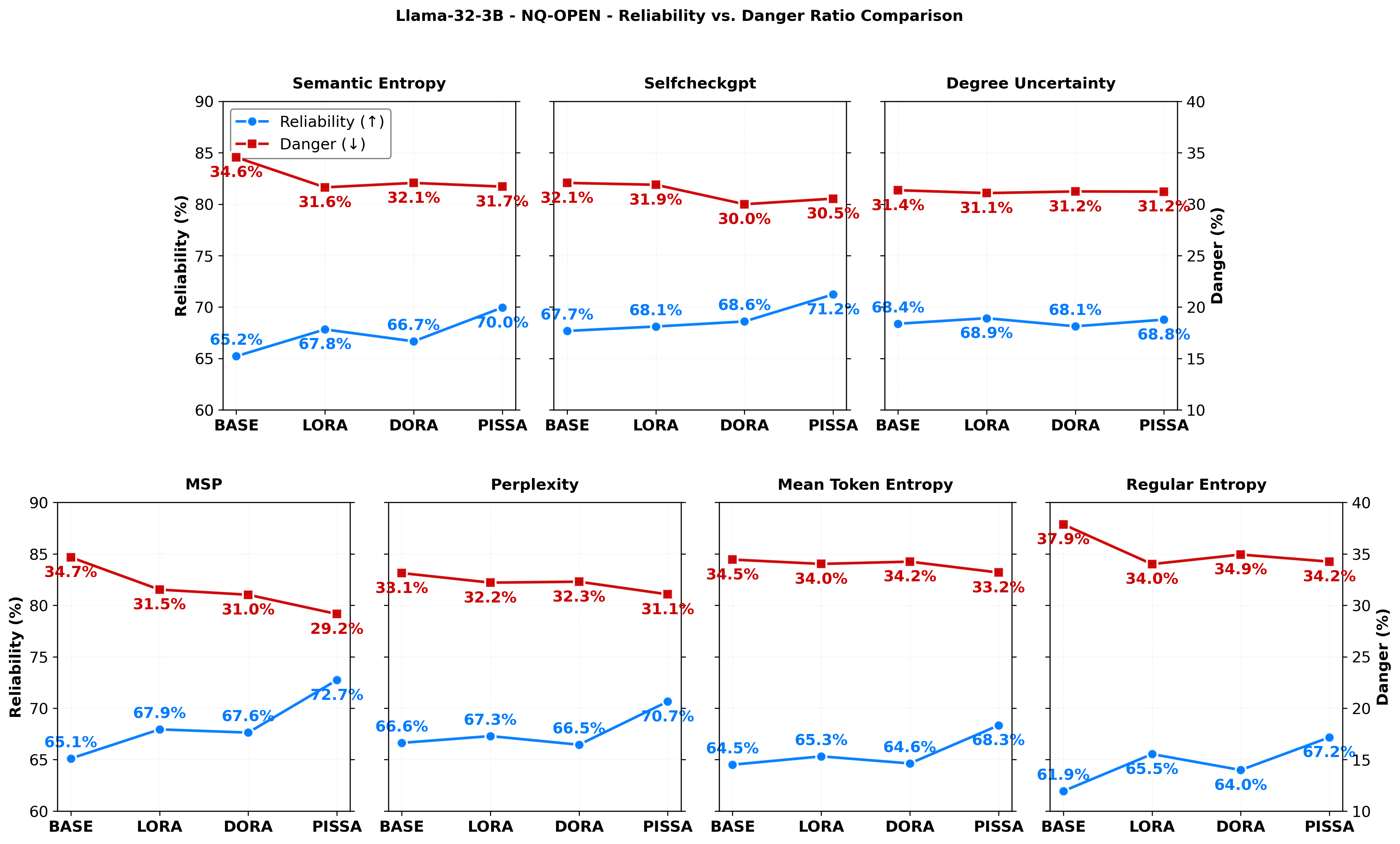}
\caption{Confidence-correctness safety analysis on NQ-Open (Llama-3.2-3B). 
Reliability ($\uparrow$): precision of high-confidence predictions. 
Danger ($\downarrow$): fraction of errors delivered with high confidence. 
fine-tuning methods reduce dangerous hallucinations while maintaining prediction reliability.}
\label{fig:cc_nqopen}
\end{figure}

(Figures~\ref{fig:cc_squad} and~\ref{fig:cc_nqopen}).
This section provides a comprehensive confidence and correctness analysis across datasets using reliability and danger ratios. The figures quantify how parameter-efficient fine-tuning shifts errors from confident hallucinations to detectable uncertain cases, offering behavioral evidence that parameter-efficient fine-tuning improves safety-relevant properties beyond aggregate AUROC metrics.

\section{PCA plots of hidden state in layer}
\begin{figure}[htbp]
\centering
\begin{tikzpicture}
\node[fill=blue, circle, inner sep=3pt, label=right:{\small Correct (label=1)}] at (0,0) {};
\node[fill=red, circle, inner sep=3pt, label=right:{\small Incorrect (label=0)}] at (4,0) {};
\end{tikzpicture}
\vspace{0.5em}

% 第一行: LLaMA-3.2-3B TriviaQA
\includegraphics[width=0.24\textwidth]{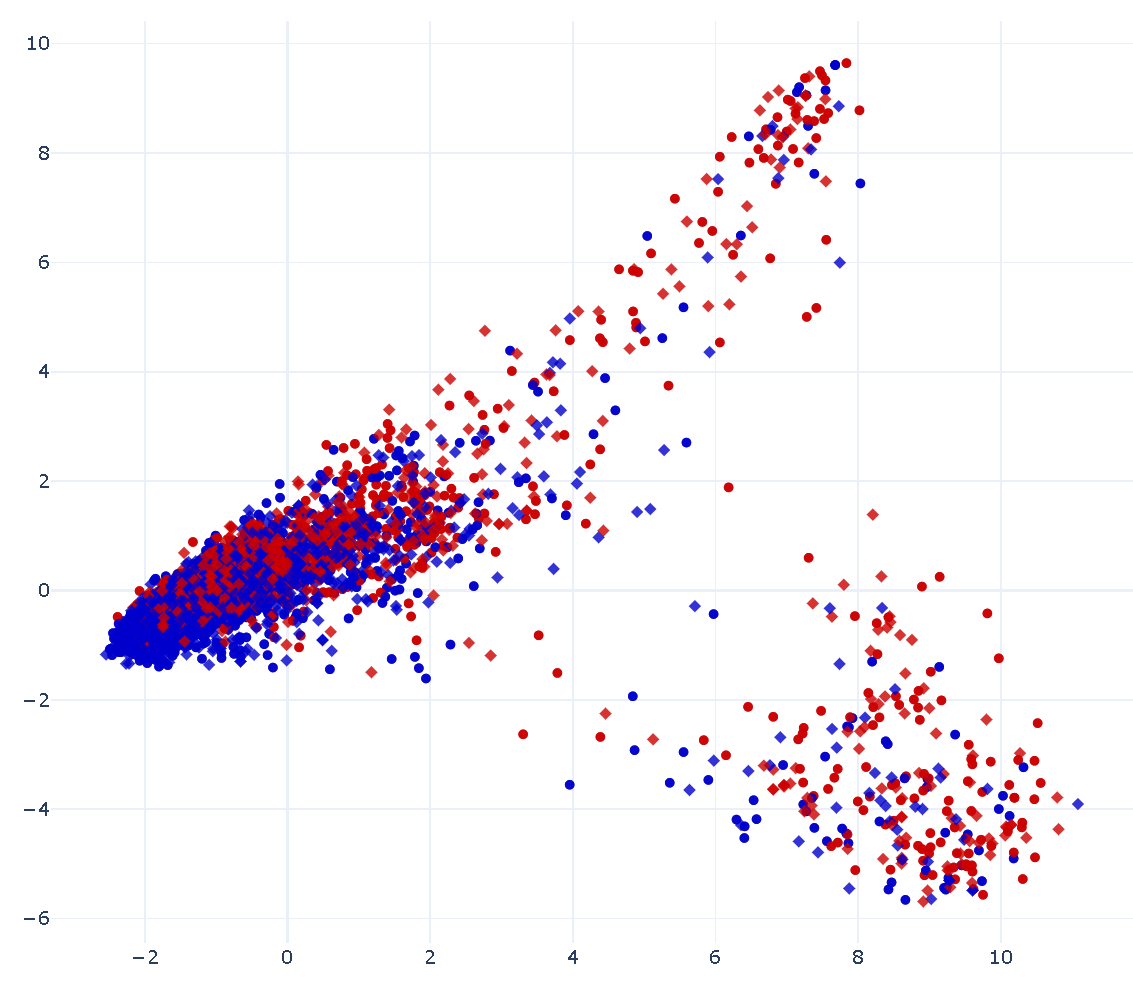}
\hfill
\includegraphics[width=0.24\textwidth]{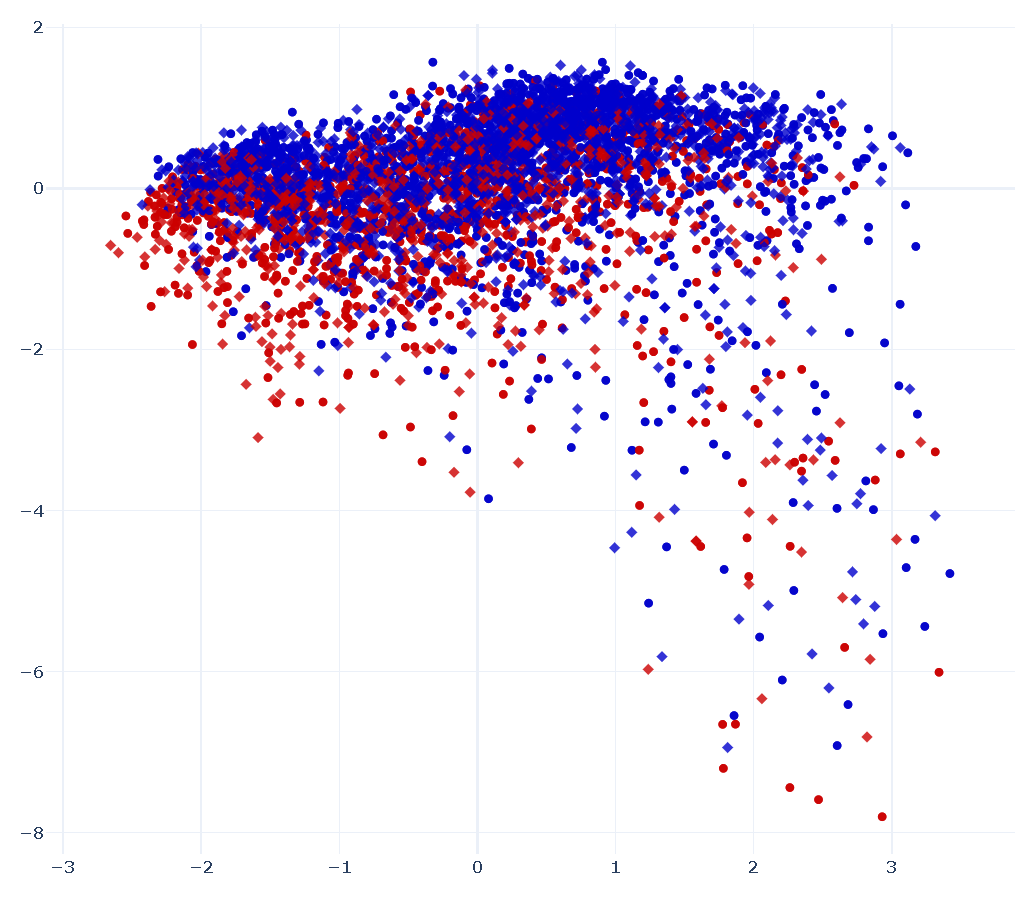}
\hfill
\includegraphics[width=0.24\textwidth]{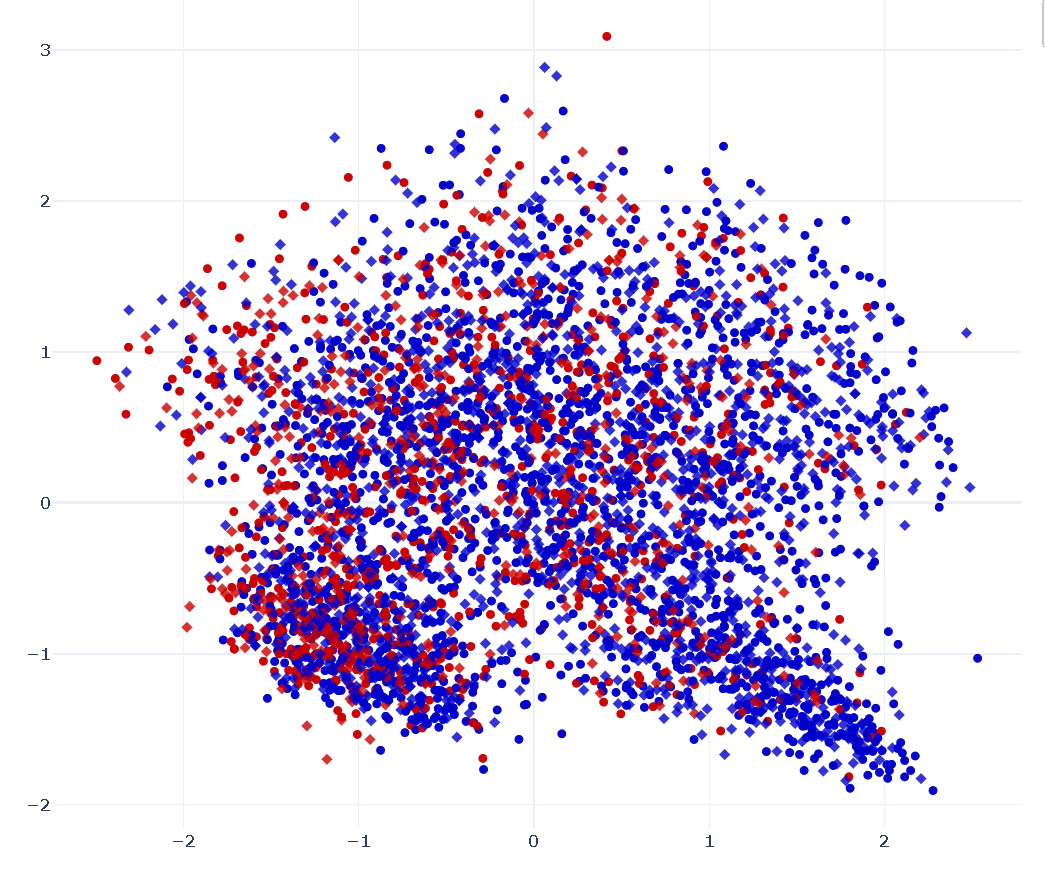}
\hfill
\includegraphics[width=0.24\textwidth]{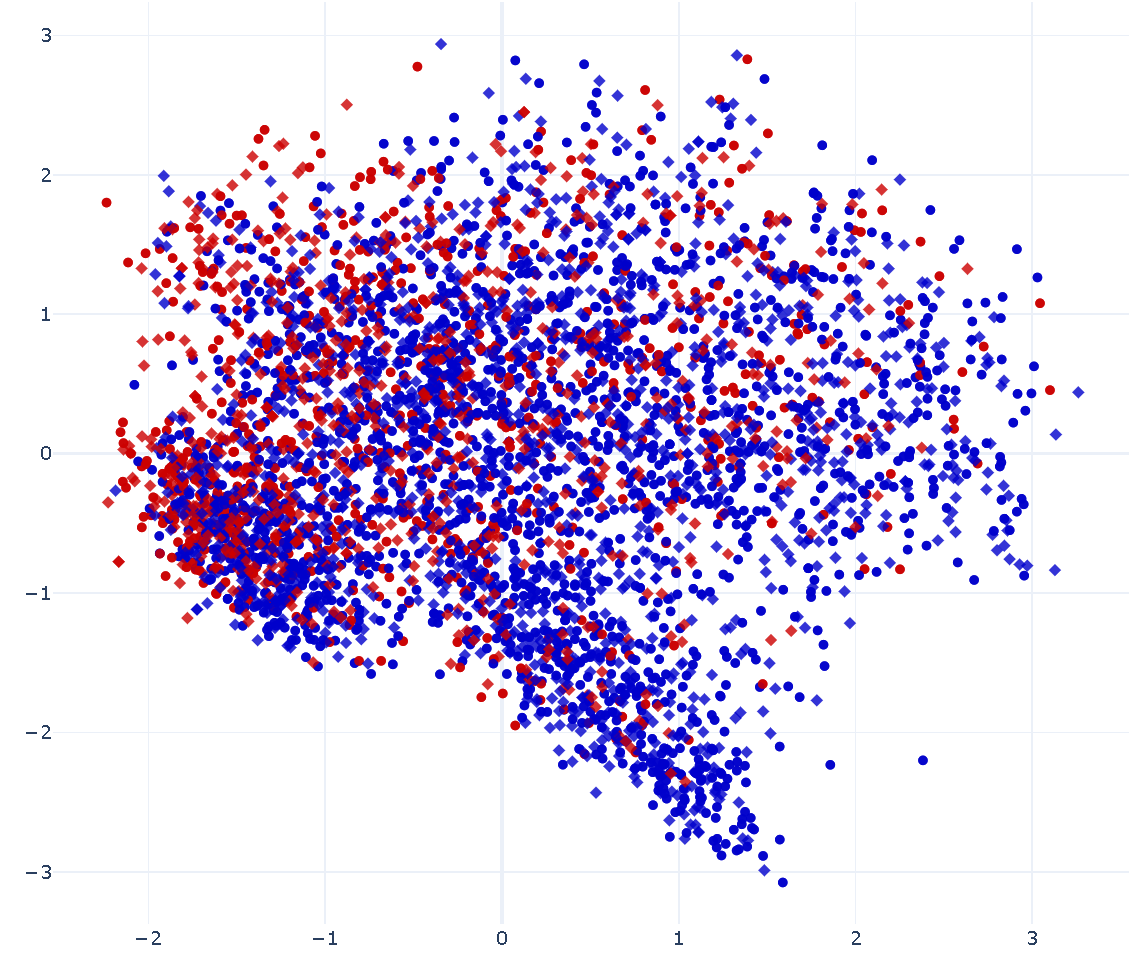}

\vspace{0.2em}
\textbf{LLaMA-3.2-3B-Instruct (TriviaQA)}

\vspace{0.8em}

% 第二行: Mistral-7B NQopen
\includegraphics[width=0.24\textwidth]{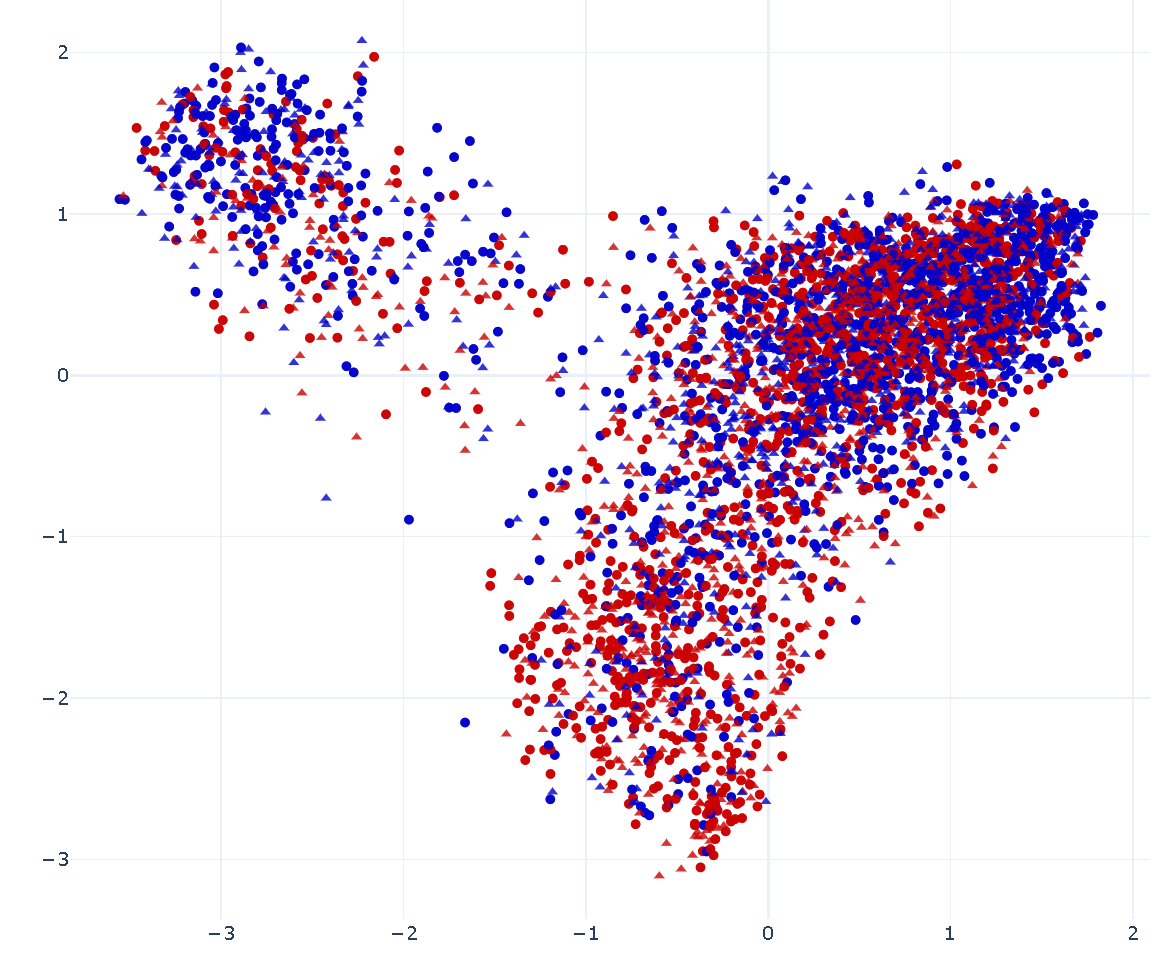}
\hfill
\includegraphics[width=0.24\textwidth]{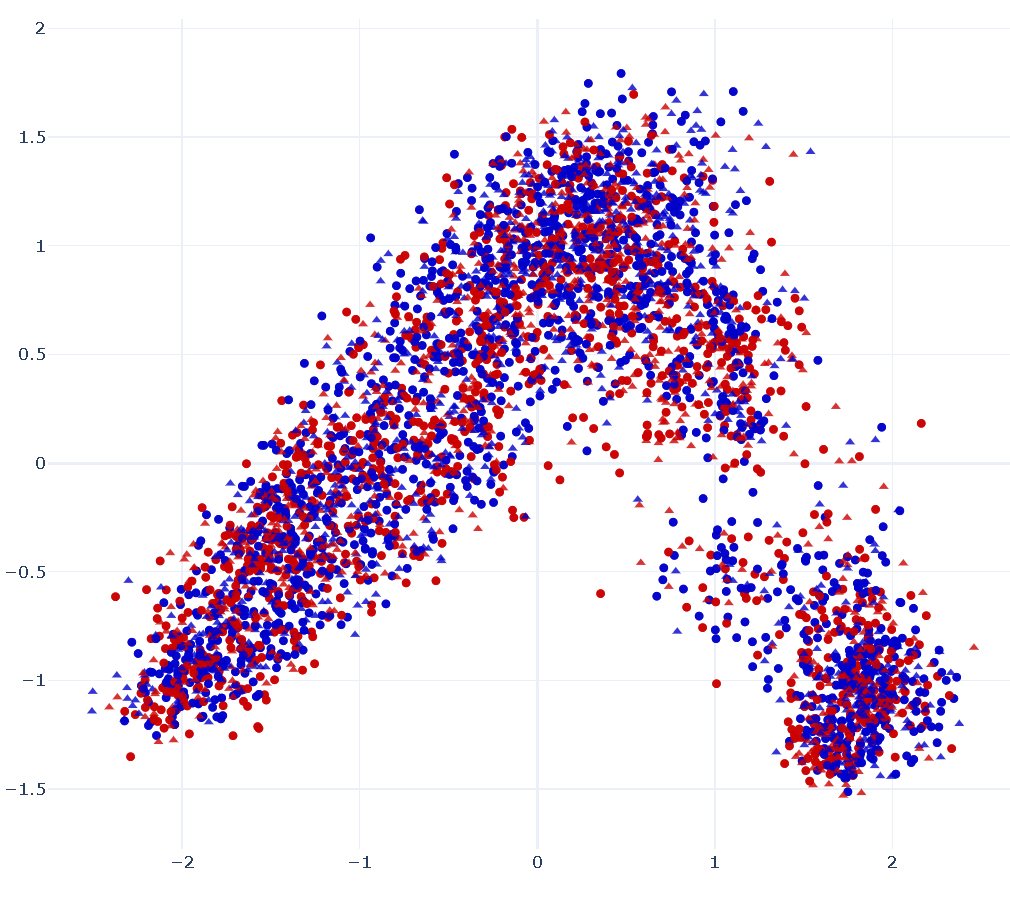}
\hfill
\includegraphics[width=0.24\textwidth]{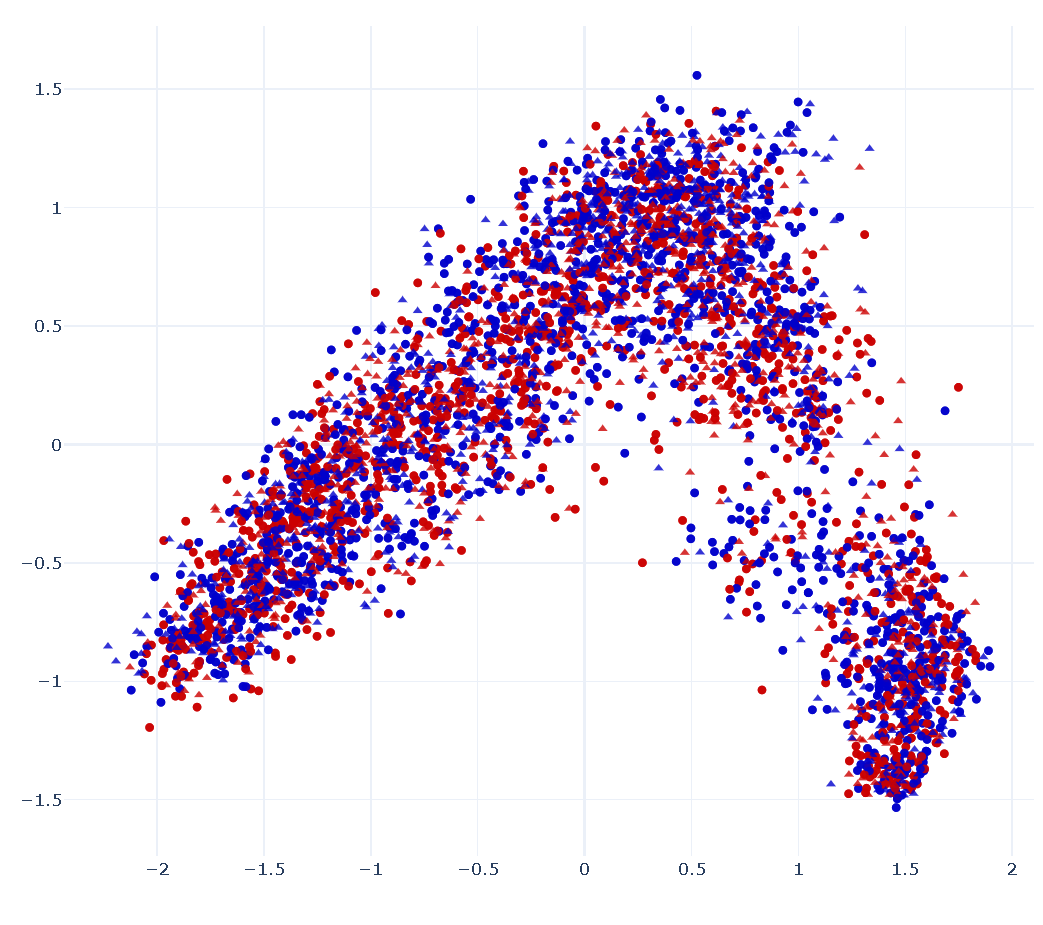}
\hfill
\includegraphics[width=0.24\textwidth]{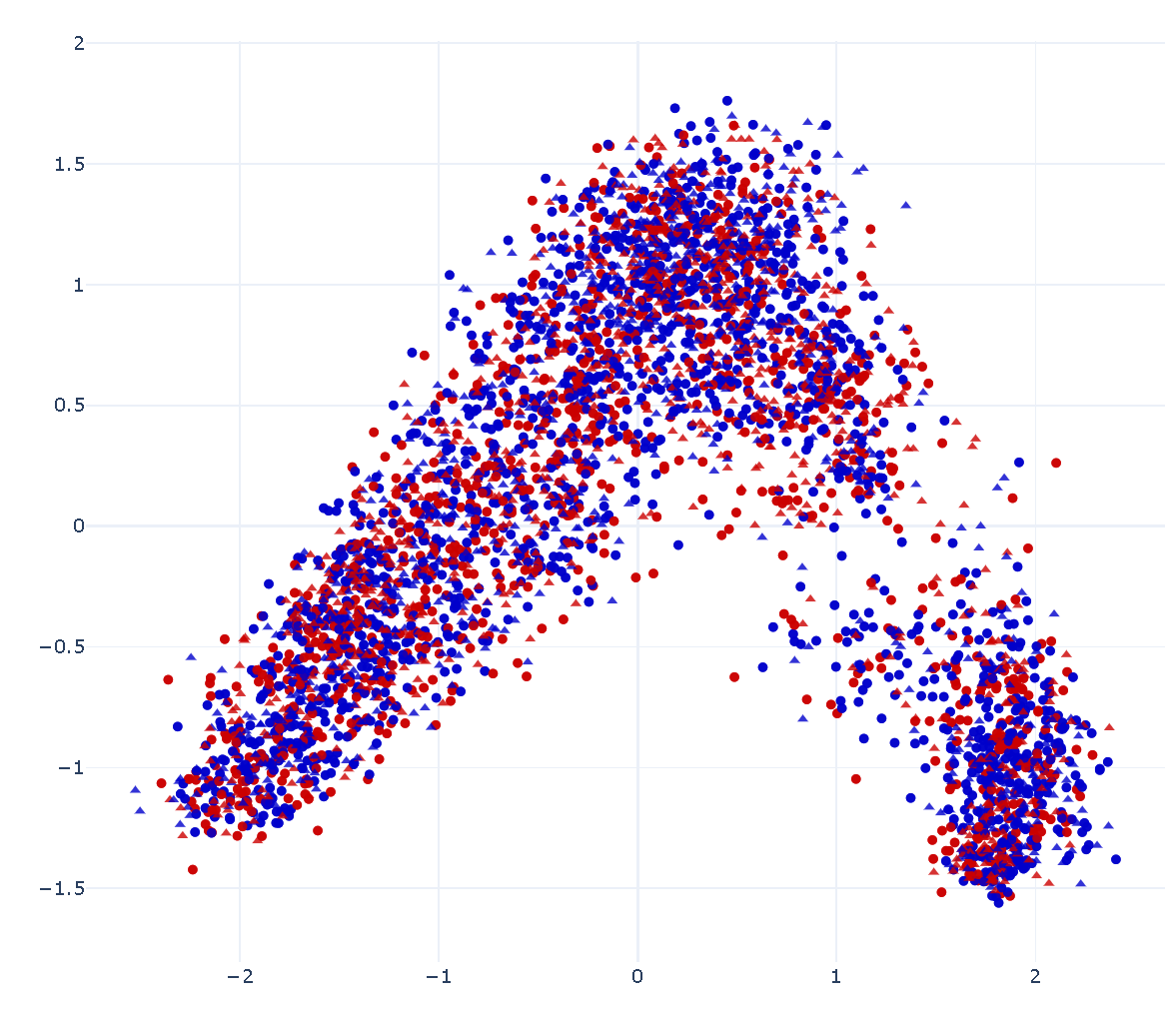}

\vspace{0.2em}
\textbf{Mistral-7B-Instruct-v0.3 (NQ-Open)}
\caption{PCA visualization of hidden representations at the best probing layer. Each row shows Base, LoRA, DoRA, and PiSSA (left to right). These plots show only the leading two principal components out of several thousand hidden dimensions, and are provided for qualitative illustration; the quantitative, full-dimensional separability test is the linear probe in Table~\ref{tab:best_probe_auroc_aupr}. In the leading components, fine-tuning does not produce clearer separation between correct and incorrect answers.}
\label{fig:pca_plots}
\end{figure}

(Figure.~\ref{fig:pca_plots}). This section presents PCA visualizations of hidden representations at the best probing layers. The plots provide geometric intuition for the probe results, illustrating that parameter-efficient fine-tuning does not necessarily increase linear separability between correct and incorrect answers, even when black-box hallucination detection improves.

We emphasize two caveats so that these visualizations are not over-interpreted. First, the projection is extremely lossy: it retains two directions out of several thousand hidden dimensions (for example, $3072$ for LLaMA-3.2-3B), so apparent inseparability in this 2D view is not, by itself, evidence of inseparability in the full representation space. The quantitative separability test in this paper is the full-dimensional linear probe of Table~\ref{tab:best_probe_auroc_aupr}, a logistic-regression classifier trained on the entire hidden-state vector; the inconsistency we report (Takeaway~\#3) is a statement about that high-dimensional linear classifier, not about the 2D projection. Second, our probes are linear, so they test linear separability alone. Whether the labels become separable under more flexible decision boundaries, for example an RBF-kernel SVM or a small MLP probe, is a robustness check we leave to future work. The qualitative pattern of the plots, namely that parameter-efficient fine-tuning does not visibly sharpen the correct/incorrect boundary in the leading components, is consistent with the probe results and is offered only as supporting illustration.
\section{Response Length Analysis}
\label{app:response_length}

This section provides the full generation length statistics referenced in the main text. We measure the token counts of model outputs on the test split of three datasets using LLaMA-3.2-3B-Instruct.

\begin{table}[!htbp]
\centering
\appendixtablestyle
\caption{Generation length statistics across datasets and adaptation methods (LLaMA-3.2-3B-Instruct, test split), including the base model with few-shot short-answer prompting (\texttt{Base (few-shot)}) used in the format control of Section~\ref{sec:format_control}, and the full-parameter SFT baseline (Full SFT) of Appendix~\ref{app:sft_baseline}. Note the standard deviation (Std): few-shot prompting lowers it only on the extractive SQuAD set, whereas weight-level adaptation lowers it on all three.}
\label{tab:generation_length_statistics}
\fontsize{8.6pt}{10.2pt}\selectfont
\setlength{\tabcolsep}{4pt}
\resizebox{0.8\textwidth}{!}{
\begin{tabular}{llccccccc}
\toprule
\textbf{Dataset} & \textbf{Method} & \textbf{Mean} & \textbf{Med.} & \textbf{Std} & \textbf{Min} & \textbf{Max} & \textbf{Q25} & \textbf{Q75} \\
\midrule
\cellcolor{gray!20}TriviaQA & \cellcolor{gray!20}Base  & \cellcolor{gray!20}7.07 & \cellcolor{gray!20}5.0  & \cellcolor{gray!20}4.30 & \cellcolor{gray!20}2 & \cellcolor{gray!20}20 & \cellcolor{gray!20}4.0 & \cellcolor{gray!20}8.0 \\
TriviaQA & Base (few-shot) & 5.89 & 4.0  & 4.42 & 1 & 20 & 3.0 & 7.0 \\
TriviaQA & LoRA  & 4.66 & 4.0  & 1.61 & 3 & 18 & 4.0 & 5.0 \\
TriviaQA & DoRA  & 4.68 & 4.0  & 1.58 & 3 & 17 & 4.0 & 5.0 \\
TriviaQA & PiSSA & 4.98 & 5.0  & 1.54 & 3 & 19 & 4.0 & 6.0 \\
TriviaQA & Full SFT & 4.02 & 4.0  & 1.61 & 2 & 13 & 3.0 & 5.0 \\
\midrule
\cellcolor{gray!20}NQ-Open  & \cellcolor{gray!20}Base  & \cellcolor{gray!20}11.92 & \cellcolor{gray!20}13.0 & \cellcolor{gray!20}5.02 & \cellcolor{gray!20}3 & \cellcolor{gray!20}20 & \cellcolor{gray!20}7.0 & \cellcolor{gray!20}16.0 \\
NQ-Open  & Base (few-shot) & 10.92 & 8.0  & 9.01 & 1 & 39 & 3.0 & 16.0 \\
NQ-Open  & LoRA  & 5.30  & 5.0  & 1.78 & 3 & 16 & 4.0 & 6.0 \\
NQ-Open  & DoRA  & 5.29  & 5.0  & 1.77 & 3 & 16 & 4.0 & 6.0 \\
NQ-Open  & PiSSA & 5.39  & 5.0  & 1.83 & 3 & 17 & 4.0 & 6.0 \\
NQ-Open  & Full SFT & 4.43  & 4.0  & 1.79 & 2 & 16 & 3.0 & 5.0 \\
\midrule
\cellcolor{gray!20}SQuAD    & \cellcolor{gray!20}Base  & \cellcolor{gray!20}9.29 & \cellcolor{gray!20}8.0 & \cellcolor{gray!20}4.72 & \cellcolor{gray!20}2 & \cellcolor{gray!20}22 & \cellcolor{gray!20}5.0 & \cellcolor{gray!20}12.0 \\
SQuAD    & Base (few-shot) & 4.16 & 4.0 & 2.42 & 1 & 12 & 2.0 & 5.0 \\
SQuAD    & LoRA  & 6.01 & 5.0 & 2.93 & 3 & 22 & 4.0 & 7.0 \\
SQuAD    & DoRA  & 6.39 & 5.0 & 3.17 & 3 & 22 & 4.0 & 7.0 \\
SQuAD    & PiSSA & 6.41 & 5.0 & 3.26 & 3 & 22 & 4.0 & 7.0 \\
SQuAD    & Full SFT & 5.28 & 4.0 & 3.14 & 2 & 22 & 3.0 & 6.0 \\
\bottomrule
\end{tabular}}
\end{table}

Across all three datasets, fine-tuned models generally generate shorter responses than the base model. This pattern is especially pronounced on TriviaQA and NQ-Open, where the mean and median generation lengths decrease substantially after fine-tuning. The reduced response length may partially reflect more concise answer generation behavior after task adaptation.
\section{Temperature Sensitivity}
\label{app:temperature_sensitivity}

We conduct a robustness check across different sampling temperatures. Specifically, Table~\ref{tab:temperature_sensitivity} reports AUROC for the base model and all three parameter-efficient methods at temperatures 0.6, 0.8, and 1.0. We include only detectors that require sampling multiple responses, since deterministic detectors are unaffected by temperature.

Overall, parameter-efficient improvements persist across all tested temperatures, supporting the robustness of our main conclusions.

\begin{table}[!htbp]
\centering
\appendixtablestyle
\caption{Hallucination detection performance across fine-tuning methods and sampling temperatures, on TriviaQA with LLaMA-3.2-3B-Instruct. Results are reported in AUROC.}
\label{tab:temperature_sensitivity}
\fontsize{8.6pt}{10.2pt}\selectfont
\resizebox{0.98\textwidth}{!}{
\begin{tabular}{lcccc}
\toprule
\textbf{Configuration} & \textbf{Degree of Uncertainty} & \textbf{Predictive Entropy} & \textbf{SelfCheckGPT} & \textbf{Semantic Entropy} \\
\midrule
\cellcolor{gray!20}Base, $T$=0.6  & \cellcolor{gray!20}0.8500 & \cellcolor{gray!20}0.7793 & \cellcolor{gray!20}0.8419 & \cellcolor{gray!20}0.8155 \\
LoRA, $T$=0.6  & 0.8603 & 0.7943 & 0.8596 & 0.8524 \\
DoRA, $T$=0.6  & 0.8921 & 0.8072 & 0.8886 & 0.8717 \\
PiSSA, $T$=0.6 & 0.8821 & 0.8086 & 0.8813 & 0.8759 \\
\midrule
\cellcolor{gray!20}Base, $T$=0.8  & \cellcolor{gray!20}0.8661 & \cellcolor{gray!20}0.7815 & \cellcolor{gray!20}0.8620 & \cellcolor{gray!20}0.8259 \\
LoRA, $T$=0.8  & 0.8831 & 0.7891 & 0.8643 & 0.8514 \\
DoRA, $T$=0.8  & 0.8914 & 0.8003 & 0.8955 & 0.8822 \\
PiSSA, $T$=0.8 & 0.8841 & 0.8112 & 0.8880 & 0.8686 \\
\midrule
\cellcolor{gray!20}Base, $T$=1.0  & \cellcolor{gray!20}0.8703 & \cellcolor{gray!20}0.7819 & \cellcolor{gray!20}0.8737 & \cellcolor{gray!20}0.8205 \\
LoRA, $T$=1.0  & 0.8779 & 0.8216 & 0.8889 & 0.8737 \\
DoRA, $T$=1.0  & 0.8738 & 0.8138 & 0.8838 & 0.8823 \\
PiSSA, $T$=1.0 & 0.8772 & 0.8087 & 0.8896 & 0.8797 \\
\bottomrule
\end{tabular}}
\end{table}

\section{Rank Sensitivity and MLP Inclusion Tests}
\label{app:rank_sensitivity_mlp}
To verify that the detectability improvement is not a narrow artifact of a single low-rank configuration, we evaluate ranks 8, 16, and 32 for three low-rank methods (LoRA, DoRA, and PiSSA), targeting both attention and MLP projection layers on TriviaQA using LLaMA-3.2-3B-Instruct. Results are reported in Tables~\ref{tab:rank_sensitivity_appendix} and \ref{tab:mlp_inclusion_appendix}, and all metrics are AUROC. Overall, detection improvements are consistent across ranks 8, 16, and 32. These results further support our main takeaways that fine-tuning improves semantic-consistency-based and confidence-based hallucination detectors. Table~\ref{tab:rank_sensitivity_appendix} shows that this pattern is stable across different ranks, while Table~\ref{tab:mlp_inclusion_appendix} shows that it also holds when comparing Attention-only tuning against MLP+Attn.

\begin{table}[!htbp]
\centering
\appendixtablestyle
\caption{Hallucination detection performance across low-rank configurations (rank sensitivity). All results are AUROC on TriviaQA with LLaMA-3.2-3B-Instruct. In the first column, each configuration specifies the low-rank method, rank, and target modules.}
\label{tab:rank_sensitivity_appendix}
\fontsize{8.6pt}{10.2pt}\selectfont
\resizebox{0.98\textwidth}{!}{%
\begin{tabular}{lccccccc}
\toprule
\textbf{Configuration} & \textbf{SE} & \textbf{SC} & \textbf{Deg} & \textbf{MSP} & \textbf{Perplexity} & \textbf{MeanEnt} & \textbf{PreEnt} \\
\midrule
\cellcolor{gray!20}Base                  & \cellcolor{gray!20}0.8205 & \cellcolor{gray!20}0.8737 & \cellcolor{gray!20}0.8703 & \cellcolor{gray!20}0.8154 & \cellcolor{gray!20}0.8328 & \cellcolor{gray!20}0.8457 & \cellcolor{gray!20}0.7819 \\
\midrule
LoRA 8, MLP+Attn      & 0.8711 & 0.8807 & 0.8726 & 0.8742 & 0.8574 & 0.8334 & 0.8063 \\
LoRA 16, MLP+Attn     & 0.8743 & 0.8801 & 0.8737 & 0.8707 & 0.8546 & 0.8344 & 0.8037 \\
LoRA 32, MLP+Attn     & 0.8703 & 0.8798 & 0.8728 & 0.8698 & 0.8522 & 0.8256 & 0.7980 \\
\midrule
DoRA 8, MLP+Attn      & 0.8778 & 0.8927 & 0.8836 & 0.8827 & 0.8615 & 0.8322 & 0.8094 \\
DoRA 16, MLP+Attn     & 0.8748 & 0.8857 & 0.8796 & 0.8734 & 0.8492 & 0.8236 & 0.7968 \\
DoRA 32, MLP+Attn     & 0.8759 & 0.8881 & 0.8745 & 0.8757 & 0.8529 & 0.8252 & 0.8042 \\
\midrule
PiSSA 8, MLP+Attn     & 0.8726 & 0.8897 & 0.8814 & 0.8832 & 0.8602 & 0.8273 & 0.8115 \\
PiSSA 16, MLP+Attn    & 0.8721 & 0.8784 & 0.8712 & 0.8763 & 0.8578 & 0.8358 & 0.8167 \\
PiSSA 32, MLP+Attn    & 0.8698 & 0.8794 & 0.8797 & 0.8694 & 0.8570 & 0.8340 & 0.8098 \\
\bottomrule
\end{tabular}}
\end{table}

\begin{table}[!htbp]
\centering
\appendixtablestyle
\caption{Hallucination detection performance across fine-tuning methods and target modules. All results are AUROC on TriviaQA with LLaMA-3.2-3B-Instruct.}
\label{tab:mlp_inclusion_appendix}
\resizebox{0.98\textwidth}{!}{%
\begin{tabular}{llccccccc}
\toprule
\textbf{Configuration} & \textbf{Modules} & \textbf{SE} & \textbf{SC} & \textbf{Deg} & \textbf{MSP} & \textbf{Perplexity} & \textbf{MeanEnt} & \textbf{PreEnt} \\
\midrule
\cellcolor{gray!20}Base  & \cellcolor{gray!20}---       & \cellcolor{gray!20}0.8205 & \cellcolor{gray!20}0.8737 & \cellcolor{gray!20}0.8703 & \cellcolor{gray!20}0.8154 & \cellcolor{gray!20}0.8328 & \cellcolor{gray!20}0.8457 & \cellcolor{gray!20}0.7819 \\
\midrule
LoRA  & Attn only & 0.8737 & 0.8889 & 0.8779 & 0.8656 & 0.8597 & 0.8303 & 0.8216 \\
LoRA  & MLP+Attn  & 0.8613 & 0.8798 & 0.8728 & 0.8648 & 0.8462 & 0.8256 & 0.7880 \\
\midrule
DoRA  & Attn only & 0.8823 & 0.8838 & 0.8738 & 0.8866 & 0.8605 & 0.8202 & 0.8138 \\
DoRA  & MLP+Attn  & 0.8759 & 0.8881 & 0.8745 & 0.8757 & 0.8529 & 0.8252 & 0.8042 \\
\midrule
PiSSA & Attn only & 0.8797 & 0.8896 & 0.8772 & 0.8878 & 0.8595 & 0.8275 & 0.8087 \\
PiSSA & MLP+Attn  & 0.8698 & 0.8794 & 0.8697 & 0.8694 & 0.8570 & 0.8340 & 0.8098 \\
\bottomrule
\end{tabular}}
\end{table}

These results show that the gains are not tied to a single rank choice. In addition, tuning attention modules alone already yields strong improvements, while extending tuning to both MLP and attention layers preserves the same qualitative conclusions.

\FloatBarrier
\section{Full Supervised Fine-Tuning (SFT) Baseline}
\label{app:sft_baseline}

Section~\ref{sec:takeaway2} reports black-box detection under full-parameter SFT in Table~\ref{tab:sft_blackbox}. This appendix records the protocol behind those runs and the AUPR evidence that accompanies them.

\paragraph{Protocol.}
We run one full-parameter SFT configuration per backbone and dataset on the same in-domain QA training sets used for the parameter-efficient runs, with no method-specific hyperparameter sweep. Detection follows the protocol applied to the Base and parameter-efficient runs: the three sample-based semantic-consistency detectors (SE, SC, Deg) are averaged over the same three random seeds and reported as mean $\pm$ standard deviation, and the confidence and entropy detectors come from a single representative seed. The Full SFT entries in the two appendix AUPR tables (Tables~\ref{tab:triviaqa_aupr} and~\ref{tab:nqopen_aupr}) are single-run point estimates. Because these runs are untuned, they stay outside the column-wise bolding that marks the strongest tuned method, and the \textbf{Avg Impr.} rows of Tables~\ref{tab:llama_auroc},~\ref{tab:qwen_auroc},~\ref{tab:mistral_auroc} and~\ref{tab:squad_aupr} average the three parameter-efficient methods alone.

\paragraph{AUPR evidence.}
The AUROC pattern of Table~\ref{tab:sft_blackbox} carries over to AUPR. In Tables~\ref{tab:triviaqa_aupr} and~\ref{tab:nqopen_aupr}, semantic-consistency and confidence detectors gain under full SFT on both datasets, and several detector--dataset pairs reach scores at or above the best parameter-efficient method by a wide margin: SE on TriviaQA with Mistral-7B reaches an AUPR of $0.843$ against $0.729$ for the best parameter-efficient method. This is confirmatory evidence that the effect arises from in-domain task supervision: it is a property of in-domain adaptation in general, and it holds beyond low-rank parameterizations. The paper studies parameter-efficient fine-tuning in depth across three methods, three backbones, and three datasets because parameter-efficient fine-tuning is the dominant adaptation method in practice, and these single-configuration SFT runs establish that the same qualitative pattern extends outside the low-rank family. Parameter-efficient fine-tuning remains practically attractive because it reaches this pattern while updating a small fraction of parameters.

\FloatBarrier

\FloatBarrier

\section{Ethical Considerations}
We used AI assistants (such as Claude, ChatGPT and Gemini) during the research process for: (1) polishing and editing the manuscript for clarity and grammar; (2) providing suggestions for figure design and visualization; and (3) assisting with code implementation and debugging. All AI-generated content was carefully reviewed, verified, and revised by the authors, who take full responsibility for the final manuscript.

% ============ LLaMA SQuAD ============

% ============ Mistral TriviaQA ============

\clearpage
\end{document}